\documentclass[preprint,12pt,sort,numbers]{elsarticle}

\usepackage{xcolor}
\usepackage{placeins}
\usepackage[most]{tcolorbox}
\usepackage{amssymb}
\usepackage{amsmath}
\usepackage{tabularx}
\usepackage{listings}
\usepackage{float}
\usepackage{booktabs} 
\usepackage{multirow}
\usepackage[hyphens,spaces,obeyspaces]{url}
\usepackage{subcaption} 
\usepackage{rotating} 
\usepackage{makecell} 

\begin{document}

\begin{frontmatter}



\title{%
An Experimental Comparison of the Most Popular Approaches to Fake News Detection\\[0.5em]
\large\bfseries \textcolor{red}{Preprint version of the paper in \textit{Information Sciences}, doi: \url{https://doi.org/10.1016/j.ins.2026.123407}}
}

\author[dii]{Pietro Dell'Oglio}
\ead{pietro.delloglio@ing.unipi.it}

\author[di]{Alessandro Bondielli}
\ead{alessandro.bondielli@unipi.it}

\author[dii]{Francesco Marcelloni\corref{cor1}}
\ead{francesco.marcelloni@unipi.it}

\author[di]{Lucia C. Passaro}
\ead{lucia.passaro@unipi.it}

\address[dii]{Dipartimento di Ingegneria dell'Informazione, University of Pisa, Largo Lucio Lazzarino, 1, Pisa, Italy}

\address[di]{Dipartimento di Informatica, University of Pisa, Largo B. Pontecorvo, 3, Pisa, Italy}

\cortext[cor1]{Corresponding author}

\begin{abstract}
In recent years, fake news detection has received increasing attention in public debate and scientific research. Despite advances in detection techniques, the production and spread of false information have become more sophisticated, driven by Large Language Models (LLMs) and the amplification power of social media.

We present a critical assessment of 12 representative fake news detection approaches, spanning traditional machine learning, deep learning, transformers, and specialized cross-domain architectures. We evaluate these methods on 10 publicly available datasets differing in genre, source, topic, and labeling rationale. We address text-only English fake news detection as a binary classification task by harmonizing labels into “Real” and “Fake” to ensure a consistent evaluation protocol. We acknowledge that label semantics vary across datasets and that harmonization inevitably removes such semantic nuances. Each dataset is treated as a distinct domain.

We conduct in-domain, multi-domain and cross-domain experiments to simulate real-world scenarios involving domain shift and out-of-distribution data.
Fine-tuned models perform well in-domain but struggle to generalize. Cross-domain architectures can reduce this gap but are data-hungry, while LLMs offer a promising alternative through zero- and few-shot learning. Given inherent dataset confounds  and possible pre-training exposure, results should be interpreted as robustness evaluations within this English, text-only protocol.

\end{abstract}




\end{frontmatter}

\section{Introduction}\label{sec:intro} 
The growing presence of fake news on social media has become a major issue requiring urgent attention. This trend is driven by the increasing role of social media as a tool for news dissemination and consumption \citep{newman2013social}. Journalists leverage these platforms to engage public opinion and breaking news stories, while users rely on verified social media accounts or their personal networks to stay informed on current events. Social networks have proven to be particularly valuable during crises, as they can disseminate news faster than traditional media outlets \citep{vieweg2010microblogged}.

The distorted use of social media has become increasingly evident in recent years, as exemplified by the rise of the first ``infodemic'' during the COVID-19 pandemic. Scholars have described this period as a Post-Truth Era \citep{lewandowsky2017beyond}, where emotions and misinformation often overshadow factual accuracy \citep{alam2021survey}. The Russo-Ukrainian conflict has exacerbated this trend, with disinformation emerging as a strategic tool, as it has often been experimented in conflicts \citep{patwa2021fighting}.

The lack of control and fact-checking mechanisms on social media platforms facilitates the spread of unverified and false information, which can substantially influence public opinion on crucial issues \citep{zubiaga2018detection}. A notable example was the 2016 USA presidential election campaign, where the proliferation of fake news played a significant role \citep{allcott2017social}.
In this context, the growing  popularity and accessibility of Large Language Models (LLMs) further enable the rapid creation of credible yet false information \citep{spitale2023ai,sun2024exploring}.

Fake news on social media can take various forms, ranging from simple rumors to misleading clickbait articles, making effective detection and mitigation challenging. This diversity has led to numerous independent fact-checking initiatives and has stimulated a surge of research in computational detection, where the problem is often framed as a supervised classification task \citep{shu2017fake,zhou2020survey,nakov2022overview,guo2022survey}.
However, this large body of literature suffers from a critical, implicit limitation: models are typically trained and evaluated in-domain, using test data drawn from the same distribution as the training data. This leads to a false sense of security. As noted by \citep{janicka2019cross}, models that achieve high accuracy by overfitting to dataset-specific artifacts (e.g., topics, writing styles, labelling scheme) often fail when applied to new, unseen domains---a scenario known as domain shift. The real-world applicability of a fake news detector depends entirely on its ability to generalize.
To address this gap, we conduct a large-scale empirical benchmarking study explicitly focused on the cross-domain generalization capabilities of fake news detection models. 

In this work, we operationalize the domain as dataset identity. This choice intentionally conflates multiple sources of distribution shift that co-occur in realistic deployment scenarios, including topic shift (e.g., elections vs. pandemics), genre or medium shift (tweets, short claims, full articles), source and platform shift, temporal shift, annotation and label-semantic shift, arising from different fact-checking and labeling policies (e.g., treatment of satire or partially false claims). While pre-processing steps such as label harmonization may introduce some mismatch in the label semantics, we consider it a realistic and unavoidable component of cross-dataset domain shift. Rather than isolating these factors, our goal is to assess model robustness under their joint effect, reflecting real-world conditions. 

At the same time, it is important to distinguish these sources of variation conceptually. Topic, time, source, and annotation differences represent distinct confounding factors that cannot be causally disentangled in our experimental setting. Throughout the paper, we therefore interpret our findings as empirical observations about how models behave under naturally bundled shifts, rather than as causal claims about which specific factor drives performance changes. The term ``domain shift'' is therefore used in an operational sense, referring to the combined effect of these intertwined sources of variation.

Our study focuses exclusively on English, text-only detection. This is a deliberate scope choice aimed at isolating the generalization behaviour of textual classifiers, without incorporating multimodal cues (e.g., images, videos), social-context signals (e.g., propagation patterns, user metadata), or retrieval-augmented fact-checking components. We do so to allow for a controlled comparison across models. Nevertheless, we acknowledge that it may limit generalizability of our findings to scenarios where such signals play a central role. Our results should therefore be interpreted as lower-bound estimates of robustness, complementary to multimodal and context-aware approaches rather than competing with them.

Furthermore, it is important to note that our evaluation is subject to several constraints, including the fact that we do not account for the time of production or diffusion of fake news, so we cannot control for potential exposure of LLMs to similar content during pretraining. In addition, we do not explicitly evaluate LLM‑generated fake news, and rely on datasets collected and annotated under heterogeneous protocols.

We design four experimental groups to rigorously test model robustness: (1) a baseline Dataset-Specific test, (2) a Cross-Dataset test to expose the domain shift problem, (3) a Mixed-Training test to simulate a multi-domain scenario, and (4) a robust Leave-One-Dataset-Out experiment to assess cross-domain generalization.

Across these settings, we compare twelve representative approaches spanning traditional machine learning (ML), deep learning (DL), Transformer-based models, specialized cross-domain architectures, and decoder-only large language models (LLMs). Note that in the rest of the paper we differentiate Traditional ``ML'' and ``DL'' purely on a technical basis, i.e., based on whether or not they use Deep Neural Networks, regardless of their internal workings.

Our findings reveal a critical failure in generalization for most standard approaches. We show that fine-tuned models excel in-domain, but struggle to generalize. Conversely, specialized cross-domain architectures prove effective in Leave-One-Dataset-Out scenarios, but they are highly data-hungry and perform poorly when training resources are limited. This underscores a central challenge in the field and highlights the need for data-efficient, robust models. Our results also suggest that the transfer learning abilities of LLMs offer a promising starting point. Although they underperform compared to specialized in-domain models and data-hungry architectures, they provide a valuable compromise between generalization capability and dependence on large, domain-specific training data.

In conclusion, this work provides a large-scale, systematic empirical benchmark aimed at assessing the generalization capabilities of existing approaches under realistic domain-shift conditions. Comparing algorithms under a unified experimental protocol, this study aims to serve as a reference point for future research on robust and deployable fake news detection systems.
Specifically, we:

\begin{itemize}
    \item evaluate 12 representative methods spanning traditional machine learning, deep learning, Transformer-based models, specialized cross-domain architectures, and decoder-only Large Language Models;
    \item conduct experiments across 10 heterogeneous datasets covering both generic and event-specific domains;
    \item design four complementary experimental settings, including dataset-specific, cross-dataset, mixed-domain training, and leave-one-dataset-out, to progressively stress-test robustness; and
    \item provide a comparative analysis that highlights when and why different paradigms fail or succeed beyond in-domain evaluation.
\end{itemize}

The remainder of the paper is structured as follows. Section \ref{sec:related} reviews related work and position this work within the existing literature. Sections \ref{sec:experimentalSetup}–\ref{sect:discussion} present the experimental design, results, and discussion.  Specifically, Section \ref{sec:experimentalSetup} provides an overview of our experimental setup, including datasets, models, and evaluation protocol.  In Section \ref{sec:experiments} we discuss the results of the four experimental scenarios, namely Dataset-Specific experiments (Section \ref{sect:exp1}), Cross-Dataset experiments (Section \ref{sect:exp2}), Mixed-Training/Single-Test experiments (Section \ref{sect:exp3}), and Leave-One-Dataset-Out experiments (Section \ref{sect:exp4}). Section \ref{sect:discussion} synthesizes and discusses the overall findings emerging from these results.  
We then discuss some potential limitations of this work in Section \ref{sec:limitations}, and finally draw some conclusions and outline future direction in Section \ref{sec:conclusions}.

\section{Related Work}\label{sec:related}

Fake news detection has received substantial attention in recent years, leading to a wide range of proposed approaches and empirical studies. Most of this literature frames the task as a supervised text classification problem and reports promising results. However, a critical limitation shared by the majority of existing works is their predominant focus on \emph{in-domain evaluation}, where training and test data are drawn from the same dataset or distribution. As a consequence, reported performance may largely reflect the exploitation of dataset-specific artifacts rather than the ability to generalize to new, unseen domains and data.

\textbf{Early evidence of domain shift.}
The work of Janicka et al.~\cite{janicka2019cross} was among the first to systematically demonstrate this issue. By training traditional ML classifiers on one dataset and testing them on others, the authors showed that performance can drop by more than 20\% in cross-dataset settings. Their findings showed that high in-domain accuracy is often tightly coupled with dataset-specific lexical and stylistic characteristics, motivating the need for explicit cross-domain evaluation. This observation directly informs the design of our experimental settings.

\textbf{Architectures designed for generalization.}
In response to these limitations, recent research has proposed models that explicitly target cross-domain robustness. Two main architectural strategies have emerged.
The first aims to learn \emph{domain-invariant representations} by encouraging models to capture signals of misinformation that are independent of topic or source. This is typically achieved by means of adversarial or self-supervised training. Representative approaches include REAL-FND~\citep{mosallanezhad2022domain}, DAFNE~\citep{comito2025learning}, and FADED~\citep{wei2025cross}.
The second strategy relies on \emph{multi-expert architectures}, most notably Mixture-of-Experts (MoE) models, which combine multiple domain-specialized components. Methods such as MDFEND~\citep{nan2021mdfend} and M3FEND~\citep{zhu2022memory} are designed for multi-domain scenarios and employ domain embeddings to weight expert contributions. While effective when domains are known in advance, this reliance limits their applicability to unseen domains. More recent approaches, such as MERMAID~\citep{liguori2025breaking}, address this limitation by routing inputs based on content rather than explicit domain labels, enabling cross-domain zero-shot generalization.

\textbf{LLM paradigms for fake news detection.}
Early work on misinformation detection is dominated by supervised fine-tuning of pretrained language models, most notably BERT-based architectures, on labeled fake-news datasets \cite{zhou2020survey,kaliyar2021fakebert}. These approaches frame fake news detection as a standard text classification problem and typically achieve strong in-domain performance after task-specific fine-tuning. However, a substantial body of evidence shows that such models show performance degrading markedly when evaluated on data drawn from different topics, sources, or time periods \citep{janicka2019cross}. This limitation has motivated interest in alternative paradigms that reduce reliance on dataset-specific supervision.
The emergence of larger models has enabled a different evaluation setting based on in-context learning. In zero-shot and few-shot prompting, model parameters remain frozen and predictions are guided solely by natural-language instructions and, optionally, a small number of examples provided in the prompt. Rather than learning a task-specific decision boundary, this paradigm probes to what extent LLMs can function as classifiers by exploiting their internal knowledge. One of the earliest systematic studies in this direction, Hu et al. \cite{hu2023bad}, showed that GPT-3.5 can often identify fake news and generate multi-perspective rationales, but still lags behind fine-tuned BERT models in absolute accuracy.
Beyond pure prompting, other lines of work adapt LLMs more directly to the task through instruction tuning \cite{raza2025fake}. While these methods often close or surpass the performance gap with traditional classifiers in-domain, they reintroduce task-specific supervision and are therefore subject to the same generalization concerns observed in earlier fine-tuned models. A further family of approaches incorporates retrieval-augmented generation (RAG), conditioning LLM predictions on external evidence retrieved from the web or curated knowledge bases \citep{bai2024large,niu2024veract,thomas2025explainable}. Although retrieval can improve factual grounding, retrieval-augmented systems are highly sensitive to retrieval quality and coverage. Moreover, externally retrieved evidence cannot be systematically aligned with dataset annotations, which entangles retrieval errors with model behavior and complicates controlled cross-domain evaluation. In this work, we focus exclusively on full fine-tuning, zero-shot and few-shot prompting as a controlled evaluation paradigm.

\textbf{Multimodal and social context fake news detection.}
Recent research has increasingly moved towards integrative approaches. Multimodal detection models, such as EANN \cite{wang2018eann} and MMLO \cite{pramanick2021multimodal}, analyze the consistency between text and visual assets to identify manipulated content. Other approaches incorporate social-context signals, leveraging Graph Neural Networks (GNNs) to model propagation patterns and user interaction networks \cite{shu2019beyond,bian2020rumor,xu2022evidence}. 

\textbf{Comparative and benchmark studies.}
Despite the growing interest in generalization, most comparative studies continue to evaluate models primarily under in-domain conditions. Table~\ref{tab:empirical_studies} summarizes the main empirical works on fake news detection, reporting the number of evaluated models, datasets, and whether cross-domain analysis is performed.

Early benchmark studies focused almost exclusively on traditional machine learning methods and limited datasets. For instance, Ahmed et al.~\cite{ahmed2018detecting}, Agarwal et al.~\cite{agarwal2019analysis}, and Dutta et al.~\cite{dutta2019fake} evaluated a small number of classifiers on one or two datasets, without addressing domain shift. Subsequent works expanded the set of models to include deep learning architectures~\cite{jiang2021novel,katsaros2019machine,rohera2022taxonomy}, but still relied on single-domain or in-domain protocols.

More recent studies have incorporated Transformer-based models and, in some cases, LLMs. Alghamdi et al.~\cite{alghamdi2022comparative} compared a wide range of ML, DL, and Transformer approaches across multiple datasets, yet their evaluation remained strictly in-domain. Follow-up studies~\cite{chen2023using,alghamdi2023towards,farhangian2024fake} further examined DL and Transformer architectures, but continued to focus on single-dataset or in-domain settings. Notably, Farhangian et al.~\cite{farhangian2024fake} explicitly identified cross-dataset evaluation as a key challenge, while leaving it outside the scope of their analysis.

The most recent work by Huang et al.~\cite{huang2025unmasking} examines the role of LLMs and includes a cross-domain analysis. However, their study is limited to LLM-based approaches and does not include a direct comparison with traditional ML and DL methods or cross-domain architectures.

\textbf{Positioning of this study.}
While the existing literature reports model performance under specific experimental conditions, none adopts a single evaluation protocol that simultaneously spans diverse model families, multiple datasets, and systematically defined generalization scenarios. In contrast, this work evaluates 12 different approaches across 10 datasets under four experimental paradigms, namely \emph{dataset-specific, cross-dataset, mixed-training, and leave-one-dataset-out}, which are aligned with realistic deployment settings. In this context, we define generalization as the model's aggregate resilience to the naturally bundled distribution shifts, encompassing topics, genres, and collection protocols, encountered when moving between heterogeneous datasets. It is important to clarify that our study measures cross-dataset transferability under label and collection heterogeneity rather than establishing a causal link to isolated factors; therefore, it establishes the practical robustness of a model across divergent data sources without claiming to isolate purely semantic or topical invariance. In doing so, we address the generalization challenge identified by Janicka et al.~\cite{janicka2019cross} and report results that are currently absent from the literature.

\begin{table}[H]
\caption{Existing studies on fake news detection that include an experimental analysis. We report the papers and the number of tested ML models (ML), DL models (DL), (Large) Language Models (LLMs), the tested Datasets and the presence of Cross-Domain experiments and methods.}\label{tab:empirical_studies}
\centering
\scriptsize
\begin{tabularx}{\textwidth}{@{}p{3.5cm}*{4}{>{\centering\arraybackslash}p{1.3cm}}>{\centering\arraybackslash}p{3cm}p{4cm}@{}}
\toprule 
\textbf{Reference} & \textbf{ML} & \textbf{DL} & \textbf{LLMs} & \textbf{Datasets} & \textbf{Cross Domain} \\
\midrule 

Ahmed et al., 2018 \cite{ahmed2018detecting} & 6 & x & x & 2 & x\\ 
Agarwal et al., 2019 \cite{agarwal2019analysis} & 5 & x & x & 1 & x\\ 
Dutta et al., 2019 \cite{dutta2019fake} & 2 & x & x & 1 & x\\
Janicka et al., 2019 \cite{janicka2019cross} & 5 & x & x & 4 & \checkmark\\
Katsaros et al., 2019 \cite{katsaros2019machine} & 6 & 2 & x & 3 & x\\
Jiang et al., 2021 \cite{jiang2021novel} & 5 & 3 & x & 2 & x\\
Rohera et al., 2022 \cite{rohera2022taxonomy} & 3 & 1 &x & 1 & x\\
Alghamdi et al., 2022 \cite{alghamdi2022comparative} & 7 & 8 & 2 & 4 & x\\
Chen et al., 2023 \cite{chen2023using} & x & 3 & x & 1 & x\\
Alghamdi et al., 2023 \cite{alghamdi2023towards} & 5 & 5 & 10 & 1 & x\\
Farhangian et al., 2024 \cite{farhangian2024fake} & 8 & 3 & 9 & 4 & x\\

Huang et al., 2025 \cite{huang2025unmasking} & x & x & 4 & 4 & \checkmark \\

\textbf{Our paper} & \textbf{3} & \textbf{4} & \textbf{5} & \textbf{10} & \checkmark\\
\bottomrule 
\end{tabularx}
\end{table}

\section{Experimental Setup} \label{sec:experimentalSetup} 
Many current approaches to fake news detection frame the task as a classification problem, relying primarily on supervised learning to distinguish between individual real and fake news. In most cases, the broader context in which misinformation circulates, and the sources from which it originates, are not taken into account.  
Although several efforts have been made to construct extensive datasets, the available resources remain limited in scope and diversity, posing significant challenges for supervised approaches.

In our experiments, we used 10 of the most representative datasets available in the literature. 
Further, we evaluated 12 methods spanning the main families of approaches, from classical ML to Transformer-based architectures, up to cross-domain specific architectures. All selected approaches were chosen based on their prevalence and established effectiveness within the fake news detection literature.  

We designed four groups of experiments both to assess the ability of the selected methods to solve the task on different datasets, and to evaluate their generalization capability on unseen data.

\begin{description}

\item[Dataset-Specific] This group of experiments reflect the standard approach, where the goal is, for each $\langle algorithm, dataset \rangle$ pair, to find the optimal parameters for the algorithm to fit the dataset. This allows us to assess the performance of each classifier on each specific dataset. 

\item[Cross-Dataset] This group aims to evaluate the generalization capability of each classifier trained on a single dataset/domain. Each classifier is trained on a specific dataset using the optimal parameters identified in the Dataset-Specific experiments, and is then tested on all the remaining datasets. In this group, we use all available data from a dataset for training, regardless of the original splits. Natively multi-domain approaches are excluded from this evaluation. For decoder-only LLMs, in zero-shot settings we only evaluate the model on the test sets, without any conditioning beyond the prompt chosen via the Dataset-Specific experiments. In few-shot settings, we randomly sample examples from the training dataset for in-context learning, and test on the remaining ones. Note that the same in-context examples are used for all models. 

\item[Mixed-Training/Single-Test] With this group of experiments, we attempt to simulate a multi-domain scenario. Each classifier is trained on a \textit{global training set} composed of portions of data sampled from all the datasets. We evaluate the trained model in two ways: first, independently on each dataset, considering as test set only the examples not used for the training; second, we test on a \textit{global test set}, built by aggregating equal portions of data from each of the datasets. Here, we also include a model from the literature specifically designed for multi-domain and cross-domain. For decoder-only models, in the zero-shot setting, the training phase is bypassed entirely and the models are evaluated directly on the test sets. In the few-shot setting, we randomly sample examples from the training dataset, and evaluate the models on the remaining ones. Note that, as in the previous group, the in-context examples are identical across all models.

\item[Leave-One-Dataset-Out] Finally, this group of experiments allows us to simulate real-world scenarios in which a single model is used to classify cross-domain news. Each classifier is trained, in turn, on a mixture of all of the datasets except one, using the optimal parameters determined in the first group of experiments. The classifiers are then tested on the dataset/domain not seen during training. Here as well we include a model from the literature specifically designed for multi-domain and cross-domain settings. For decoder-only models, in the zero-shot setting, the training phase is bypassed and the models are evaluated directly on the test sets. In the few-shot setting, the in-context examples were randomly selected from the 9-dataset training set. The set of examples was kept consistent across all models. 

\end{description}

All the experiments assume a binary categorization of data into either \textit{fake} or \textit{real} news.

\subsection{The Used Datasets}\label{sect:dataset}

For our experiments, we used only publicly available datasets in English. We categorise these datasets into two families:  \textit{generic} datasets which contain data concerning broad topics such as politics or mixed themes, and \textit{narrow} datasets which contain data pertaining to a specific topic or real world event (e.g., \cite{passaro2022context}). Table  \ref{tab:dataset_selected} provides an overview of the datasets used.

All instances used in our experiments comprise: i) a textual entry representing an article, a social media post, or a short statement, and ii) a binary label indicating wether the content is  \textit{fake} or \textit{real}.  
Most of the datasets considered were originally annotated with binary labels. For the few datasets whose annotations did not directly align with our binary classification setup, we applied minimal preprocessing steps to derive a consistent binary label. These transformations were designed to preserve the original intent of annotations while ensuring cross-dataset comparability. A detailed description of the preprocessing applied to each dataset is provided in \ref{sec:appendix_dataset}.
We also note that, while the labeling rationale is mostly consistent across datasets, some of them also consider satirical texts as real (e.g., FakeVsSatire, Horne), whereas others restrict the real to factual reporting. In this work we consider the unique characteristics of each dataset as a domain when we refer to cross-domain generalization. We acknowledge that the differences in label semantics across datasets can partially affect results. To address this, we also conduct a replication of one of the analysis excluding dataset with satire in  \ref{sect:lodo_satire_analysis}. Results show that, despite some fluctuations, our overarching conclusions hold.

\begin{table}[H]
\centering\tiny
\caption{Description of all datasets used in the experiments 
($\dagger$ generic datasets; $\ddagger$ narrow datasets). 
For each dataset we report topic scope, source types, annotation criteria, 
time span (when available), binary mapping rule used in the experiments, and number of Fake and Real instances.}
\label{tab:dataset_selected}

\begin{tabular}{@{}l l p{1.4cm} p{1.4cm} p{1cm} p{2cm} p{1.3cm}@{}}
\toprule
\textbf{Dataset} & 
\textbf{Topic Scope} & 
\textbf{Source Types} & 
\textbf{Ann. Criteria} & 
\textbf{Time Span} &
\textbf{Mapping} & 
\textbf{Fake/Real} \\
\midrule

Celebrity \citep{perez2017automatic} &
Vips$\dagger$ &
Articles &
Fact-checked fake/real gossip & 2016--2017 &
Fake = fake;\newline Real = real & 250 / 250\\

Cidii \citep{hamed2023disinformation} &
Islamic Issues$\ddagger$ &
Articles;\newline Social Media &
Manual annotation of correct information vs.\ disinformation &
2016--2021 &
Fake = fake;\newline Real = real & 300 / 422\\

FakeVsSatire \citep{golbeck2018fake} &
Politics$\dagger$ &
Articles &
Manual annotation of fake news vs.\ satire &
2016--2017 &
Fake = fake;\newline Real = satire & 283 / 203 \\

Fakes \citep{salem2019fa} &
War in Syria$\ddagger$ &
Articles &
Semi-automated fact-checking of fake/real news &
2011--2018 &
Fake = fake;\newline Real = real & 378 / 426 \\

Horne \citep{horne2017just} &
Politics$\dagger$ &
Articles &
Manual annotation of real, fake, and satirical news &
2015--2016 &
Fake = fake;\newline Real = real or satire & 123 / 203\\

Infodemic \citep{patwa2021fighting} &
Covid-19$\ddagger$ &
Social Media &
Fact-checked COVID-19 misinformation &
2020 &
Fake = fake;\newline Real = real & 5031 / 5526\\

Isot \citep{ahmed2017detection} &
Multiple$\dagger$ &
Articles &
Manual annotation of fake/real news &
2016--2017 &
Fake = fake;\newline Real = real & 22855 / 21416 \\

LIAR-PLUS \citep{alhindi2018your} &
Politics$\dagger$ &
Short statements &
Six-class fact-checking taxonomy &
2007--2016 &
Fake = \{pants-on-fire, false, barely-true\};\newline 
Real = \{half-true, mostly-true, true\} & 5654 / 7130 \\

NDF \citep{passaro2022context} &
Notre Dame Fire$\ddagger$ &
Articles;\newline Social Media &
Fact-checked fake/real news &
2019 &
Fake = fake;\newline Real = real & 216 / 338 \\ \\

Politifact \citep{shu2017fake} &
Politics$\dagger$ &
Articles &
Six-class fact-checking taxonomy &
2007--2016 &
Fake = \{pants-on-fire, false, mostly-false\};\newline 
Real = \{half-true, mostly-true, true\} & 183 / 321\\
\bottomrule
\end{tabular}
\end{table}

All the datasets that were not already available with training, validation, and test sets were split using a stratified approach, following the rationale used for the Infodemic dataset (60\% training, 20\% validation, 20\% test). 
We used the same training, validation, and test sets for all comparison approaches.

For the generalization assessment, we adopt a cross-domain perspective where each dataset is treated as a distinct domain,  regardless of its generic or narrow categorization. The rationale is that each dataset, even those on the same topic, embodies unique biases arising from its specific collection sources, time frames, and labeling methods. Models can exploit these dataset-specific biases (e.g., lexical or stylistic artifacts) as shortcuts \cite{hoy2022exploring}. Consequently, cross-dataset evaluation in the second experiment (training on one dataset, testing on another dataset) serves to measure the generalization capability of each model trained on a single dataset.

Specifically for the third group of experiments we create a new training set, named \textit{global training set}, comprising instances extracted from all the datasets. For each dataset, we randomly selected the same number of instances for the Fake and Real classes. This number was set to 100 and was imposed by the Horne dataset size, the smallest dataset. Thus, the global training set is a balanced set comprising 2000 instances, divided into 1000 fake and 1000 real. As for the test set for this group of experiments, we exploited each dataset, considering only the portion not used for the training phase. Table \ref{tab:dataset_reduced} reports the number of instances for the two classes used in the test phase.

\begin{table}[H]
 \centering\scriptsize
\caption{Distribution of the instances in the subsets of the datasets used in the test phase of the third group of experiments.}\label{tab:dataset_reduced}
 
\begin{tabular}{@{}lcccc@{}}
\toprule
 \textbf{Dataset} & \textbf{Fake class} & \textbf{Real class}\\
\midrule 

Celebrity \citep{perez2017automatic} &  150 & 150\\
Cidii \citep{hamed2023disinformation}& 200 & 322\\
FakeVsSatire \citep{golbeck2018fake}&  183 & 103\\
Fakes \citep{salem2019fa} &  278 & 326\\
Horne \citep{horne2017just}&  23 & 103\\
Infodemic \citep{patwa2021fighting}&  4933 & 5426\\
Isot \citep{ahmed2017detection} &  22755 & 21316\\
LIAR-PLUS \citep{alhindi2018your}&  5556 & 7030\\
NDF \citep{passaro2022context}&  116 & 238\\
Politifact \citep{shu2017fake} &  83 & 221\\

\bottomrule 
\end{tabular}
 
\end{table}

We also created another test set, denoted as \textit{global test set}, by taking equal portions of instances (20 for the fake class and 20 for the real class) from all instances of each dataset not used in the training phase.

The setup of the Leave-One-Dataset-Out experiment allows us to measures model's true ability to generalize to unseen domains by assessing its capacity to overcome dataset-specific biases \cite{hoy2022exploring}.
It simulates the deployment of a general-purpose detector encountering a new topic for the first time, thereby providing a clear indication of a model's ability to learn universal signals of misinformation rather than domain-specific artifacts.

In constructing the training sets for the Leave-One-Dataset-Out experiment (see Section \ref{sect:exp4}), we aggregated all available instances from the $N-1$ training datasets without downsampling the larger ones (i.e., Isot, Infodemic and LIAR-PLUS) to match the smaller ones. This choice is motivated by the fact that in a realistic deployment scenario, a detection system would leverage the entirety of its historical database to maximize coverage, mirroring the natural imbalance in news volume across topics and events. Furthermore, specialized cross-domain architectures like MERMAID are highly data-hungry; providing the full volume of data is essential for enabling these models to converge and effectively learn the routing mechanisms without underfitting. 

To further support the research community and promote transparency, we provide a dedicated GitHub repository which includes the code for the models used and the references and pointers to the publicly available datasets.
\footnote{The repository will be made available upon acceptance of the paper.} 

\subsection{The Tested Approaches}\label{sect:approach} 
In this section, we provide a detailed description of all the approaches evaluated in our study, including their architectures, feature representations, training procedures, and implementation frameworks. This methodological overview ensures that each experimental result can be directly traced back to a clearly specified model configuration.

For our experiments we selected 12 methods which represent a mix of traditional ML classifiers and state-of-the-art DL techniques, including Transformer models. These methods were selected based on their prevalence, established effectiveness, and availability. We summarize the approaches in Table \ref{tab:approaches_selected}.

\begin{table}[H]\centering
 \scriptsize
\caption{Approaches used for the experimentation.}\label{tab:approaches_selected}

\begin{tabular}{@{}lll@{}}
\toprule
 \textbf{Approach} & \textbf{Type} & \textbf{Features/Method}\\
\midrule 
LR \cite{jiang2021novel} & Traditional ML & Tf-Idf (Bow) \\
SVM \cite{rohera2022taxonomy} & Traditional ML & Tf-Idf (Bow)  \\
NB \cite{rohera2022taxonomy} & Traditional ML & Tf-Idf (Bow) \\
CNN \cite{wang2017liar} & DL (Training from scratch) & Word2Vec (Embeddings) \\
BiLSTM \cite{chen2023using} & DL (Training from scratch) & Word2Vec (Embeddings) \\
CNN-BERT \citep{kaliyar2021fakebert} & DL (Training from scratch) & BERT (Embeddings) \\
BERT \citep{devlin2018bert} & Encoder-only Transformer & BERT (Fine tuninig) \\
DeBERTa \citep{he2020deberta} & Encoder-only Transformer & DeBERTa (Fine tuninig) \\
MERMAID \citep{liguori2025breaking} & Cross-domain architecture & DeBERTa (Embeddings), MoE \\
Llama3-8B \citep{touvron2023llama} & Decoder-only Transformer & Zero- and few-shot Prompting \\
Qwen3-32B \citep{qwen3} & Decoder-only Transformer & Zero- and few-shot Prompting \\
Zephyr-7B-beta \citep{tunstall2023zephyr} & Decoder-only Transformer & Zero- and few-shot Prompting \\

\bottomrule 
\end{tabular}
 
\end{table}

It is worth noting that the choices made in our experimental setup were guided by our main objective: to compare various methods in a rigorous and fair manner while also ensuring their suitability for the task of fake news detection. We have applied these choices consistently across all groups of experiments. 

\paragraph{Traditional ML Models} We exploited a Logistic Regression (LR) classifier inspired by \cite{jiang2021novel}, a Support Vector Machine classifier (SVM), and a Multinomial Naïve Bayes (NB), both inspired by \cite{rohera2022taxonomy}. We implemented the classifiers with \texttt{scikit-learn}.\footnote{\url{https://scikit-learn.org}} The feature representation method used for the ML classifiers is Term Frequency-Inverse Document Frequency (TF-IDF), which aligns with common practices in the literature \citep{ahmed2017detection,ahmed2018detecting,della2018automatic,jiang2021novel,mohsen2024automated}.

\paragraph{DL Models} We employed three different approaches: a standard CNN and a BiLSTM, inspired respectively by the works of \cite{wang2017liar} and \cite{chen2023using}, and a BERT-based deep convolutional approach inspired by FakeBERT \citep{kaliyar2021fakebert}, which we refer to as CNN-BERT.
For all implementations we used TensorFlow Keras.\footnote{\url{https://keras.io/}} 
For CNN and BiLSTM we used pre-trained 300-dimensional word2vec embeddings from Google News \citep{mikolov2013efficient} to warm-start the text embeddings. The CNN consists of an Embedding layer followed by a Dropout, a single 1D-convolutional layer with a Global Max Pooling, and a single dense hidden layer. The output layer uses a sigmoid activation function. The BiLSTM consists of an Embedding layer followed by two LSTMs, one that processes the sequence forward and one that processes it backward. The outputs of both directions are then concatenated and passed to a single dense hidden layer. The output layer exploits a sigmoid activation function. We used Adam\footnote{\url{https://keras.io/api/optimizers/adam/}} as optimizer for both the CNN and BiLSTM, as common in the literature \citep{bugueno2019empirical, bahad2019fake, liao2021integrated}. 
For CNN-BERT we reproduced  the implementation described in \citep{kaliyar2021fakebert}.
We acknowledge that this ``DL'' family of approaches encompasses a wide range of methods, differing significantly in architecture complexity and training/inference costs. 

\paragraph{Transformer-based Models} As for encoder-only models, we fine-tuned BERT \citep{devlin2018bert} (\texttt{bert-base-cased}) and DeBERTa \citep{he2020deberta} (\texttt{deberta-base}) using Huggingface.\footnote{\url{https://huggingface.co}}
We selected the cased version of BERT as capitalization can be a relevant signal for fake news detection \citep{affelt2019spot}. As a representative cross-domain architectures, we included MERMAID \citep{liguori2025breaking}, adopting the implementation and hyperparameters provided in the original paper.

For the decoder-only models, we used the HuggingFace implementation of LLama3-8B \citep{touvron2023llama}, Qwen3-32B \citep{qwen3} and Zephyr-7B-Beta \citep{tunstall2023zephyr}. We evaluated them in both zero-shot and few-shot settings. We conducted a prompt optimization process on a validation set. Full details of the prompts are available in \ref{sec:appendix_optimization}. To ensure deterministic predictions we set temperature to zero. For Qwen3-32B, we explicitly disabled the thought- generation mechanism, ensuring that only the final answer is produced. In the few-shot scenario, examples were randomly selected in a stratified manner from the training set of the experiment. For comparability, inference was performed on exactly the same test data used for the individual experiments of the various groups. Full inference scripts will be released in our GitHub repository after acceptance.

Finally, we note that the MERMAID approach is only used in the Mixed-Training/Single-Test and Leave-One-Dataset-Out groups of experiments. This limitation is due to the MERMAID's MoE architecture, which is specifically designed to leverage  knowledge from multiple specialized models through a trainable gating sub-network \citep{liguori2025breaking}. In single-domain settings, this gating mechanism would collapse into a trivial identity mapping, as the model requires a multi-domain environment to train the gating logic and differentiate between expert specializations. 

The training and inference times for each approach are presented and discussed in \ref{sec:appendix_training_time}.

\paragraph{Tuning Budget}
To ensure fairness across model families, we adopt a uniform tuning budget. All models are tuned once per dataset using the same train/validation/test split. Traditional ML models, DL models, and fine-tuned encoder-only Transformers are each optimized through a single grid search over the hyperparameter ranges reported in \ref{sect:appendix_opt_param}. For decoder-only LLMs, which do not involve parameter tuning, we instead evaluate a set of 3 prompt templates on the same validation split, as it is reported in \ref{sect:appendix_opt_prompt}. 

\paragraph{Practical Interpretation of Statistical Results} 
To bridge the gap between statistical significance and operational utility, we define criteria for interpreting our findings beyond null-hypothesis testing. While p-values indicate the reliability of the observed performance gaps, we focus on the magnitude of these differences to assess their practical impact on real-world fake news detection. In a deployment scenario, a ``meaningful difference'' is characterized by a consistent improvement in F1-score that suggests a higher resilience to domain shift. Specifically, we consider a model's performance gain to be practically significant if it demonstrates a clear separation in the Critical Difference (CD) diagrams, indicating that the improvement is robust across diverse datasets and not merely an artifact of a specific domain's characteristics.

\section{Results} \label{sec:experiments} 

In this section, we discuss the results obtained across the four groups of experiments. Due to space constraints, we focus on the key findings and use the F1-score as the primary metric to compare model performance across experiments. Detailed results, including precision and recall values, as well as the statistical tests used to assess differences among models within each experimental group, are provided in the Supplementary Material. Here, we summarize the main conclusions derived from these statistical analyses.

We adopt a performance-oriented evaluation framework, comparing models based on their best attainable F1-score rather than on performance relative to computational budget. Introducing compute-budgeted evaluations would risk conflating modeling and generalization capabilities with hardware- and implementation-specific factors across heterogeneous model families. Accordingly, computational efficiency is treated as an orthogonal consideration and lies outside the scope of the present work.

\subsection{Dataset-Specific Experiments}\label{sect:exp1} 

Table \ref{tab:all_f1_models_by_dataset} shows the F1-scores obtained in the Dataset-Specific experiments by the 12 different approaches.\footnote{See Section 1 of the Supplementary Material for detailed results.}

\begin{table}[H]
\scriptsize
\caption{F1-score obtained by all models on each dataset. The decoder-only models are considered both in zero-shot (ZS) and few-shot (FS) scenarios. The best F1-score per dataset is in bold.}
\label{tab:all_f1_models_by_dataset}
\begin{tabularx}{\textwidth}{@{}l*{10}{>{\centering\arraybackslash}X}@{}}
\toprule
\textbf{Model} &
\rotatebox{90}{\textbf{Celebrity}} &
\rotatebox{90}{\textbf{Cidii}} &
\rotatebox{90}{\textbf{FakeVsSatire}} &
\rotatebox{90}{\textbf{Fakes}} &
\rotatebox{90}{\textbf{Horne}} &
\rotatebox{90}{\textbf{Infodemic}} &
\rotatebox{90}{\textbf{Isot}} &
\rotatebox{90}{\textbf{LIAR-PLUS}} &
\rotatebox{90}{\textbf{NDF}} &
\rotatebox{90}{\textbf{Politifact}} \\
\midrule
LR                & .60         & .97         & .70         & .43         & .76         & .93         & .99          & .56         & .86         & .77         \\
SVM                 & .67         & .86         & .69         & .44         & .71         & .93         & .99          & .57         & .82         & .75         \\
NB                  & .47         & .86         & .68         & .42         & .68         & .88         & .96          & .56         & .79         & .71         \\ \midrule
CNN                 & .54         & .58         & .56         & .53         & .56         & .55         & .96          & .57         & .56         & .64         \\
BiLSTM              & .50         & .59         & .51         & \textbf{.56} & .62        & .70         & .97          & .57         & .63         & .65         \\
CNN-BERT            & .50         & .59         & .58         & .53         & .70         & .68         & .99          & .58         & .59         & .64         \\ \midrule
BERT                & .76         & \textbf{.98} & .72        & .44         & .77         & \textbf{.97} & \textbf{1.0} & .61         & .84         & .84         \\
DeBERTa             & \textbf{.82} & .97        & \textbf{.78} & .47        & \textbf{.95} & \textbf{.97} & \textbf{1.0} & .63         & .87         & \textbf{.86} \\ 
\midrule

LLaMa3-8B (ZS)     & .76         & .89         & .57         & .53         & .67         & .78         & .73          & .61         & .80         & .75         \\
LLaMa3-8B (FS)      & .75         & .97         & .56         & .50         & .67         & .84         & .88          & .60         & .85         & .75         \\
Qwen3-32B (ZS)      & .81         & .94         & .58         & .48         & .67         & .84         & .75          & .60         & .86         & .79         \\
Qwen3-32B (FS)      & .76         & .96         & .58         & .47         & .73         & .86         & .81          & .62         & \textbf{.92} & .75         \\
Zephyr-7B-beta (ZS) & .67         & .86         & .45         & .52         & .56         & .76         & .63          & .62         & .82         & .66         \\
Zephyr-7B-beta (FS) & .71         & .91         & .54         & .53         & .52         & .85         & .64          & \textbf{.65} & .85         & .54         \\
\bottomrule
\end{tabularx}
\end{table}

We observe that DeBERTa outperforms the other models, excelling on most datasets, independently of sizes and topics. The second best performing model is BERT. This confirms that fine-tuning large pre-trained models on fake news datasets are more suitable than ML models trained from scratch. This aligns with previous research \cite{alghamdi2022comparative}.

Traditional ML models perform surprisingly well, particularly on smaller datasets. DL models show more mixed results: they excel on some datasets but underperform on others, especially the smaller ones. In particular, CNN-BERT and CNN achieve comparable average performance, yet CNN-BERT outperforms CNN on larger datasets such as Infodemic and Isot, which is expected given its more complex architecture. A critical issue encountered with these models is their tendency to overfit the training data relatively quickly. This can be observed in the limited number of epochs required before the loss stopped decreasing. This behavior suggests that these models might be highly sensitive to the specific characteristics of the training data, hindering their ability to generalize to unseen examples. This pattern is even more evident when the models are trained on one dataset and tested on different datasets (see Section \ref{sect:exp2}). Overall, it seems that the parameter search performed has not been sufficient for these DL models, which seem to require more extensive training data and architecture adjustments to achieve optimal performance. In the original paper, CNN-BERT was tested on a large dataset downloaded from Kaggle that we could not identify, 
obtaining an F1-score of 0.98. The result, however, is comparable to the one obtained by CNN-BERT on Isot.

It is worth noting that neural models are known to exhibit variability across different training runs due to random initialization and stochastic optimization. To ensure that our conclusions are robust, we report mean performance and standard deviation over multiple runs in \ref{sect:appendix_seed_neural}.

The results of the decoder-only LLMs offer a useful alternative perspective. Even in a zero-shot setting, models such as Qwen3-32B and Llama3-8B demonstrate good capabilities. Notably, Qwen3-32B achieves highly competitive F1-scores of 0.94 on Cidii and 0.86 on NDF, outperforming traditional baselines and nearing the performance of fine-tuned BERT on NDF without requiring weight updates.

The transition from zero-shot to few-shot prompting yields a substantial performance leap on some datasets, confirming the models' ability to learn from limited context. We observe significant gains, such as Llama3-8B improving from 0.73 to 0.88 on Isot and from 0.89 to 0.97 on Cidii. Notably, on the NDF dataset, Qwen3-32B in the few-shot setting achieves an F1-score of 0.92, surpassing both the fine-tuned BERT (0.84) and DeBERTa (0.87). This suggests that while DeBERTa remains the most consistent model across all domains, few-shot LLMs can achieve state-of-the-art performance on specific tasks without the need for full model training.

\begin{figure}[!htb]
\centering
\includegraphics[width=0.95\linewidth]{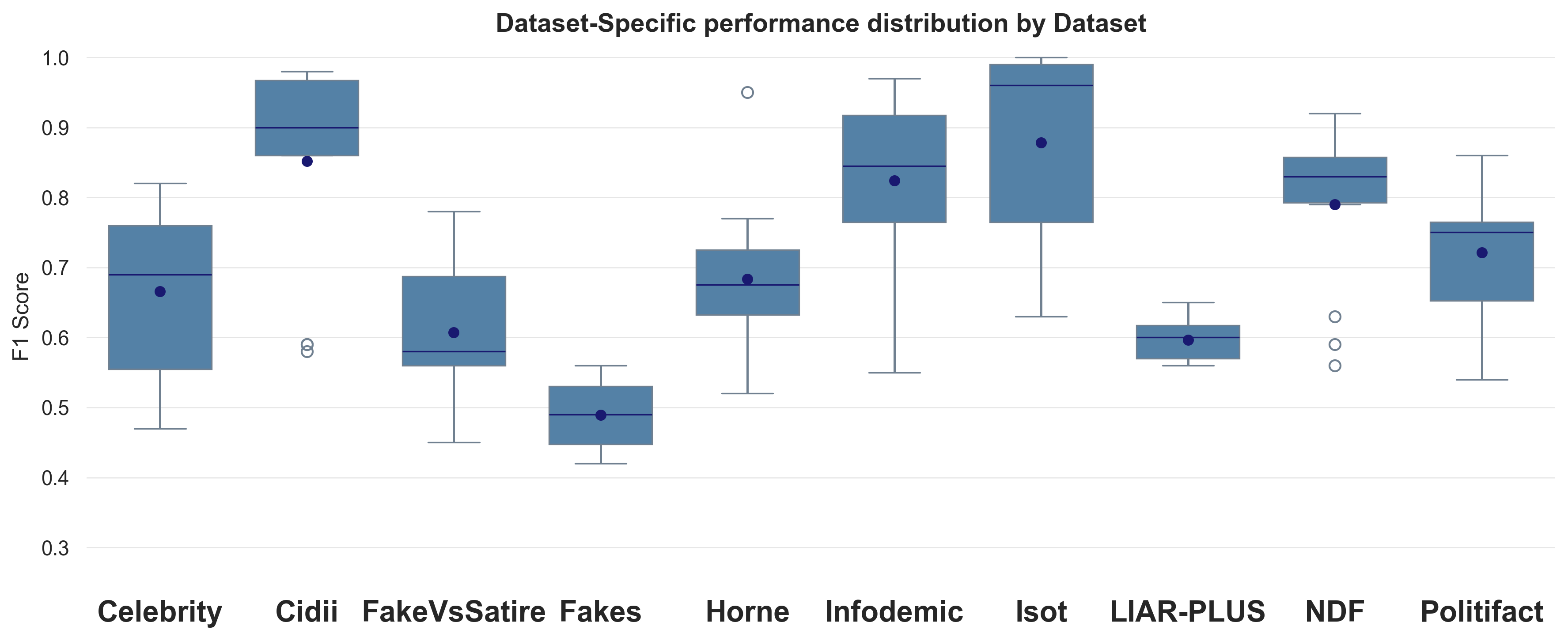} 
\caption{Distribution of F1-scores across the datasets for the Dataset-Specific experiments.}
\label{fig:distr-dataset-specific} 
\end{figure}

Figure \ref{fig:distr-dataset-specific} shows the F1-score distribution across datasets. We observe substantial variation in the average performance achieved on the different datasets. However, when compared with prior literature on these datasets, our average performances generally align with previously reported results. The Fakes and LIAR-PLUS datasets appear to be challenging for all the models when compared with datasets such as Politifact,
as also shown in \cite{mohsen2024automated} and \cite{alghamdi2024power}. The complexity of the Fakes dataset is likely attributable to the fact that it describes a multifaceted real-world event, making the classification task inherently more difficult.
In contrast, Isot is known to be relatively simple (see e.g., \cite{hakak2021ensemble}), a trend that our results also confirm.
The performances on the FakeVsSatire dataset are also interesting. The dataset is challenging, and even BERT and DeBERTa struggle to achieve high F1-scores. In other datasets, such as Infodemic and Cidii, the models exhibit more variable results.
The results obtained in the Politifact dataset, which is widely used in the literature, are consistent with previous findings \cite{alghamdi2022comparative,alghamdi2024enhancing,alghamdi2024power}.
To further address the practical significance of the model's performances Table \ref{tab:effect_sizes_se} reports the mean effect sizes ($\Delta$F1) and their corresponding Standard Errors (SE), using DeBERTa as the reference best-performing baseline. 

\begin{table}[ht]
\centering
\scriptsize
\caption{Mean effect size ($\Delta$F1) and Standard Error (SE) of all models compared to DeBERTa (best baseline). Calculations are performed across the 10 datasets.}
\label{tab:effect_sizes_se}
\begin{tabular}{lcc}
\toprule
\textbf{Model} & \textbf{Mean $\Delta$F1} & \textbf{Standard Error (SE)} \\ \midrule
LR & -0.075 & 0.022 \\
SVM & -0.089 &  0.020 \\
NB & -0.131 &  0.030 \\ \midrule
CNN & -0.227 & 0.049 \\
BiLSTM & -0.202 & 0.045 \\
CNN-BERT & -0.194 &  0.043 \\ \midrule
BERT & -0.039 & 0.016 \\ \midrule
LLaMa3-8B (ZS) & -0.123 & 0.033 \\
LLaMa3-8B (FS) & -0.095 & 0.029 \\
Qwen3-32B (ZS) & -0.100 & 0.032 \\
Qwen3-32B (FS) & -0.086 &  0.028 \\
Zephyr-7B-beta (ZS) & -0.177 & 0.045 \\
Zephyr-7B-beta (FS) & -0.158 & 0.050 \\ 
\bottomrule
\end{tabular}
\end{table}

The analysis confirms that DeBERTa consistently outperforms traditional ML and DL models. Notably, it maintains a stable advantage even over the best-performing LLM (i.e., Qwen3-32B). The low SE values across all categories further demonstrate that these gains are consistent across diverse domains.

To better understand the behavior of DeBERTa as the best-performing model and to assess the nature of its mistakes, we show a dedicated qualitative error analysis in \ref{sec:error_analysis}.

\begin{figure}[!htb]
\centering
\includegraphics[width=0.95\linewidth]{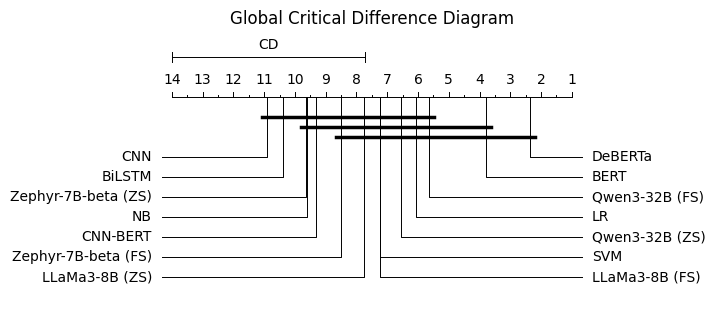} 
\caption{Critical-difference diagram showing the statistically significant differences among models according to the Nemenyi post-hoc test.}
\label{fig:critical_difference_diagram_exp1} 
\end{figure}

Finally, to assess whether statistically significant differences exist among the models, we first computed the Friedman test and then applied the post-hoc Nemenyi test. The critical-difference diagram in Figure \ref{fig:critical_difference_diagram_exp1} summarizes the statistical comparisons across all models.\footnote{See Section 1 of the Supplementary Material for the results of the statistical tests.} The results indicate that, overall, the models do not differ significantly from one another, with two exceptions: BERT, which exhibits statistically significant differences when compared with CNN and BiLSTM, and DeBERTa, which shows statistically significant differences when compared with NB, CNN, BiLSTM, CNN-BERT, and Zephyr-7B-beta.

Taken together, dataset-specific performance should not be interpreted as evidence of general robustness, as strong in-domain results remain poor predictors of cross-domain generalization.

\subsection{Cross-Dataset Experiments}\label{sect:exp2}

The main results of the Cross-Dataset experiments are summarized in Figure \ref{fig:distr-cross-dataset}.\footnote{See Section 2 of the Supplementary Material for detailed results.} For each model, we show the box-plot representing the distribution of the F1-scores (90 for all the models except for decoder-only LLMs used in the zero-shot setting) obtained by training on one dataset and testing on all the others. 

\begin{figure}[!htb]
\centering
\includegraphics[width=0.95\linewidth]{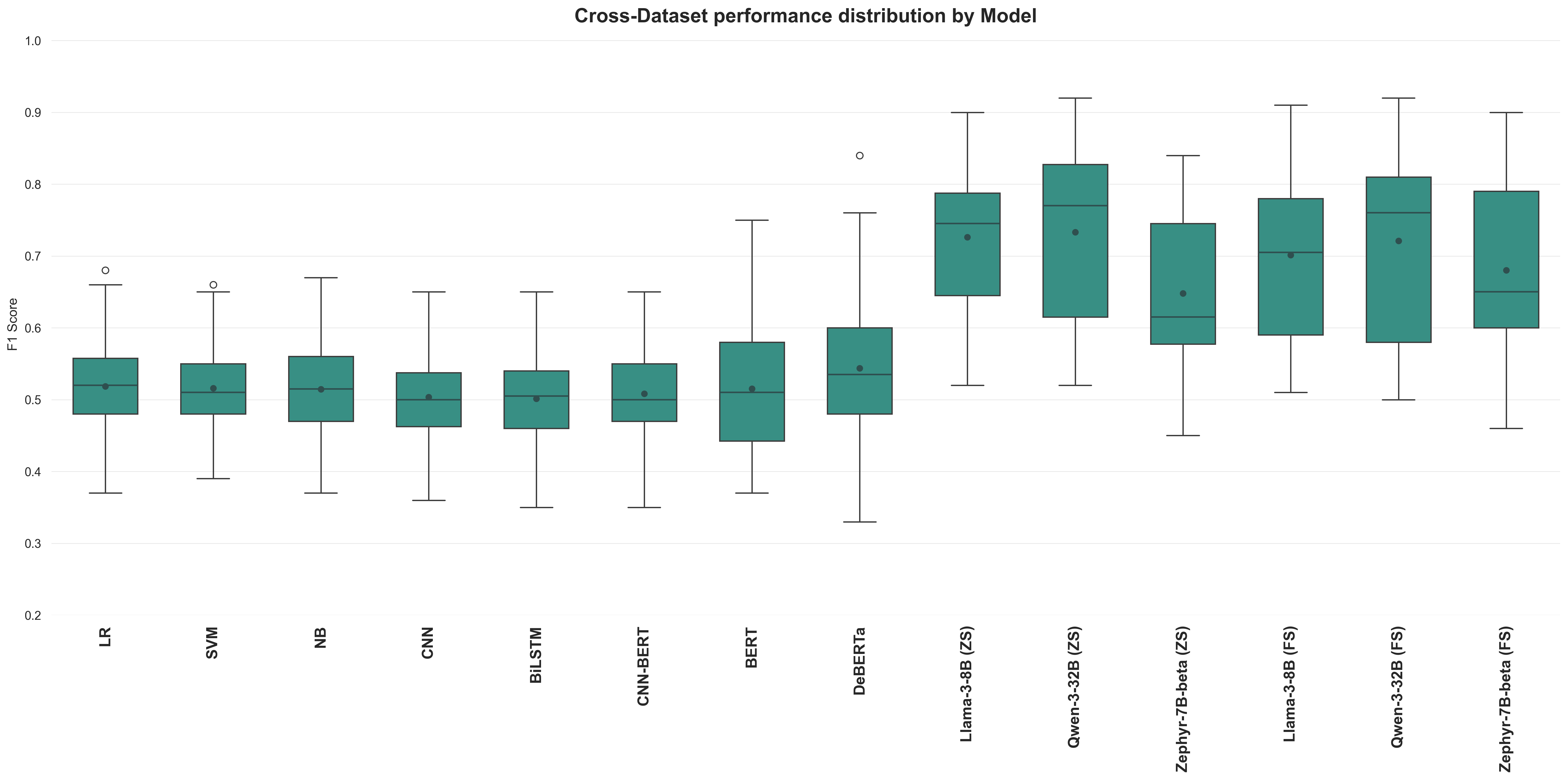} 
\caption{Distribution of the average F1-scores across the datasets for the Cross-Dataset  experiments.}
\label{fig:distr-cross-dataset} 
\end{figure}

Traditional ML models and standard DL architectures trained from scratch exhibit limited generalization, as expected, with median F1-scores around $0.50$ and relatively compact interquartile ranges. This suggests that these models fail to capture domain-invariant features, performing only marginally better than random guessing on unseen domains. Fine-tuned Transformers exhibit a wider performance spread, indicating that they may generalize to domains that are semantically close to the training data, but fail when the semantic gap between source and target domains is too large.

Decoder-only models perform markedly better. In the zero-shot setting they obtain results comparable to the Dataset-Specific experiments, as they are evaluated on a subset of data used for that experiment. The few-shot scenario poses a different challenge, since examples drawn from one domain are used as in-context demonstrations to predict another. On average, performances are similar to or worse than the zero-shot condition, suggesting that given the models' understanding of the task, out-of-domain examples may be even counterproductive. 

A more fine-grained analysis of the results
highlights that some datasets, such as Isot, contain strong, dataset-specific artifacts. Models trained on such data often fail to generalize in a meaningful way.

\begin{figure}[!htb]
\centering
\includegraphics[width=0.95\linewidth]{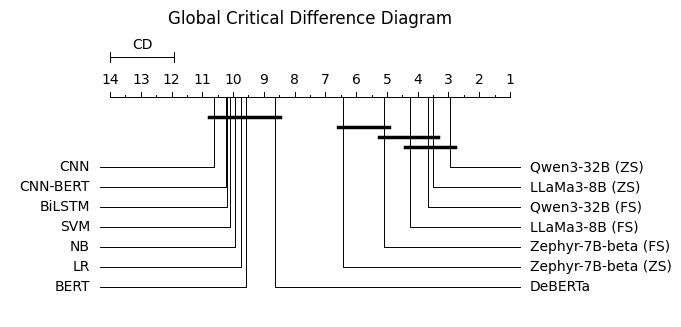} 
\caption{Critical-difference diagram showing the statistically significant differences among models according to the Nemenyi post-hoc test.}
\label{fig:critical_difference_diagram_exp2} 
\end{figure}

To assess whether statistically significant differences exist among the models, we first computed the Friedman test and then applied the post-hoc Nemenyi test. The critical-difference diagram in Figure \ref{fig:critical_difference_diagram_exp2} summarizes the statistical comparisons across all models.\footnote{See Section 2 of the Supplementary Material for the results of the statistical tests.} Results show that the decoder-only LLMs are statistically similar to each other, yet different from all other fully trained models, which in turn appear statistically similar among themselves. This suggests that fully trained models are less resilient to domain shifts. The only exception is Zephyr-7B-beta, which is also statistically different from other LLMs as well.

To further explore generalization limitations, specifically in fully trained models, we performed an additional analysis. Recall that we distinguish between \textit{generic} datasets, that include open-domain data, and \textit{narrow} datasets, that include a specific time-span or topic. We, therefore, explore how the type of training input (\textit{generic} or \textit{narrow}) affects generalization performance.

We report per-dataset results in Table \ref{tab:generic_narrow}.

\begin{table}[]

\centering
\scriptsize
\caption{Average F1-scores achieved from the models on the generic$\dagger$ (i.e Horne, Isot, FakeVsSatire, Politifact, LIAR-PLUS, Celebrity) and narrow$\ddagger$ (i.e. Cidii, NDF, Fakes and Infodemic) datasets, when trained on the dataset in the column. Best F1-scores are highlighted in bold for all models except decoder-only LLMs, in order to emphasize approaches involving training or fine-tuning on the target data.}\label{tab:generic_narrow}
\begin{tabularx}{\textwidth}{@{}lc*{10}{>{\centering\arraybackslash}X}@{}}
\toprule
\textbf{Model} & \textbf{Test} & \rotatebox{90}{\textbf{Celebrity$\dagger$}} & \rotatebox{90}{\textbf{Cidii$\ddagger$}} & \rotatebox{90}{\textbf{FakeVsSatire$\dagger$}} & \rotatebox{90}{\textbf{Fakes$\ddagger$}} & \rotatebox{90}{\textbf{Horne$\dagger$}} & \rotatebox{90}{\textbf{Infodemic$\ddagger$}} & \rotatebox{90}{\textbf{Isot$\dagger$}} & \rotatebox{90}{\textbf{LIAR$\dagger$}} & \rotatebox{90}{\textbf{NDF$\ddagger$}} & \rotatebox{90}{\textbf{Politifact$\dagger$}} \\
\midrule
\multirow{2}{*}{LR}
& Generic & .55 & .45 & .45 & .51 & .54 & \textbf{.50} & .49 & .55 & .50 & .52 \\
& Narrow & .53 & .47 & \textbf{.50} & .53 & .49 & .47 & .48 & .52 & .47 & \textbf{.55} \\
\multirow{2}{*}{SVM}
& Generic & .55 & .48 & \textbf{.52} & .49 & .55 & .48 & .50 & .54 & .50 & .56 \\
& Narrow & .51 & .47 & .49 & .50 & .49 & \textbf{.50} & .49 & .50 & .46 & .51 \\
\multirow{2}{*}{NB}
& Generic & .53 & .47 & .48 & .52 & .52 & .48 & \textbf{.52} & .55 & .48 & .51 \\
& Narrow & .53 & .50 & \textbf{.50} & .51 & .50 & .47 & \textbf{.50} & .51 & .47 & .52 \\
\midrule
\multirow{2}{*}{CNN}
& Generic & .48 & .47 & .45 & .52 & .50 & .47 & .49 & .53 & .52 & .50 \\
& Narrow & .47 & .44 & .49 & .49 & .50 & \textbf{.50} & .49 & .52 & .48 & .51 \\
\multirow{2}{*}{BiLSTM}
& Generic & .45 & .49 & .47 & .52 & .51 & \textbf{.50} & .48 & .53 & .49 & .50 \\
& Narrow & .48 & .49 & .49 & .43 & .52 & .48 & .48 & .51 & .48 & .50 \\
\multirow{2}{*}{CNN-BERT}
& Generic & .52 & .46 & .43 & \textbf{.53} & .51 & .49 & .44 & .52 & .49 & .51 \\
& Narrow & .52 & .44 & .43 & .51 & .51 & .46 & .44 & .52 & .45 & .51 \\
\midrule
\multirow{2}{*}{BERT}
& Generic & .63 & \textbf{.53} & .49 & \textbf{.53} & \textbf{.60} & .46 & .48 & \textbf{.62} & .50 & .53 \\
& Narrow & \textbf{.54} & \textbf{.55} & .49 & .51 & .39 & .49 & .48 & .54 & .51 & .52 \\
\multirow{2}{*}{DeBERTa}
& Generic & \textbf{.68} & .48 & \textbf{.52} & \textbf{.53} & .59 & .47 & .47 & .57 & \textbf{.64} & \textbf{.63} \\
& Narrow & \textbf{.54} & .45 & .43 & \textbf{.57} & \textbf{.55} & .43 & .46 & \textbf{.57} & \textbf{.61} & .53 \\\midrule
\multirow{2}{*}{Llama (FS)}
& Generic & .83 & .88 & .58 & .52 & .71 & .81 & .70 & .59 & .72 & .68 \\
& Narrow & .81 & .89 & .57 & .53 & .70 & .80 & .79 & .59 & .67 & .71 \\
\multirow{2}{*}{Qwen (FS)}
& Generic & .81 & .87 & .58 & .52 & .73 & .82 & .74 & .57 & .83 & .77 \\
& Narrow & .77 & .84 & .58 & .53 & .73 & .80 & .73 & .56 & .82 & .76 \\
\multirow{2}{*}{Zephyr (FS)}
& Generic & .77 & .84 & .51 & .52 & .62 & .83 & .70 & .61 & .79 & .60 \\
& Narrow & .77 & .88 & .48 & .53 & .63 & .80 & .64 & .62 & .80 & .65 \\
\bottomrule
\end{tabularx}

\end{table}

Fine-tuned Transformers achieve the best results across almost all scenarios. We observe that training on the same kind of data (Generic$\rightarrow$Generic and Narrow$\rightarrow$Narrow) yields the best and most consistent results regardless of the model. We also observe that fine-tuned Transformers are slightly better at leveraging \textit{generic} knowledge when applied to \textit{narrow} data; the reverse, however, does not hold. In fact, all models perform poorly when trained on \textit{narrow} data and tested on \textit{generic} data.

Focusing on DeBERTa, we observe notable improvements when trained on NDF, Politifact, and Celebrity. DeBERTa generally performs better on \textit{generic} test sets than on \textit{narrow} ones, regardless of the training data source. The notable exception is when the model is trained on \textit{Fakes}: in this case, the model aligns with the domain specificity, performing better on \textit{narrow} test sets than on \textit{generic} ones.

We note that the LLMs operate in a few-shot setting and are therefore not directly comparable to the fully trained models discussed above. Their consistently high scores reflect their prior knowledge rather than adaptation to the training dataset, and their cross-dataset behaviour is consequently less informative for assessing generalization under domain shift. 

We further introduce a second controlled analysis based on genre. Specifically, we group datasets into two broad genres, Articles (datasets composed exclusively of news articles) and Others (datasets consisting of social media posts, short claims, or mixtures of short-form and long-form content). We evaluate models when training and testing within the same genre as well as across genres, as reported in Table \ref{tab:articles_others}. 

\begin{table}[]
\centering
\scriptsize
\caption{Average F1-scores achieved from the models on the datasets grouped by genre: Articles (i.e Celebrity, FakeVsSatire, Fakes, Horne, Isot, Politifact) and Others (i.e. Cidii, Infodemic, LIAR-PLUS and NDF), when trained on the dataset indicated by each column. Best F1-scores are highlighted in bold, except for LLMs, to emphasize models trained from scratch or fine tuned on the target data.}\label{tab:articles_others}
\begin{tabularx}{\textwidth}{@{}lc*{10}{>{\centering\arraybackslash}X}@{}}
\toprule
\textbf{Model} & \textbf{Test} & \rotatebox{90}{\textbf{Celebrity$\dagger$}} & \rotatebox{90}{\textbf{Cidii$\ddagger$}} & \rotatebox{90}{\textbf{FakeVsSatire$\dagger$}} & \rotatebox{90}{\textbf{Fakes$\ddagger$}} & \rotatebox{90}{\textbf{Horne$\dagger$}} & \rotatebox{90}{\textbf{Infodemic$\ddagger$}} & \rotatebox{90}{\textbf{Isot$\dagger$}} & \rotatebox{90}{\textbf{LIAR$\dagger$}} & \rotatebox{90}{\textbf{NDF$\ddagger$}} & \rotatebox{90}{\textbf{Politifact$\dagger$}} \\
\midrule
\multirow{2}{*}{LR}
& Articles & .56 & \textbf{.56} & .46 & .51 & .53 & .50 & .51 & .54 & .49 & .52 \\
& Others   & .54 & .51 & .49 & .55 & .56 & .44 & .46 & .56 & .51 & .57 \\
\multirow{2}{*}{SVM}
& Articles & .56 & .53 & \textbf{.52} & .49 & .55 & .49 & .51 & .54 & .50 & .55 \\
& Others   & .55 & .51 & .49 & .49 & .55 & .45 & .45 & .55 & .47 & .54 \\
\multirow{2}{*}{NB}
& Articles & .53 & .53 & .48 & .52 & .52 & .48 & \textbf{.53} & .55 & .48 & .51 \\
& Others   & .56 & .49 & .50 & .54 & .56 & .44 & .48 & .47 & .50 & .57 \\
\midrule
\multirow{2}{*}{CNN}
& Articles & .49 & .53 & .45 & .53 & .50 & .48 & .50 & .53 & .52 & .50 \\
& Others   & .47 & \textbf{.56} & .45 & .52 & .50 & .49 & .44 & .55 & .48 & .56 \\
\multirow{2}{*}{BiLSTM}
& Articles & .46 & .47 & .49 & .51 & .51 & .51 & .49 & .53 & .49 & .50 \\
& Others   & .44 & .45 & .48 & .56 & \textbf{.57} & \textbf{.56} & .43 & .57 & .52 & .54 \\
\multirow{2}{*}{CNN-BERT}
& Articles & .53 & .53 & .44 & .53 & .52 & .50 & .51 & .52 & .48 & .51 \\
& Others   & .48 & .56 & .43 & .57 & .43 & .54 & .44 & .58 & .54 & .57 \\
\midrule
\multirow{2}{*}{BERT}
& Articles & .62 & .47 & .49 & .53 & .59 & .48 & .49 & \textbf{.61} & .49 & .52 \\
& Others   & .60 & .44 & .46 & \textbf{.57} & .48 & .43 & .43 & .51 & .51 & .52 \\
\multirow{2}{*}{DeBERTa}
& Articles & \textbf{.68} & .49 & .51 & .53 & .59 & .48 & .49 & .57 & \textbf{.64} & \textbf{.62} \\
& Others   & .55 & .44 & .46 & \textbf{.57} & \textbf{.57} & .43 & .46 & .59 & .62 & .55 \\ \midrule
\multirow{2}{*}{Llama (FS)}
& Articles & .82 & .88 & .58 & .52 & .71 & .80 & .73 & .58 & .69 & .67 \\
& Others   & .82 & .89 & .57 & .53 & .70 & .81 & .66 & .60 & .72 & .73 \\
\multirow{2}{*}{Qwen (FS)}
& Articles & .79 & .85 & .58 & .52 & .73 & .81 & .73 & .57 & .82 & .77 \\
& Others   & .80 & .88 & .58 & .53 & .73 & .82 & .76 & .56 & .82 & .78 \\
\multirow{2}{*}{Zephyr (FS)}
& Articles & .76 & .85 & .52 & .52 & .60 & .61 & .71 & .82 & .79 & .60 \\
& Others   & .79 & .86 & .47 & .53 & .65 & .62 & .63 & .81 & .80 & .65 \\
\bottomrule
\end{tabularx}
\end{table}

Across all model families, we observe that training and testing within the same genre consistently yields higher and more stable performance. Conversely, genre-shift scenarios lead to systematic degradation, even when topic and annotation conventions are relatively aligned. This indicates that genre mismatch is an independent and substantial contributor to cross-dataset failures. Moreover, the degradation is asymmetric: models trained on article-style datasets transfer more effectively to short-form datasets than the reverse. This suggests that richer, longer inputs provide more robust signals for generalization, whereas models trained on short claims struggle to adapt to more complex structures.

A detailed analysis of the results obtained by both the topic-shift and genre-shift investigations highlights how these different dimensions drive the observed generalization drop. Genre-shift represents a major barrier: as shown in Table \ref{tab:articles_others}, models trained on a specific genre exhibit substantial performance degradation when tested on a different one. For instance, when DeBERTa is trained on the Articles genre, it achieves an F1-score of 0.68 on Celebrity, but this performance drops significantly when evaluated on other datasets within the same genre-training aggregate. This suggests that structural features and stylistic markers learned from a specific genre do not translate seamlessly even within the same macro-category if the underlying source varies.

Although topic shift also leads to severe lexical depletion, its impact is most evident in the analysis of Narrow vs. Generic datasets (Table \ref{tab:generic_narrow}): models trained on Narrow datasets show consistent degradation when tested on Generic data. For example, DeBERTa trained on the Narrow aggregate achieves an F1-score of only 0.54 on Celebrity and 0.43 on FakeVsSatire (Generic datasets), with performance improving slightly when training and testing occur within the same type of dataset. This suggests that the models overfit the specific vocabulary of historical or current events, making them ineffective when the news cycle shifts to broader topics.

We also observe an annotation shift as an additional driver of degradation. Even when the genre or topic-type is consistent, the shift from clear binary definitions to the nuanced fact-checking taxonomies of LIAR-PLUS or Politifact creates a performance ceiling. As shown in Table \ref{tab:generic_narrow}, DeBERTa trained on the Generic aggregate struggles to exceed an F1-score of 0.57 on LIAR-PLUS and 0.63 on Politifact. This suggests that the semantic boundary between ``fabrication'' and ``half-truth'' remains inherently blurry for classifiers trained on simpler binary rationales, representing a fundamental drift in the definition of misinformation.

Finally, to assess whether statistically significant differences exist among models within each stratum (narrow vs. generic and articles vs. others), we applied a Friedman test followed by a post-hoc Nemenyi test.\footnote{See Section 2 of the Supplementary Material for the results of the statistical tests.}

We can conclude that a classification-based approach relying on training on one specific dataset to learn the distinction between real and fake news may be sub-optimal. Indeed, the domain and the timing of the news, as well as its relationship with other news, may play a crucial role that is overlooked when leveraging simple text-based classification.

\subsection{Mixed-Training/Single-Test Experiments}\label{sect:exp3}

Main results from the Mixed-Training/Single-Test experiments are summarized in Table \ref{tab:all_f1_models_by_dataset_mixed}.\footnote{See Section 3 of the Supplementary Material for detailed results.}

\begin{table}[H]
\scriptsize
\caption{F1-score obtained by all models on each dataset and on the Global test set. The decoder-only models are considered in both zero-shot (ZS) and few-shot (FS) scenarios. The best F1-score for each dataset is highlighted in bold.}
\label{tab:all_f1_models_by_dataset_mixed}
\begin{tabularx}{\textwidth}{@{}l*{11}{>{\centering\arraybackslash}X}@{}}
\toprule
\textbf{Model} &
\rotatebox{90}{\textbf{Celebrity}} &
\rotatebox{90}{\textbf{Cidii}} &
\rotatebox{90}{\textbf{FakeVsSatire}} &
\rotatebox{90}{\textbf{Fakes}} &
\rotatebox{90}{\textbf{Horne}} &
\rotatebox{90}{\textbf{Infodemic}} &
\rotatebox{90}{\textbf{Isot}} &
\rotatebox{90}{\textbf{LIAR-PLUS}} &
\rotatebox{90}{\textbf{NDF}} &
\rotatebox{90}{\textbf{Politifact}} &
\rotatebox{90}{\textbf{Global}} \\ 
\midrule
LR                  & .58          & .85          & .72          & .49          & .71          & .76          & .68          & .53          & .82          & .71          & .61 \\
SVM                 & .66          & .80          & .72          & .53          & .64          & .77          & .74          & .55          & .81          & .75          & .60 \\
NB                  & .54          & .75          & .60          & .51          & .76          & .66          & .55          & .53          & .69          & .72          & .58 \\ \midrule
CNN                 & .54          & .62          & .44          & \textbf{.54}          & .66          & .51         & .53          & .55          & .62          & .63          & .49 \\
BiLSTM              & .49          & .64          & .38          & .53 & .68         & .54          & .50          & .55          & .58          & .65          & .52 \\
CNN-BERT            & .53          & .53          & .36          & \textbf{.54}          & .68          & .55          & .59          & .49          & .54          & .64          & .52 \\ \midrule
BERT               & .77          & .89 & .69          & .52          & .71          & .87 & .95 & .57          & .80          & .84          & .69 \\
DeBERTa             & .83 & \textbf{.94}          & \textbf{.78} & .50          & \textbf{.87} & \textbf{.89} & \textbf{.98} & .58          & \textbf{.89}          & \textbf{.85} & \textbf{.73} \\

MERMAID             & .62 & .25          & .77 & .48          & .67 & .84 & .85 & .56          & .71          & .82 & .66 \\ \midrule 

LLaMa3-8B (ZS)      & \textbf{.87} & .91          & .66          & \textbf{.54} & .63          & .78          & .64          & .61          & .83          & .75          & .62 \\
LLaMa3-8B (FS)      & .86          & .91          & .66          & \textbf{.54} & .64          & .83          & .72          & .59          & .75          & .72          & .58 \\
Qwen3-32B (ZS)      & .85          & .92          & .66          & .52          & .60          & .83          & .75          & .58          & .86          & .83          & .64 \\
Qwen3-32B (FS)      & .82          & .87          & .66          & .52          & .64          & .83          & .74          & .59          & .86          & .78          & .66 \\
Zephyr-7B-beta (ZS) & .73          & .85          & .52          & \textbf{.54} & .51          & .75          & .61          & .62          & .83          & .69          & .54 \\
Zephyr-7B-beta (FS) & .71          & .86          & .54          & .51          & .57          & .79          & .59          & \textbf{.63} & .81          & .64          & .55 \\
\bottomrule
\end{tabularx}
\end{table}

We observe results with trends similar to those of the Dataset-Specific experiments. Model performance varies widely across datasets, further confirming the difficulty of generalizing fake news detection across domains. In several cases models improve their performances compared to Dataset-Specific experiments, particularly on smaller datasets.  This suggests that, in data-constrained settings, incorporating data from other domains may act as a beneficial form of data augmentation.

Again, DeBERTa emerges as the top performer, confirming that pre-trained Transformers can leverage even a limited, mixed-domain training data to learn more robust decision boundaries. In contrast, MERMAID, despite being specifically designed for cross-domain tasks, shows inconsistent performance in this specific setup that could be linked to a routing collapse within the gating mechanism (see Table \ref{tab:mermaid-routing}). The Gate must not only classify Fake/Real, but also learn to identify the originating domain of each instance in order to route it to the appropriate expert. We quantify routing collapse using two complementary diagnostics: (i) gate entropy, computed as the Shannon entropy of the router’s probability distribution over experts, and (ii) the coefficient of variation (CV) of expert importance \cite{shazeer2017outrageously}, defined as the standard deviation divided by the mean of the experts’ importance weights (i.e., the batchwise sum of gate probabilities).  
Gate entropy captures how spread the routing distribution is, whereas CV reflects how uneven the cumulative expert importance becomes across a batch. Their combination allows distinguishing different routing regimes: low entropy + high CV signals a collapse where the Gate tends to select the same expert for the majority of inputs; high entropy + low CV indicates balanced and stable routing; intermediate patterns correspond to partial specialization.

\begin{table*}[t]
\centering
\caption{Routing diagnostics for MERMAID across all test sets. For each dataset we report the gate entropy (lower = more deterministic routing)  
and the CV of expert utilization (higher = routing collapse).
}
\tiny
\begin{tabular}{lccccccccccc}
\toprule
& \rotatebox{90}{\textbf{Celebrity}} & \rotatebox{90}{\textbf{Cidii}} & 
\rotatebox{90}{\textbf{FakeVsSatire}} & \rotatebox{90}{\textbf{Fakes}} &  
\rotatebox{90}{\textbf{Horne}} & \rotatebox{90}{\textbf{Infodemic}} & \rotatebox{90}{\textbf{Isot}} & \rotatebox{90}{\textbf{LIAR-PLUS}} & \rotatebox{90}{\textbf{NDF}} & 
\rotatebox{90}{\textbf{Politifact}} & \rotatebox{90}{\textbf{Global}} \\
\midrule
\textbf{Gate Entropy} 
& 0.003 & 0.053 & 0.002 & 0.007 & 0.057 & 0.175 & 0.007 & 0.780 & 0.175 & 0.034 & 0.078 \\
\textbf{CV} 
& 2.797 & 2.013 & 2.998 & 2.565 & 2.737 & 2.625 & 2.943 & 1.638 & 2.185 & 2.321 &  2.420 \\
\bottomrule
\end{tabular}

\label{tab:mermaid-routing}
\end{table*}

In the Mixed-Training/Single-Test setting, MERMAID exhibits low entropy together with consistently high CV. This pattern suggests that in this setting and with little data per domain, the gating mechanism fails to learn domain-aware routing, causing a domain embedding mismatch where instances are routed to suboptimal experts. To assess the impact of training data availability in this setup, we performed a controlled scaling study on six datasets, which  were the only ones with enough examples to support all three balanced sampling regimes. For each dataset, we sampled balanced subsets of increasing size (100/100, 150/150, and 200/200 real/fake instances) comparing MERMAID and DeBERTa. Table \ref{tab:ablation_mixed} reports the F1‑scores obtained by the two approaches across the three regimes.

\begin{table}[H]
\scriptsize
\caption{F1‑scores for DeBERTa and MERMAID trained on balanced mixed‑domain subsets of increasing size. }
\label{tab:ablation_mixed}

\begin{tabularx}{\textwidth}{@{}l*{7}{>{\centering\arraybackslash}X}@{}}
\toprule
\textbf{Model} &
\rotatebox{90}{\textbf{Celebrity}} &
\rotatebox{90}{\textbf{Cidii}} &
\rotatebox{90}{\textbf{Fakes}} &
\rotatebox{90}{\textbf{Infodemic}} &
\rotatebox{90}{\textbf{Isot}} &
\rotatebox{90}{\textbf{LIAR-PLUS}} &
\rotatebox{90}{\textbf{Global}} \\ 
\midrule
{Deberta (train 100/100)}                  & .84          & .98          & .48          & .88          & .99          & .57          & .64          \\
{Mermaid (train 100/100)}                 & .47          & .52          & .48          & .84          & .92          & .56          & .64          \\\midrule
{Deberta (train 150/150)}                 & .79          & .98          & .47          & .87          & .99          & .59          & .64          \\
{Mermaid (train 150/150)}              & .57          & .78          & .49          & .80 & .92         & .54          & .59           \\\midrule
{Deberta (train 200/200)}            & .84          & .98          & .40          & .91          & .99          & .60          & .66          \\
{Mermaid (train 200/200)}                & .68          & .82 & .51          & .82          & .85          & .57 & .54  \\

\bottomrule
\end{tabularx}
\end{table}

\begin{figure}[!htb]
\centering
\includegraphics[width=0.95\linewidth]{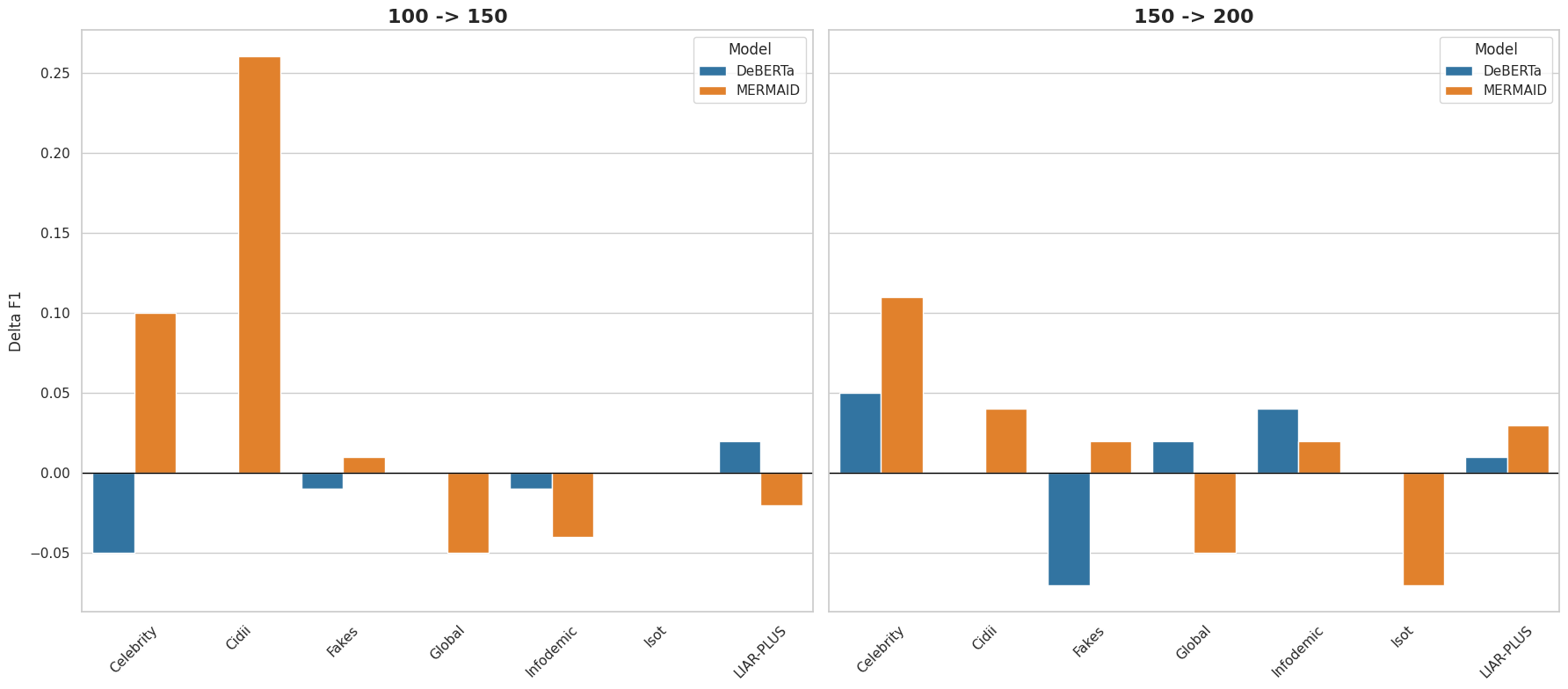} 
\caption{F1‑score gains for DeBERTa and MERMAID at each increase in training data size.}
\label{fig:delta-f1-mixed-ablation}
\end{figure}

Figure \ref{fig:delta-f1-mixed-ablation} shows the marginal gain achieved by the two models at each increment of data.

Across all data regimes, DeBERTa exhibits consistent and stable performance on almost all datasets, with few and negligible increases. MERMAID improves on several datasets as the per‑dataset sample size increases, most notably on Celebrity, Cidii, and Fakes, confirming that its architecture requires a minimum data threshold to develop the specialization needed for effective cross‑domain routing. However, consistent performance drops occur on datasets such as Isot and on the Global test set. In general, MERMAID remains competitive but generally underperforms compared to DeBERTa. It is worth noting that increasing the per‑dataset sampling cap does not always lead to more stable training. In fact, as noted in the Dataset Specific experiment in Section \ref{sect:exp1}, performances across datasets are variable. This suggests that, beyond a certain threshold, additional training data may introduce domain‑specific noise or inconsistencies that contaminate the training signal rather than strengthen it.
This suggests that MERMAID does not simply (or not necessarily) require  more data, but rather data that is homogeneous enough to allow its experts to specialize effectively. At the same time, its architecture does benefit from large-scale training  data, as evidenced by the Leave-One-Dataset-Out experiment in Section \ref{sect:exp4}), where MERMAID performs more robustly. In contrast, when constrained to smaller or heterogeneous training sets, the model struggles to generalize reliably.

As for the decoder-only models, zero-shot performance remains consistent with the previous experiments, with only minor fluctuations due to the different composition of the test sets. The few-shot results are akin to those of the Cross-Dataset experiments. Changes over zero-shot are marginal, nonexistent, or even negative. This further suggests that in-context examples are beneficial primarily when classifying same-domain data, whereas more varied sampling may introduce noise that misleads the model and degrades performance on out-of-domain instances.

We included a mixed training experiment variant that mirrors the natural distribution of the ten datasets in \ref{sect:mixed_imbalanced_variant}.

\begin{figure}[!htb]
\centering
\includegraphics[width=0.95\linewidth]{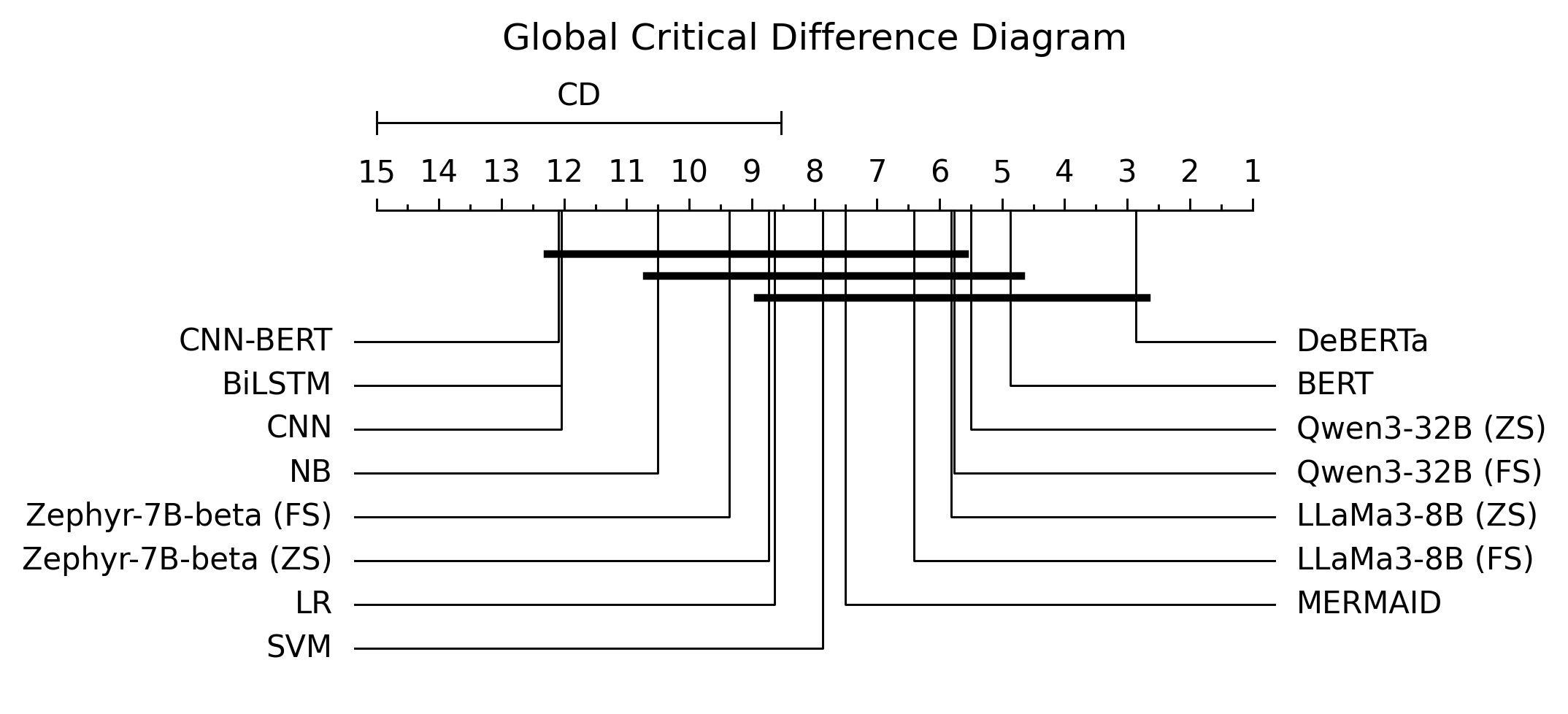} 
\caption{Critical-difference diagram showing the statistically significant differences among models according to the Nemenyi post-hoc test.}
\label{fig:critical_difference_diagram_exp3} 
\end{figure}

To assess whether statistically significant differences exist among the models, we first computed the Friedman test and then applied the post-hoc Nemenyi test. The critical-difference diagram in Figure \ref{fig:critical_difference_diagram_exp3} summarizes the statistical comparisons across all models.\footnote{See Section 3 of the Supplementary Material for the results of the statistical tests.} Results reveal a scenario similar to that of the Dataset-Specific experiments. Most models do not statistically differ from each other, except for DeBERTa, which is statistically different from most of the other models.

\subsection{Leave-One-Dataset-Out Experiments}\label{sect:exp4}

Table \ref{tab:all_f1_models_by_dataset_lodo} reports the main results obtained in the Leave-One-Dataset-Out experiments.\footnote{See Section 4 of the Supplementary Material for detailed results.}

\begin{table}[H]
\scriptsize
\caption{F1-scores obtained by all models on each dataset in the Leave-One-Dataset-Out experiments. The
decoder-only models are considered both in zero-shot (ZS) and few-shot (FS) scenarios. The best F1-score for each dataset is highlighted in bold.}
\label{tab:all_f1_models_by_dataset_lodo}
\begin{tabularx}{\textwidth}{@{}l*{10}{>{\centering\arraybackslash}X}@{}}
\toprule
\textbf{Model} &
\rotatebox{90}{\textbf{Celebrity}} &
\rotatebox{90}{\textbf{Cidii}} &
\rotatebox{90}{\textbf{FakeVsSatire}} &
\rotatebox{90}{\textbf{Fakes}} &
\rotatebox{90}{\textbf{Horne}} &
\rotatebox{90}{\textbf{Infodemic}} &
\rotatebox{90}{\textbf{Isot}} &
\rotatebox{90}{\textbf{LIAR-PLUS}} &
\rotatebox{90}{\textbf{NDF}} &
\rotatebox{90}{\textbf{Politifact}} \\
\midrule
{LR}                  & .57          & .46          & \textbf{.63} & .48          & .74          & .52          & .73          & .45          & .42          & .61 \\
{SVM}                 & .51          & .51          & .55          & .53          & .57          & .52          & .55          & .46          & .52          & .47 \\
{NB}                  & .50          & .43          & .55          & .53          & .53          & .55          & .61          & .53          & .39          & .52 \\ \midrule
{CNN}                 & .50          & .51          & .55          & .53          & .57          & .52          & .55          & .46          & .52          & .47 \\
{BiLSTM}              & .51          & .51          & .55          & .53          & .57          & .52          & .55          & .46          & .52          & .47 \\
{CNN-BERT}            & .55          & .45          & .59          & .51          & .64          & .44          & .49          & .47          & .38          & .57 \\ \midrule
{BERT}                & .57          & .46          & \textbf{.63} & .48          & .74          & .52          & .73          & .45          & .42          & .61 \\
{DeBERTa}             & .65          & .44          & .60          & .51          & .74          & .60          & \textbf{.83} & .45          & .47          & .67 \\
{MERMAID}             & .69          & .47          & \textbf{.63} & .54          & \textbf{.83} & .48          & .76          & .47          & .66          & \textbf{.81} \\\midrule
{LLaMa3-8B (ZS)}      & \textbf{.86} & .90          & .58          & .52          & .72          & .78          & .74          & \textbf{.62} & .79          & .54 \\
{LLaMa3-8B (FS)}      & .85          & .90          & .57          & .52          & .70          & .82          & .63          & .60          & .73          & .73 \\
{Qwen3-32B (ZS)}      & .82          & \textbf{.92} & .52          & .57          & .72          & .83          & .75          & .58          & \textbf{.83} & .79 \\
{Qwen3-32B (FS)}      & .81          & .85          & .52          & \textbf{.58} & .73          & .82          & .76          & .57          & .81          & .79 \\
{Zephyr-7B-beta (ZS)} & .73          & .84          & .45          & .53          & .57          & .75          & .61          & \textbf{.62} & .78          & .60 \\
{Zephyr-7B-beta (FS)} & .77          & .72          & .48          & .51          & .65          & \textbf{.84} & .60          & \textbf{.62} & .78          & .65 \\
\bottomrule
\end{tabularx}
\end{table}

Here, MERMAID achieves robust performance. Among the fully-trained models, it performs best in 6 out of 10 unseen domains, providing evidence that a Mixture-of-Experts,\footnote{Here, Mixture-of-Expert is used as defined in \citep{liguori2025breaking} not with the now-common definition typically used for MoE LLMs.} approach can generalize effectively when provided with a sufficiently large and diverse set of source domains.

As for the other fully-trained models, DeBERTa is the most consistently high performer. It achieves the best score in 2 of the 10 domains and frequently ranks close to the top-performing model in the remaining ones. This suggests that its pre-training captures more abstract and transferable signals of misinformation than BERT.

However, results also reveal that not all held-out datasets pose the same type of generalization challenge. As shown in Table \ref{tab:dataset_selected} (see Sec. \ref{sec:experiments}), most datasets adopt a binary fake/real annotation grounded in fact-checking (e.g., Celebrity, Cidii, Fakes, Infodemic, Isot, NDF), or a closely related binary distinction such as fake vs.\ satire (FakeVsSatire, Horne). In contrast, LIAR-PLUS originally uses a six-class fact-checking taxonomy, which we remap into binary labels for consistency.

Across traditional ML models, transformer encoders, and decoder-only LLMs, performance is generally higher and more stable on datasets whose binary labels directly reflect fact-checking judgments or simple fake/real distinctions (e.g., FakeVsSatire, Horne, Isot). LIAR-PLUS is the most challenging held-out dataset for nearly all models, with substantially lower scores and limited improvements in the few-shot setting. However, the sensitivity analysis in \ref{sect:lodo_satire_analysis} shows that when satire-related datasets are excluded from the training, trends in model performances remain relatively stable.

Taken together, these results suggest that the difficulty does not stem solely from domain differences, but also from the fact that the underlying label definitions do not fully align with the binary fake/real framing used elsewhere. In other words, part of the observed variability reflects differences in how “truthfulness” is operationalized across datasets, rather than cross-domain transfer alone.

\begin{figure}[!htb]
\centering
\includegraphics[width=0.95\linewidth]{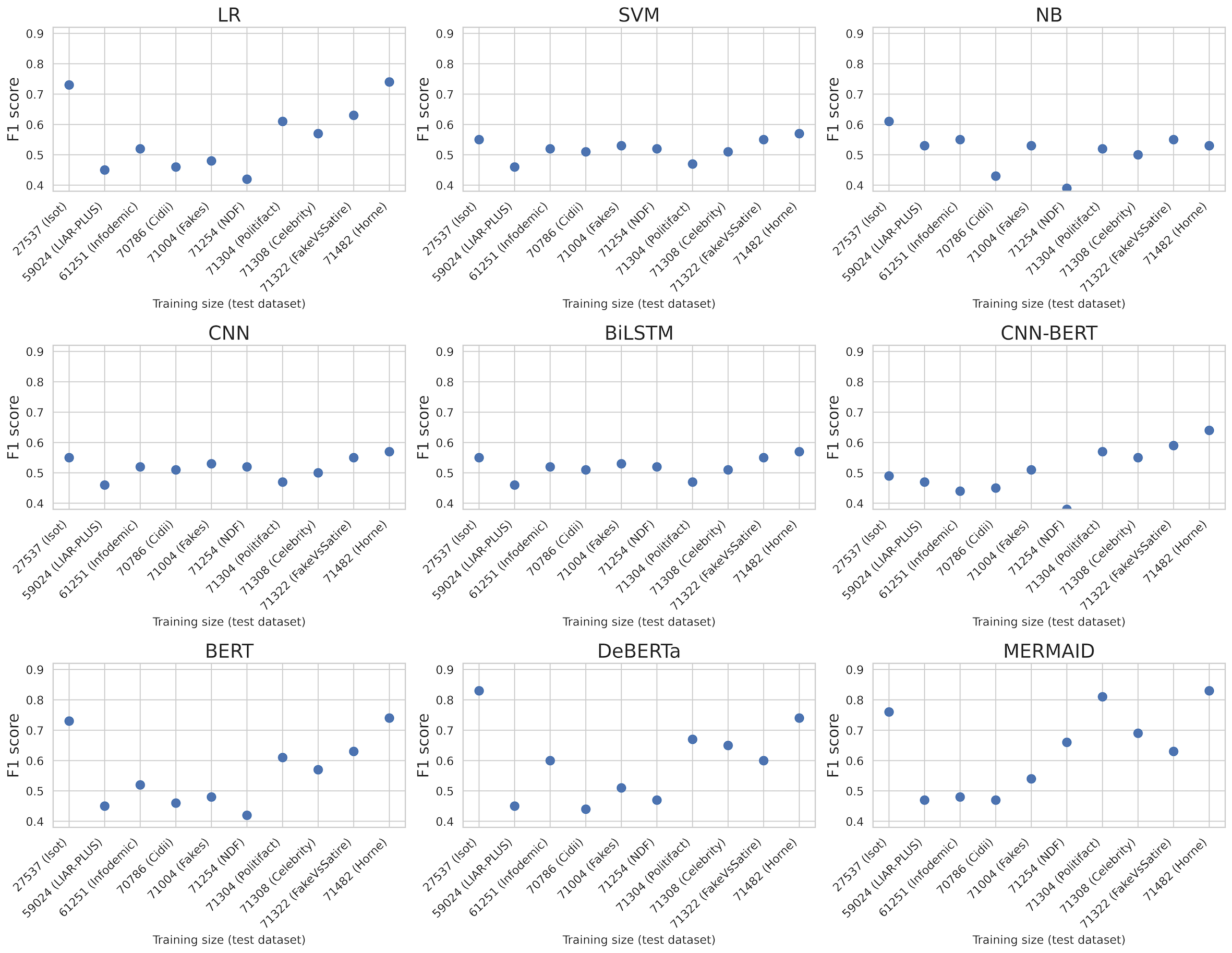} 
\caption{Relationship between each model’s held‑out performance ranking and the size of the remaining training pool.}
\label{fig:mask_generalization_issue} 
\end{figure}

To assess whether large training pools can mask generalization issues we report a per‑model scatter plot in Figure \ref{fig:mask_generalization_issue}. Each point corresponds to a held‑out dataset: the x‑axis ranks datasets from the smallest to the largest training pool (i.e. the union of all other datasets), while the y‑axis reports the model's F1 score on that dataset. 

We observe that for traditional ML and DL models, training dataset size weakly correlates with performances on the held-out dataset. Transformer-based models on the other hand, exhibit a more pronounced scaling trend, with larger training sizes correlating to better performances on the held-out data. Thus, we can conclude that large aggregated training sets can partially obscure generalization difficulties, especially for better-performing models.

As for LLMs, all models exhibit remarkable robustness in the zero-shot setting. However, the few-shot setting again shows limitations. Because the in-context examples are drawn from the other domains, as for Cross-Dataset and Mixed-Training/Single-Test experiments, performance can improve on challenging datasets such as Fakes, but can degrade on datasets  with strong unique characteristics, such as Cidii or Isot. This reaffirms that providing out-of-domain examples can sometimes introduce noise that may mislead the model rather than guiding it. To better understand the stability of decoder-only LLMs in the few-shot setting, we additionally evaluate the impact of exemplar selection. Further details are reported in \ref{sect:appendix_few_shot_calibration}.

\begin{figure}[!htb]
\centering
\includegraphics[width=0.95\linewidth]{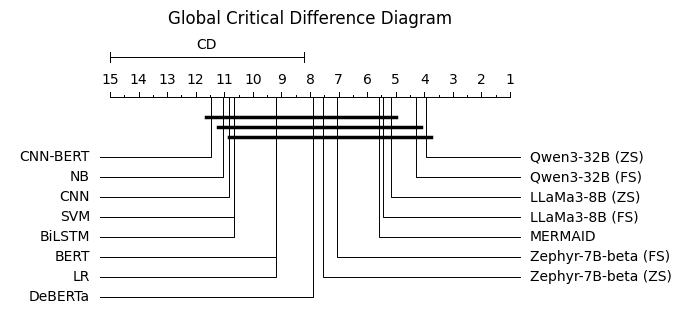} 
\caption{Critical-difference diagram showing the statistically significant differences among
models according to the Nemenyi post-hoc test.}
\label{fig:critical_difference_diagram_exp4} 
\end{figure}

To assess whether statistically significant differences exist among the
models, we first computed the Friedman test and then applied the post-
hoc Nemenyi test. The critical-difference diagram in Figure \ref{fig:critical_difference_diagram_exp4} summarizes the statistical comparisons across all models.\footnote{See Section 4 of the Supplementary Material for the results of the statistical tests.} The results reveal that most models are statistically similar to each other. We observe that decoder-only LLMs are, for the most part, closer to being different (i.e., to the 0.05 threshold), but only Qwen3-32B actually exhibit statistical differences from few of the other models.

\section{Overall Findings}
\label{sect:discussion}

A consistent picture emerges from the experiments: fake news detection remains highly domain-sensitive, and models struggle to acquire domain-invariant representations that transfer reliably to new, unseen contexts. 

Table \ref{tab:robustness_summary} summarizes the median and interquartile range (IQR) of each model across all four experimental settings.

\begin{table}[H]
\scriptsize
\centering
\caption{Robustness summary across the four experiments. 
For each model we report the median and IQR of F1-scores for the dataset-specific (E1), cross-dataset (E2), 
 mixed-training/single-test (E3), 
and leave-one-dataset-out  (E4) experiments.}
\label{tab:robustness_summary}
\begin{tabularx}{\textwidth}{l *{8}{>{\centering\arraybackslash}X}}
\toprule
\textbf{Model} &
\textbf{Med (E1)} & \textbf{IQR (E1)} &
\textbf{Med (E2)} & \textbf{IQR (E2)} &
\textbf{Med (E3)} & \textbf{IQR (E3)} &
\textbf{Med (E4)} & \textbf{IQR (E4)} \\
\midrule
LR & 0.765 & 0.2875 & 0.520 & 0.0775 & 0.71 & 0.155 & 0.545 & 0.16 \\
SVM & 0.73 & 0.175 & 0.510 & 0.0700 & 0.72 & 0.145 & 0.52 & 0.035 \\
\midrule
NB & 0.695 & 0.2525 & 0.515 & 0.0900 & 0.60 & 0.1675 & 0.53 & 0.04 \\
CNN & 0.56 & 0.025 & 0.500 & 0.0750 & 0.54 & 0.10 & 0.52 & 0.0425 \\
BiLSTM & 0.605 & 0.0825 & 0.505 & 0.0800 & 0.54 & 0.115 & 0.52 & 0.035 \\
CNN-BERT & 0.59 & 0.09 & 0.500 & 0.0800 & 0.54 & 0.055 & 0.50 & 0.11 \\
\midrule
BERT & 0.805 & 0.2075 & 0.510 & 0.1375 & 0.77 & 0.1725 & 0.545 & 0.16 \\
DeBERTa & 0.865 & 0.175 & 0.535 & 0.1200 & 0.85 & 0.135 & 0.60 & 0.185 \\
MERMAID & -- & -- & -- & -- & 0.67 & 0.2175 & 0.645 & 0.2475 \\
\midrule
LLaMa3-8B (ZS) & 0.74 & 0.15 & 0.745 & 0.1700 & 0.66 & 0.1925 & 0.73 & 0.1975 \\
LLaMa3-8B (FS) & 0.75 & 0.23 & 0.705 & 0.1900 & 0.72 & 0.195 & 0.715 & 0.19 \\
Qwen3-32B (ZS) & 0.77 & 0.215 & 0.770 & 0.2500 & 0.75 & 0.225 & 0.77 & 0.2125 \\
Qwen3-32B (FS) & 0.755 & 0.20 & 0.760 & 0.2300 & 0.74 & 0.1775 & 0.775 & 0.1925 \\
Zephyr-7B-beta (ZS) & 0.645 & 0.1625 & 0.615 & 0.1800 & 0.62 & 0.205 & 0.615 & 0.1675 \\
Zephyr-7B-beta (FS) & 0.645 & 0.275 & 0.650 & 0.1900 & 0.63 & 0.21 & 0.65 & 0.1525 \\
\bottomrule
\end{tabularx}
\end{table}

The robustness summary reveals distinct patterns depending on the deployment scenario. Traditional transformers such as BERT and DeBERTa deliver the strongest in‑domain performance (high medians in E1 and E3), but their sharp drops in E2 and E4 indicate that they remain highly sensitive to domain shift. The cross‑domain architecture MERMAID, despite lower peak performance, exhibits more stable behavior under distributional drift, making them preferable when robustness is a priority and training data are heterogeneous. Decoder‑only LLMs show the most balanced profile: their medians remain consistently high across all settings, and their IQRs, although larger than those of cross‑domain models, are markedly more stable than those of fully trained transformers. This suggests that LLMs are particularly suitable when deployment involves unseen or evolving domains, provided that their variability can be mitigated through prompting strategies or exemplar selection.

The Dataset-Specific experiments confirm that large pre-trained Transformers achieve the strongest in-domain performance, outperforming other approaches. However, for fully-trained Transformers, this advantage diminishes sharply in cross-domain settings.

The Cross-Dataset experiments show that all supervised models, including fine-tuned Transformers, generalize poorly. Training on data from the same type (e.g., generic$\rightarrow$generic or narrow$\rightarrow$narrow) yields the most stable outcomes, whereas mismatched scenarios show systematically worse performances. These findings indicate that ``generalist'' fine-tuned models such as DeBERTa may overfit dataset-specific patterns rather than build robust indicators of misinformation. Table \ref{tab:robustness_summary} reinforces this interpretation: models such as BERT and DeBERTa exhibit high in-domain medians but low cross-domain medians, suggesting that their representations remain specialized to the training distribution.

The Mixed-Training/Single-Test experiments further refine this observation. When trained on a small but heterogeneous mix of domains, models sometimes improve their performance on smaller datasets, suggesting that mixed-domain training can offer gains in generalization. However, these improvements remain limited: fully trained models still struggle when compared with domains that differ substantially from those seen during training.

The inclusion of a model explicitly designed for cross-domain learning yields some interesting insights. It consistently outperforms other fully trained models on truly out-of-domain data, confirming that this kind of architecture can actually transfer knowledge more effectively. However, this is only true provided that the training set is both large and varied enough, as demonstrated by its instability in the Mixed-Dataset experiments.

Decoder-only LLMs offer a different and compelling perspective. Their zero-shot robustness is remarkably consistent, often matching or surpassing supervised baselines without requiring task-specific training. This suggests that large-scale pre-training endows these models with broad contextual knowledge that transfers effectively to new domains. This is reflected in their strong and relatively stable medians across experiments. However, the few-shot results demonstrate an important nuance: in-context examples help only when sampled from the same domain, but become detrimental when they originate from unrelated domains. This reinforces the idea that LLMs possess a strong implicit understanding of the task, and that poorly curated in-context examples may introduce misleading domain cues rather than facilitating adaptation.

To assess whether this sensitivity is an inherent limitation of decoder-only LLMs or primarily a consequence of mismatched demonstrations, we further evaluated a simple exemplar-selection baseline. To ensure a controlled setting, we relied on the same mixed-domain configuration used in the Mixed-Training/Single-Test experiments, drawing exemplars from the global training set rather than from dataset-specific pools. For each test instance, we computed its embedding and retrieved the six most similar training examples based on cosine similarity, to match the same number of exemplars taken in the standard configuration. These exemplars were then used as few-shot demonstrations in place of randomly sampled ones.

\begin{table}[H]
\scriptsize
\caption{F1‑scores of decoder‑only models using similarity‑based and random exemplar selection in the mixed‑domain few‑shot setting}
\label{tab:f1_llama_qwen_zephyr_emb_retrieval}
\begin{tabularx}{\textwidth}{@{}l*{11}{>{\centering\arraybackslash}X}@{}}
\toprule
\textbf{Model} &
\rotatebox{90}{\textbf{Celebrity}} &
\rotatebox{90}{\textbf{Cidii}} &
\rotatebox{90}{\textbf{FakeVsSatire}} &
\rotatebox{90}{\textbf{Fakes}} &
\rotatebox{90}{\textbf{Horne}} &
\rotatebox{90}{\textbf{Infodemic}} &
\rotatebox{90}{\textbf{Isot}} &
\rotatebox{90}{\textbf{LIAR-PLUS}} &
\rotatebox{90}{\textbf{NDF}} &
\rotatebox{90}{\textbf{Politifact}} &
\rotatebox{90}{\textbf{Global}} \\
\midrule
{LLaMa3-8B (FS - Similar)}      
& .83 & .96 & .64 & .52 & .52 & .83 & .71 & .57 & .89 & .60 & .58 \\

{LLaMa3-8B (FS - Random)}      
& .86 & .91 & .66 & .54 & .64 & .83 & .72 & .59 & .75 & .72 & .58 \\

{Qwen3-32B (FS - Similar)}      
& .83 & .97 & .65 & .53 & .64 & .88 & .78 & .61 & .89 & .82 & .65 \\

{Qwen3-32B (FS - Random)}      
& .82 & .87 & .66 & .52 & .64 & .83 & .74 & .59 & .86 & .78 & .66 \\

{Zephyr-7B-beta (FS - Similar)} 
& .78 & .96 & .61 & .51 & .50 & .85 & .75 & .62 & .92 & .71 & .71 \\

{Zephyr-7B-beta (FS - Random)} 
& .71 & .86 & .54 & .51 & .57 & .79 & .59 & .63 & .81 & .64 & .55 \\
\bottomrule
\end{tabularx}
\end{table}

The results, reported in Table~\ref{tab:f1_llama_qwen_zephyr_emb_retrieval}, show that similarity-based exemplar selection has a clear stabilizing effect. For instance, LLaMa3-8B improves on datasets such as Cidii and NDF, while Qwen3-32B exhibits consistent gains across nearly all datasets. Zephyr-7B-beta shows the largest improvements, particularly on Cidii, NDF, and the Global test set. Although these improvements are not uniform across datasets, the results indicate that part of the few-shot brittleness stems from mismatched demonstrations rather than intrinsic model limitations. At the same time, the effectiveness of this mitigation depends on the availability of sufficiently similar training instances: when such examples are scarce or only loosely related, even embedding-based retrieval may introduce misleading cues that still confuse the model.

It is also worth noting that the datasets differ not only in topic or source but also in how their labels are defined. This suggests that part of the variability across experiments reflects differences in label semantics rather than domain shift alone.

Taken together, the results highlight the limitations of current supervised classifiers under the distribution shift represented in our experimental protocol. 
The fake news detection task is content-dependent but also deeply contextual, influenced by topical, temporal, and stylistic factors that vary dramatically across datasets. In this context, simple text-based classification show limited transferability under the tested shifts and may not fully capture the contextual complexity present in current benchmarks.
Models that perform well in-domain often do so because they exploit dataset-specific regularities rather than learning generalizable misinformation cues. The statistical tests further reinforce this observation. We observe in fact a statistically similar behavior across models in experiments where the evaluation is conducted in-domain, namely the Dataset-Specific and Mixed-Dataset experiments. In both cases, the models are evaluated on data, at least partly, from the same distribution as the training set. Statistical differences are also almost absent in the Leave-One-Dataset-Out experiment, likely because the larger training set partly masks generalization issues. On the other hand, the Cross-Dataset experiments, where the model has to generalize from a distribution to a completely different one, clearly show the differences between decoder-only LLMs and fully trained models. The latter exhibit poor performances with no statistical differences among them, while the former are much more resilient to this kind of domain shifts, and are significantly different from fully trained models.

These observations lead to three overarching conclusions:
\begin{itemize}
    \item Fine‑tuned Transformers generally achieve strong in‑domain performance, but their ability to generalize across domains appears limited, suggesting that their representations may remain partly tied to the training distribution.
    \item Architectures explicitly designed for cross-domain learning can generalize, but only when provided with extensive and diverse training data, limiting their applicability in low-resource or rapidly evolving scenarios.
    \item Decoder-only LLMs show a promising balance between robustness and data efficiency, achieving strong zero-shot performance without requiring domain-specific supervision, although their sensitivity to inappropriate in-context examples must be taken into account.
\end{itemize}

Overall, these findings suggest that progress in fake news detection requires moving beyond narrowly supervised models trained on isolated datasets, and toward data-efficient cross-domain methods that leverage broad pre-training, architectural specialization, or hybrid approaches. Developing models that can generalize reliably under realistic distribution shifts remains a central challenge, particularly within the constraints of current datasets and evaluation protocols.

\section{Limitations}\label{sec:limitations}

This work presents some limitations that should be acknowledged.

First, we do not account for the time of production/diffusion of fake news with the tested models. This may limit our analysis in two respects. On the one hand, we cannot assess the extent to which temporal factors influence model performance. On the other hand, we cannot control for potential exposure of LLMs to similar content during pretraining, potentially influencing their downstream performance. However, the performance drops observed in few-shot settings with out-of-domain examples suggest that any such effect is likely limited. The only consistent decline is observed in LLaMa3-8B (FS),  whose median F1-score decreases from 0.75 in the in-domain setting to 0.70 in the cross-dataset setting, an aspect that warrants further investigation.
Nevertheless, examining this issue in depth is not straightforward, as it would require precise knowledge of the texts included in the pre-training corpora—information that is typically unavailable. A more rigorous leakage-controlled evaluation would require a dedicated protocol. In future work, this could include: time-based splits whenever publication dates are available, ensuring that evaluation samples post-date the model’s pretraining window as well as overlap-detection heuristics based on lexical or semantic similarity to flag potentially memorized content. Although such measures cannot eliminate leakage entirely, they provide a principled framework for mitigating its impact and for interpreting LLM performance under more controlled conditions. 

Second, we do not explicitly address LLM-generated fake news, which represents an increasingly relevant threat to the robustness of current methods and a natural direction for future work.

Third, the datasets considered are collected and annotated by different sources using heterogeneous annotation protocols. Although we harmonize labels into a binary setting, some semantic nuances may be lost in this process. 

Fourth, our evaluation does not incorporate community- or propagation-based signals. While this choice is consistent with the scope of current text-only benchmarks, integrating social-contextual or group-dynamic features remains an important direction for future work.

Finally, our evaluation focuses on performances reflecting robustness and generalization rather than efficiency. While we report model size, training times, and inference cost proxies in \ref{sec:appendix_training_time}, we do not perform a systematic comparison under a fixed computational budget. The architectures considered span multiple design paradigms and technological generations, with heterogeneous resource requirements, making a strict compute-aware ranking nontrivial. Consequently, our conclusions should be interpreted as comparative in terms of performance and robustness, rather than efficiency, and a fully normalized cost-aware evaluation is left for future work.

These limitations primarily stem from the nature of currently available datasets. By focusing on peer-reviewed, publicly available benchmarks, we intentionally prioritize data quality and reproducibility, at the cost of reduced control over temporal and annotation-specific factors. 

\section{Conclusions and future work}\label{sec:conclusions} 
In this paper, we presented a large-scale empirical comparison of 12 fake news detection approaches across 10 datasets, focusing on the critical challenge of cross-domain generalization.

Our experiments, spanning in-domain, cross-dataset, multi-domain, and leave-one-dataset-out scenarios, reveal a fundamental flaw in the standard classification paradigm. Our study quantifies the severity of this issue: while fine-tuned transformers such as DeBERTa excel in-domain settings, achieving a median F1-score of 0.86, cross-dataset evaluation demonstrates that they fail to generalize to unseen topics. Specifically, performance drops substantially in cross-dataset settings, whit a median F1-score of 0.53.

Our findings identify two distinct paths to improving robustness. On the one hand, specialized cross-domain architectures (i.e., MERMAID) proved highly effective in generalizing to new domains, though they are data-hungry and require large, diverse training datasets to perform adequately.
On the other hand, decoder-only LLMs in zero-shot and few-shot settings could offer a data-efficient alternative, by demonstrating remarkable stability. This is exemplified by Qwen3-32B (ZS), which maintained a consistent median F1-score across the in-domain (0.77), cross-dataset (0.77), multi-domain (0.75), and leave-one-out (0.77) scenarios, showing only slight fluctuation.

Finally, this work highlights the need to explore novel paradigms for fake news detection. Rather than relying solely on generic features and isolated news, we argue that the focus should be shifted to more realistic scenarios where the context in which a news appears plays a crucial role.

Our future research will focus on strengthening the evaluation framework along three complementary directions that extend the analyses presented in this study.

A first direction involves analyzing the information divergence between a piece of fake news, the factual events from which it originates, and its surrounding contextual elements. This includes the exploration of retrieval‑augmented strategies, which may help mitigate current generalization limitations by grounding model predictions in external evidence and by reducing the impact of uncontrolled exposure or potential leakage.

A second direction concerns the integration of multimodal and socially grounded information. Since communication is inherently multimodal, extending cross‑dataset evaluations to settings that combine textual, visual, and contextual metadata could provide a more comprehensive understanding of model robustness and help address semantic inconsistencies that arise from heterogeneous annotation protocols.

A third direction focuses on the growing challenge posed by LLM‑generated content. As generative models become increasingly capable, evaluating the robustness of detection systems against both human‑ and LLM‑generated fake and real news becomes essential for assessing their reliability in realistic, rapidly evolving information environments.

\section*{Acknowledgements}

This work has been partly funded by the PNRR - M4C2 - Investimento 1.3, Partenariato Esteso PE00000013 - ``FAIR - Future Artificial Intelligence Research'' - Spoke 1 ``Human-centered AI'' under the NextGeneration EU programme, and the Italian Ministry of University and Research (MUR) in the framework of the PRIN 2022JLB83Z ``Psychologically-tailored approaches to Debunk Fake News detected automatically by an innovative artificial intelligence approach'', by the European Innovation Council project ``EMERGE'' (Grant No. 101070918), and the FoReLab and CrossLab projects (Departments of Excellence).

\bibliographystyle{elsarticle-num} 
\bibliography{references}

@inproceedings{wang2018eann,
  title={Eann: Event adversarial neural networks for multi-modal fake news detection},
  author={Wang, Yaqing and others},
  booktitle={KDD},
  pages={849--857},
  year={2018}
}

@inproceedings{pramanick2021multimodal,
  title={Multimodal learning for multi-label fake news detection},
  author={Pramanick, Rishabh and others},
  booktitle={ACM Multimedia},
  pages={4302--4310},
  year={2021}
}

@inproceedings{xu2022evidence,
  title={Evidence-aware fake news detection with graph neural networks},
  author={Xu, Weizhi and Wu, Junfei and Liu, Qiang and Wu, Shu and Wang, Liang},
  booktitle={Proceedings of the ACM web conference 2022},
  pages={2501--2510},
  year={2022}
}

@article{shu2019beyond,
  title={Beyond news content: The role of social context for fake news detection},
  author={Shu, Kai and others},
  journal={WSDM},
  year={2019}
}

@article{bian2020rumor,
  title={Rumor detection on social media with bi-directional graph convolutional networks},
  author={Bian, Tian and others},
  journal={AAAI},
  year={2020}
}

@article{shazeer2017outrageously,
  title={Outrageously large neural networks: The sparsely-gated mixture-of-experts layer},
  author={Shazeer, Noam and Mirhoseini, Azalia and Maziarz, Krzysztof and Davis, Andy and Le, Quoc and Hinton, Geoffrey and Dean, Jeff},
  journal={arXiv preprint arXiv:1701.06538},
  year={2017}
}

@article{raza2025fake,
  title={Fake news detection: comparative evaluation of BERT-like models and large language models with generative AI-annotated data},
  author={Raza, Shaina and Paulen-Patterson, Drai and Ding, Chen},
  journal={Knowledge and Information Systems},
  volume={67},
  number={4},
  pages={3267--3292},
  year={2025},
  publisher={Springer}
}

@article{niu2024veract,
  title={VeraCT scan: Retrieval-augmented fake news detection with justifiable reasoning},
  author={Niu, Cheng and Guan, Yang and Wu, Yuanhao and Zhu, Juno and Song, Juntong and Zhong, Randy and Zhu, Kaihua and Xu, Siliang and Diao, Shizhe and Zhang, Tong},
  journal={arXiv preprint arXiv:2406.10289},
  year={2024}
}

@inproceedings{thomas2025explainable,
  title={An Explainable KG-RAG-Based Approach to Evidence-Based Fake News Detection Using LLMs},
  author={Thomas, Jonathan John and Mihailescu, Radu-Casian},
  booktitle={Proceedings of the AAAI Symposium Series},
  volume={6},
  number={1},
  pages={152--154},
  year={2025}
}

@inproceedings{bai2024large,
  title={A Large Language Model-based Fake News Detection Framework with RAG Fact-Checking},
  author={Bai, Yangxiao and Fu, Kaiqun},
  booktitle={2024 IEEE International Conference on Big Data (BigData)},
  pages={8617--8619},
  year={2024},
  organization={IEEE}
}

@article{spitale2023ai,
  title={AI model GPT-3 (dis) informs us better than humans},
  author={Spitale, Giovanni and Biller-Andorno, Nikola and Germani, Federico},
  journal={Science Advances},
  volume={9},
  number={26},
  pages={eadh1850},
  year={2023},
  publisher={American Association for the Advancement of Science}
}

@article{guo2022survey,
  title={A survey on automated fact-checking},
  author={Guo, Zhijiang and Schlichtkrull, Michael and Vlachos, Andreas},
  journal={Transactions of the association for computational linguistics},
  volume={10},
  pages={178--206},
  year={2022},
  publisher={MIT Press One Rogers Street, Cambridge, MA 02142-1209, USA journals-info~…}
}

@inproceedings{janicka2019cross,
  title={Cross-domain failures of fake news detection},
  author={Janicka, Maria and Pszona, Maria and Wawer, Aleksander},
  booktitle={Computaci{\'o}n y Sistemas},
  volume={23},
  number={3},
  pages={1089--1097},
  year={2019}
}

@inproceedings{huang2025unmasking,
  title={Unmasking digital falsehoods: A comparative analysis of LLM-based misinformation detection strategies},
  author={Huang, Tianyi and Yi, Jingyuan and Yu, Peiyang and Xu, Xiaochuan},
  booktitle={2025 8th International Conference on Advanced Algorithms and Control Engineering (ICAACE)},
  pages={2470--2476},
  year={2025},
  organization={IEEE}
}

@article{comito2025learning,
  title={Learning Domain-Agnostic Fake News Detectors through Deep Self-Supervised Learning},
  author={Comito, Carmela and Guarascio, Massimo and Liguori, Angelica and Manco, Giuseppe and Pisani, Francesco Sergio},
  journal={IEEE Access},
  year={2025},
  publisher={IEEE}
}

@inproceedings{hoy2022exploring,
  title={Exploring the Generalisability of Fake News Detection Models},
  author={Hoy, Nathaniel and Koulouri, Theodora},
  booktitle={2022 IEEE International Conference on Big Data (Big Data)},
  pages={5731--5740},
  year={2022},
  organization={IEEE}
}

@misc{tunstall2023zephyr,
      title={Zephyr: Direct Distillation of LM Alignment}, 
      author={Lewis Tunstall and Edward Beeching and Nathan Lambert and Nazneen Rajani and Kashif Rasul and Younes Belkada and Shengyi Huang and Leandro von Werra and Clémentine Fourrier and Nathan Habib and Nathan Sarrazin and Omar Sanseviero and Alexander M. Rush and Thomas Wolf},
      year={2023},
      eprint={2310.16944},
      archivePrefix={arXiv},
      primaryClass={cs.LG}
}

@article{qwen3,
    title={Qwen3 Technical Report}, 
    author={An Yang and Anfeng Li and Baosong Yang et al.},
    journal = {arXiv preprint arXiv:2505.09388},
    year={2025}
}

@article{liguori2025breaking,
  title={Breaking domain barriers: mixture of experts for cross-domain fake news detection},
  author={Liguori, Angelica and Pisani, Francesco Sergio and Comito, Carmela and Guarascio, Massimo and Manco, Giuseppe},
  journal={Machine Learning},
  volume={114},
  number={8},
  pages={188},
  year={2025},
  publisher={Springer}
}

@inproceedings{wei2025cross,
  title={Cross-Domain Fake News Detection based on Dual-Granularity Adversarial Training},
  author={Wei, Wenjie and Zhang, Yanyue and Li, Jinyan and Liu, Panfei and Zhou, Deyu},
  booktitle={Proceedings of the 31st International Conference on Computational Linguistics},
  pages={9407--9417},
  year={2025}
}

@inproceedings{mosallanezhad2022domain,
  title={Domain adaptive fake news detection via reinforcement learning},
  author={Mosallanezhad, Ahmadreza and Karami, Mansooreh and Shu, Kai and Mancenido, Michelle V and Liu, Huan},
  booktitle={Proceedings of the ACM web conference 2022},
  pages={3632--3640},
  year={2022}
}

@inproceedings{nan2021mdfend,
  title={MDFEND: Multi-domain fake news detection},
  author={Nan, Qiong and Cao, Juan and Zhu, Yongchun and Wang, Yanyan and Li, Jintao},
  booktitle={Proceedings of the 30th ACM international conference on information \& knowledge management},
  pages={3343--3347},
  year={2021}
}

@article{zhu2022memory,
  title={Memory-guided multi-view multi-domain fake news detection},
  author={Zhu, Yongchun and Sheng, Qiang and Cao, Juan and Nan, Qiong and Shu, Kai and Wu, Minghui and Wang, Jindong and Zhuang, Fuzhen},
  journal={IEEE Transactions on Knowledge and Data Engineering},
  volume={35},
  number={7},
  pages={7178--7191},
  year={2022},
  publisher={IEEE}
}

@article{he2020deberta,
  title={Deberta: Decoding-enhanced bert with disentangled attention},
  author={He, Pengcheng and Liu, Xiaodong and Gao, Jianfeng and Chen, Weizhu},
  journal={arXiv preprint arXiv:2006.03654},
  year={2020}
}

@incollection{affelt2019spot,
  title={How to Spot Fake News},
  author={Affelt, Amy},
  booktitle={All That's Not Fit to Print},
  pages={57--84},
  year={2019},
  publisher={Emerald Publishing Limited}
}

@article{mohsen2024automated,
  title={Automated Detection of Misinformation: A Hybrid Approach for Fake News Detection},
  author={Mohsen, Fadi and Chaushi, Bedir and Abdelhaq, Hamed and Karastoyanova, Dimka and Wang, Kevin},
  journal={Future Internet},
  volume={16},
  number={10},
  pages={352},
  year={2024},
  publisher={MDPI}
}

@article{agarwal2019analysis,
  title={Analysis of classifiers for fake news detection},
  author={Agarwal, Vasu and Sultana, H Parveen and Malhotra, Srijan and Sarkar, Amitrajit},
  journal={Procedia Computer Science},
  volume={165},
  pages={377--383},
  year={2019},
  publisher={Elsevier}
}

@inproceedings{ahmed2017detection,
  title={Detection of online fake news using n-gram analysis and machine learning techniques},
  author={Ahmed, Hadeer and Traore, Issa and Saad, Sherif},
  booktitle={Intelligent, Secure, and Dependable Systems in Distributed and Cloud Environments: First International Conference, ISDDC 2017, Vancouver, BC, Canada, October 26-28, 2017, Proceedings 1},
  pages={127--138},
  year={2017},
  organization={Springer}
}

@article{ahmed2018detecting,
  title={Detecting opinion spams and fake news using text classification},
  author={Ahmed, Hadeer and Traore, Issa and Saad, Sherif},
  journal={Security and Privacy},
  volume={1},
  number={1},
  pages={e9},
  year={2018},
  publisher={Wiley Online Library}
}

@article{alam2021survey,
  title={A survey on multimodal disinformation detection},
  author={Alam, Firoj and Cresci, Stefano and Chakraborty, Tanmoy and Silvestri, Fabrizio and Dimitrov, Dimiter and Martino, Giovanni Da San and Shaar, Shaden and Firooz, Hamed and Nakov, Preslav},
  journal={arXiv preprint arXiv:2103.12541},
  year={2021}
}

@inproceedings{alhindi2018your,
  title={Where is your evidence: Improving fact-checking by justification modeling},
  author={Alhindi, Tariq and Petridis, Savvas and Muresan, Smaranda},
  booktitle={Proceedings of the first workshop on fact extraction and verification (FEVER)},
  pages={85--90},
  year={2018}
}

@article{allcott2017social,
  title={Social media and fake news in the 2016 election},
  author={Allcott, Hunt and Gentzkow, Matthew},
  journal={Journal of economic perspectives},
  volume={31},
  number={2},
  pages={211--236},
  year={2017},
  publisher={American Economic Association 2014 Broadway, Suite 305, Nashville, TN 37203-2418}
}

@article{bahad2019fake,
  title={Fake news detection using bi-directional LSTM-recurrent neural network},
  author={Bahad, Pritika and Saxena, Preeti and Kamal, Raj},
  journal={Procedia Computer Science},
  volume={165},
  pages={74--82},
  year={2019},
  publisher={Elsevier}
}

@inproceedings{bugueno2019empirical,
  title={An empirical analysis of rumor detection on microblogs with recurrent neural networks},
  author={Bugue{\~n}o, Margarita and Sepulveda, Gabriel and Mendoza, Marcelo},
  booktitle={Social Computing and Social Media. Design, Human Behavior and Analytics: 11th International Conference, SCSM 2019, Held as Part of the 21st HCI International Conference, HCII 2019, Orlando, FL, USA, July 26-31, 2019, Proceedings, Part I 21},
  pages={293--310},
  year={2019},
  organization={Springer}
}

@article{alghamdi2024power,
  title={The Power of Context: A Novel Hybrid Context-Aware Fake News Detection Approach},
  author={Alghamdi, Jawaher and Lin, Yuqing and Luo, Suhuai},
  journal={Information},
  volume={15},
  number={3},
  pages={122},
  year={2024},
  publisher={MDPI}
}

@article{alghamdi2024enhancing,
  title={Enhancing hierarchical attention networks with CNN and stylistic features for fake news detection},
  author={Alghamdi, Jawaher and Lin, Yuqing and Luo, Suhuai},
  journal={Expert Systems with Applications},
  volume={257},
  pages={125024},
  year={2024},
  publisher={Elsevier}
}

@article{alghamdi2022comparative,
  title={A comparative study of machine learning and deep learning techniques for fake news detection},
  author={Alghamdi, Jawaher and Lin, Yuqing and Luo, Suhuai},
  journal={Information},
  volume={13},
  number={12},
  pages={576},
  year={2022},
  publisher={MDPI}
}

@article{alghamdi2023towards,
  title={Towards COVID-19 fake news detection using transformer-based models},
  author={Alghamdi, Jawaher and Lin, Yuqing and Luo, Suhuai},
  journal={Knowledge-Based Systems},
  volume={274},
  pages={110642},
  year={2023},
  publisher={Elsevier}
}

@article{chen2023using,
  title={Using deep learning models to detect fake news about COVID-19},
  author={Chen, Mu-Yen and Lai, Yi-Wei and Lian, Jiunn-Woei},
  journal={ACM Transactions on Internet Technology},
  volume={23},
  number={2},
  pages={1--23},
  year={2023},
  publisher={ACM New York, NY}
}

@inproceedings{della2018automatic,
  title={Automatic online fake news detection combining content and social signals},
  author={Della Vedova, Marco L and Tacchini, Eugenio and Moret, Stefano and Ballarin, Gabriele and DiPierro, Massimo and De Alfaro, Luca},
  booktitle={2018 22nd conference of open innovations association (FRUCT)},
  pages={272--279},
  year={2018},
  organization={IEEE}
}

@article{devlin2018bert,
  title={Bert: Pre-training of deep bidirectional transformers for language understanding},
  author={Devlin, Jacob and Chang, Ming-Wei and Lee, Kenton and Toutanova, Kristina},
  journal={arXiv preprint arXiv:1810.04805},
  year={2018}
}

@article{dutta2019fake,
  title={Fake news prediction: a survey},
  author={Dutta, Pinky Saikia and Das, Meghasmita and Biswas, Sumedha and Bora, Mriganka and Saikia, Sankar Swami},
  journal={International Journal of Scientific Engineering and Science},
  volume={3},
  number={3},
  pages={1--3},
  year={2019}
}

@inproceedings{golbeck2018fake,
  title={Fake news vs satire: A dataset and analysis},
  author={Golbeck, Jennifer and Mauriello, Matthew and Auxier, Brooke and Bhanushali, Keval H and Bonk, Christopher and Bouzaghrane, Mohamed Amine and Buntain, Cody and Chanduka, Riya and Cheakalos, Paul and Everett, Jennine B and others},
  booktitle={Proceedings of the 10th ACM conference on web science},
  pages={17--21},
  year={2018}
}

@inproceedings{katsaros2019machine,
  title={Which machine learning paradigm for fake news detection?},
  author={Katsaros, Dimitrios and Stavropoulos, George and Papakostas, Dimitrios},
  booktitle={IEEE/WIC/ACM International Conference on Web Intelligence},
  pages={383--387},
  year={2019}
}

@article{hakak2021ensemble,
  title={An ensemble machine learning approach through effective feature extraction to classify fake news},
  author={Hakak, Saqib and Alazab, Mamoun and Khan, Suleman and Gadekallu, Thippa Reddy and Maddikunta, Praveen Kumar Reddy and Khan, Wazir Zada},
  journal={Future Generation Computer Systems},
  volume={117},
  pages={47--58},
  year={2021},
  publisher={Elsevier}
}

@article{sun2024exploring,
  title={Exploring the Deceptive Power of LLM-Generated Fake News: A Study of Real-World Detection Challenges},
  author={Sun, Yanshen and He, Jianfeng and Cui, Limeng and Lei, Shuo and Lu, Chang-Tien},
  journal={arXiv preprint arXiv:2403.18249},
  year={2024}
}

@article{jiang2021novel,
  title={A novel stacking approach for accurate detection of fake news},
  author={Jiang, TAO and Li, Jian Ping and Haq, Amin Ul and Saboor, Abdus and Ali, Amjad},
  journal={IEEE Access},
  volume={9},
  pages={22626--22639},
  year={2021},
  publisher={IEEE}
}

@inproceedings{salem2019fa,
  title={Fa-kes: A fake news dataset around the syrian war},
  author={Salem, Fatima K Abu and Al Feel, Roaa and Elbassuoni, Shady and Jaber, Mohamad and Farah, May},
  booktitle={Proceedings of the international AAAI conference on web and social media},
  volume={13},
  pages={573--582},
  year={2019}
}

@article{hamed2023disinformation,
  title={DISINFORMATION DETECTION ABOUT ISLAMIC ISSUES ON SOCIAL MEDIA USING DEEP LEARNING TECHNIQUES},
  author={Hamed, Suhaib Kh and Ab Aziz, Mohd Juzaiddin and Yaakub, Mohd Ridzwan},
  journal={Malaysian Journal of Computer Science},
  volume={36},
  number={3},
  pages={242--270},
  year={2023}
}

@article{kaliyar2021fakebert,
  title={FakeBERT: Fake news detection in social media with a BERT-based deep learning approach},
  author={Kaliyar, Rohit Kumar and Goswami, Anurag and Narang, Pratik},
  journal={Multimedia tools and applications},
  volume={80},
  number={8},
  pages={11765--11788},
  year={2021},
  publisher={Springer}
}

@article{hu2023bad,
  title={Bad actor, good advisor: Exploring the role of large language models in fake news detection},
  author={Hu, Beizhe and Sheng, Qiang and Cao, Juan and Shi, Yuhui and Li, Yang and Wang, Danding and Qi, Peng},
  journal={arXiv preprint arXiv:2309.12247},
  year={2023}
}

@article{rohera2022taxonomy,
  title={A taxonomy of fake news classification techniques: Survey and implementation aspects},
  author={Rohera, Dhiren and Shethna, Harshal and Patel, Keyur and Thakker, Urvish and Tanwar, Sudeep and Gupta, Rajesh and Hong, Wei-Chiang and Sharma, Ravi},
  journal={IEEE Access},
  volume={10},
  pages={30367--30394},
  year={2022},
  publisher={IEEE}
}

@article{liao2021integrated,
  title={An integrated multi-task model for fake news detection},
  author={Liao, Qing and Chai, Heyan and Han, Hao and Zhang, Xiang and Wang, Xuan and Xia, Wen and Ding, Ye},
  journal={IEEE Transactions on Knowledge and Data Engineering},
  volume={34},
  number={11},
  pages={5154--5165},
  year={2021},
  publisher={IEEE}
}

@article{mikolov2013efficient,
  title={Efficient estimation of word representations in vector space},
  author={Mikolov, Tomas and Chen, Kai and Corrado, Greg and Dean, Jeffrey},
  journal={arXiv preprint arXiv:1301.3781},
  year={2013}
}

@article{nakov2022overview,
  title={Overview of the CLEF-2022 CheckThat! lab task 2 on detecting previously fact-checked claims},
  author={Nakov, Preslav and Da San Martino, Giovanni and Alam, Firoj and Shaar, Shaden and Mubarak, Hamdy and Babulkov, Nikolay},
  year={2022}
}

@article{zhou2020survey,
  title={A survey of fake news: Fundamental theories, detection methods, and opportunities},
  author={Zhou, Xinyi and Zafarani, Reza},
  journal={ACM Computing Surveys (CSUR)},
  volume={53},
  number={5},
  pages={1--40},
  year={2020},
  publisher={ACM New York, NY, USA}
}

@article{zubiaga2018detection,
  title={Detection and resolution of rumours in social media: A survey},
  author={Zubiaga, Arkaitz and Aker, Ahmet and Bontcheva, Kalina and Liakata, Maria and Procter, Rob},
  journal={ACM Computing Surveys (CSUR)},
  volume={51},
  number={2},
  pages={1--36},
  year={2018},
  publisher={ACM New York, NY, USA}
}

@article{newman2013social,
  title={Social media in the changing ecology of news: The fourth and fifth estates in Britain},
  author={Newman, Nic and Dutton, William and Blank, Grant},
  journal={International journal of internet science},
  volume={7},
  number={1},
  year={2013},
  publisher={University of Konstanz}
}

@article{wang2017liar,
  title={" liar, liar pants on fire": A new benchmark dataset for fake news detection},
  author={Wang, William Yang},
  journal={arXiv preprint arXiv:1705.00648},
  year={2017}
}

@article{passaro2022context,
  title={In-context annotation of topic-oriented datasets of fake news: A case study on the notre-dame fire event},
  author={Passaro, Lucia C and Bondielli, Alessandro and Dell’Oglio, Pietro and Lenci, Alessandro and Marcelloni, Francesco},
  journal={Information Sciences},
  volume={615},
  pages={657--677},
  year={2022},
  publisher={Elsevier}
}

@inproceedings{patwa2021fighting,
  title={Fighting an infodemic: Covid-19 fake news dataset},
  author={Patwa, Parth and Sharma, Shivam and Pykl, Srinivas and Guptha, Vineeth and Kumari, Gitanjali and Akhtar, Md Shad and Ekbal, Asif and Das, Amitava and Chakraborty, Tanmoy},
  booktitle={Combating Online Hostile Posts in Regional Languages during Emergency Situation: First International Workshop, CONSTRAINT 2021, Collocated with AAAI 2021, Virtual Event, February 8, 2021, Revised Selected Papers 1},
  pages={21--29},
  year={2021},
  organization={Springer}
}

@article{perez2017automatic,
  title={Automatic detection of fake news},
  author={P{\'e}rez-Rosas, Ver{\'o}nica and Kleinberg, Bennett and Lefevre, Alexandra and Mihalcea, Rada},
  journal={arXiv preprint arXiv:1708.07104},
  year={2017}
}

@article{shu2017fake,
  title={Fake news detection on social media: A data mining perspective},
  author={Shu, Kai and Sliva, Amy and Wang, Suhang and Tang, Jiliang and Liu, Huan},
  journal={ACM SIGKDD explorations newsletter},
  volume={19},
  number={1},
  pages={22--36},
  year={2017},
  publisher={ACM New York, NY, USA}
}

@inproceedings{horne2017just,
  title={This just in: Fake news packs a lot in title, uses simpler, repetitive content in text body, more similar to satire than real news},
  author={Horne, Benjamin and Adali, Sibel},
  booktitle={Proceedings of the international AAAI conference on web and social media},
  volume={11},
  number={1},
  pages={759--766},
  year={2017}
}

@article{lewandowsky2017beyond,
  title={Beyond misinformation: Understanding and coping with the “post-truth” era},
  author={Lewandowsky, Stephan and Ecker, Ullrich KH and Cook, John},
  journal={Journal of applied research in memory and cognition},
  volume={6},
  number={4},
  pages={353--369},
  year={2017},
  publisher={Elsevier}
}

@article{touvron2023llama,
  title={Llama: Open and efficient foundation language models},
  author={Touvron, Hugo and Lavril, Thibaut and Izacard, Gautier and Martinet, Xavier and Lachaux, Marie-Anne and Lacroix, Timoth{\'e}e and Rozi{\`e}re, Baptiste and Goyal, Naman and Hambro, Eric and Azhar, Faisal and others},
  journal={arXiv preprint arXiv:2302.13971},
  year={2023}
}

@article{vieweg2010microblogged,
  title={Microblogged contributions to the emergency arena: Discovery, interpretation and implications},
  author={Vieweg, Sarah},
  journal={Computer Supported Collaborative Work},
  pages={515--516},
  year={2010},
  publisher={Citeseer}
}

@article{farhangian2024fake,
  title={Fake news detection: Taxonomy and comparative study},
  author={Farhangian, Faramarz and Cruz, Rafael MO and Cavalcanti, George DC},
  journal={Information Fusion},
  volume={103},
  pages={102140},
  year={2024},
  publisher={Elsevier}
}

\appendix 

\small 

\section{Data Preprocessing}\label{sec:appendix_dataset}

We report in Table \ref{tab:dataset_split}, for each data set, the provenance, the number of instances used in the train/valid/test split (dataset-specific experiment) for each data set after preprocessing and filtering, and the date the original dataset was downloaded.

\setcounter{table}{0}
\begin{table}[H]
\centering\scriptsize
\caption{Datasets used for the experiments.}
\label{tab:dataset_split}

\begin{tabular}{@{}l p{3.4 cm} c c c c@{}}

\toprule
\textbf{Dataset} & \textbf{Source} & \textbf{\# Training} & \textbf{\# Validation} & \textbf{\# Test} & \textbf{Date}\\
\midrule

Celebrity \citep{perez2017automatic} 
& \url{https://www.kaggle.com/datasets/sumanthvrao/fakenewsdataset} 
& 300 & 100 & 100 & 24/10/2023\\

Cidii \citep{hamed2023disinformation} 
& \url{https://www.kaggle.com/datasets/suhaibkhalil/cidii-dataset} 
& 433 & 144 & 145 & 14/11/2023\\

FakeVsSatire \citep{golbeck2018fake} 
& \url{https://github.com/jgolbeck/fakenews} 
& 291 & 97 & 98 & 24/10/2023\\

Fakes \citep{salem2019fa} 
& \url{https://zenodo.org/records/2607278}
& 482 & 161 & 161 & 24/10/2023\\

Horne \citep{horne2017just} 
& \url{https://github.com/rpitrust/fakenewsdata1} 
& 195 & 65 & 66 & 27/10/2023\\

Infodemic \citep{patwa2021fighting} 
& https://constraint-shared-task-2021.github.io/ 
& 6299 & 2139 & 2119 & 03/03/2024\\

Isot \citep{ahmed2017detection} 
& \url{https://onlineacademiccommunity.uvic.ca/isot/2022/11/27/fake-news-detection-datasets/}
& 23752 & 8388 & 8854 & 24/10/2023\\

LIAR-PLUS \citep{alhindi2018your} 
& \url{https://github.com/Tariq60/LIAR-PLUS} 
& 7660 & 2556 & 2557 & 24/10/2023\\

NDF \citep{passaro2022context} 
& \url{https://github.com/Unipisa/NDFDataset} 
& 332 & 111 & 111 & 24/10/2023\\

FakeNewsNet\citep{shu2017fake} \\(Politifact version)  
& \url{https://github.com/KaiDMML/FakeNewsNet} 
& 302 & 101 & 101 & 14/11/2023\\

\bottomrule
\end{tabular}

\end{table}

The LIAR-PLUS dataset is a six-class dataset. We have binarized it, including the classes ``True'', ``Mostly-True'', and ``Half-True'' in the Real class and the ``False'', ``Pants-Fire'', and ``Barely-True'' in the Fake class, as it is commonly done in several other works \citep{bahad2019fake}.

The NDF dataset is a binary-class dataset. 
It is annotated manually and with two different crowdsourcing experiments, producing three labelling variants, named Manually Labelling, Out-of-Context Labelling (OOC), and In-Context Labelling (IC). The IC variant has been made by humans during a crowdsourcing experiment, with the benefit of contextual information to guide their decisions. Annotators in the OOC experiments did not have additional information. For our research, we utilized the IC variant. 

The FakeVsSatire dataset presents a unique challenge. It includes satirical information in its real class. Satirical newspaper articles often share features with fake news, making this dataset particularly challenging.

The FakeNewsNet dataset is a comprehensive collection of fake and real articles from Politifact and Gossipcop. However,  
at the time we downloaded the texts from the links in the dataset, only data from Politifact were available.

Finally, as our study focuses exclusively on text-based fake news detection, we do not extract or model community-level or group-dynamic features (e.g., user interactions, propagation traces, or social-network structures). All experiments rely solely on the textual content and the dataset-provided labels.

\section{Parameter and prompt optimization}
\label{sec:appendix_optimization}
Before carrying out the experiments, we applied different optimization strategies depending on the model architecture. For most approaches, we searched for the best parameterization of each approach on each dataset by exploiting Optuna,\footnote{https://optuna.org/} an open source hyperparameter optimization framework to automate hyperparameter search. 
The details of this procedure are provided in  \ref{sect:appendix_opt_param}. 
For the MERMAID model, we adopted the hyperparameters as reported in the original paper. For the decoder-only models, we conducted a dedicated prompt optimization process. The goal of this process was to select a single, robust prompt for each model (e.g., one prompt for Llama3-8B, one for Qwen3-32B, and one for Zephyr-7B-beta). This selected prompt was the one that achieved the highest F1-score on the validation split across the largest number of datasets. This selection was performed independently for both the zero-shot and few-shot settings. Details of this procedure are provided in \ref{sect:appendix_opt_prompt}.

\subsection{Parameter Optimization}\label{sect:appendix_opt_param}

Table \ref{tab:parameter_optimization} details the parameters tuned for each algorithm and the range of values considered during optimization. 

\setcounter{table}{0}
\begin{table}[H]
\footnotesize
\centering
\caption{Values for parameters optimization.}\label{tab:parameter_optimization}
\begin{tabular}{@{}lcc@{}}
\toprule
Model                               & \textbf{Parameter}         & \textbf{Tested Range}                 \\ \toprule
\multirow{3}{*}{LR} & Penalty           & L2; None                    \\
                                    & C                 & 0.01; 0.1; 1.0; 10.0; 100.0 \\
                                    & Solver            & lbfgs; liblinear; sag; saga \\ \midrule
\multirow{4}{*}{SVM}                & Penalty           & L1; L2                      \\
                                    & C                 & 0.01; 0.1; 1.0; 10.0; 100.0 \\
                                    & Loss              & Hinge; Squared Hinge        \\
                                    & Dual              & True; False                 \\ \midrule
\multirow{2}{*}{NB}        & Alpha             & 0.01; 0.1; 1.0; 10.0; 100.0 \\
                                    & fit\_prior        & True; False                 \\ \midrule
\multirow{5}{*}{CNN}                & Filter size       & 3; 4; 5                     \\
                                    & Number of Filters & 16; 32; 64; 96; 128         \\
                                    & Dropout           & 0.2; 0.4; 0.6; 0.8          \\
                                    & Hidden Units      & 8; 16; 32; 64               \\
                                    & Learning Rate     & 1e-5; 1e-4; 1e-3; 1e-2      \\ \midrule
\multirow{4}{*}{BiLSTM}            & Number of Units   & 16; 32; 64; 96; 128         \\
                                    & Dropout           & 0.2; 0.4; 0.6; 0.8          \\
                                    & Hidden Units      & 8; 16; 32; 64               \\
                                    & Learning Rate     & 1e-5; 1e-4; 1e-3; 1e-2      \\ \midrule
\multirow{4}{*}{CNN-BERT}           & CNN Filters       & 64; 96; 128                 \\
                                    & Kernel Size       & 3; 4; 5                     \\
                                    & Dense Units       & 16; 32; 64                  \\
                                    & Learning Rate     & 1e-5; 1e-4; 1e-3; 1e-2      \\ \midrule
\multirow{2}{*}{BERT}               & Learning Rate     & 4e-5; 2e-5; 3e-2            \\
                                    & Weight Decay      & 0.001; 0.01; 0.1; None      \\ \midrule
\multirow{2}{*}{DeBERTa}               & Learning Rate     & 4e-5; 2e-5; 3e-2            \\
                                    & Weight Decay      & 0.001; 0.01; 0.1; None      \\ \bottomrule
\end{tabular}
\end{table}

For CNN, BiLSTM, CNN-BERT, and BERT, we set a default of 50 epochs and early stopping at 2 epochs to stop training if the validation loss 
does not decrease for two consecutive epochs. We used F1-score 
computed on the validation set to select the best model for each dataset. Due to computational constraints, we set the batch size to $8$. The other parameters were left to their default configuration.

Table \ref{tab:parameter_optimization_best} shows the best hyperparameter values obtained by each approach on each dataset.

\begin{table}[H]
\centering
\scriptsize
\caption{The best hyperparameter values after parameter optimization.}\label{tab:parameter_optimization_best}
\begin{tabularx}{\textwidth}{@{}p{1.2cm}*{11}{>{\centering\arraybackslash}X}@{}} 
\toprule
\fontsize{8pt}{8pt}\selectfont Model & \rotatebox{65}{\textbf{Parameters}} & \rotatebox{65}{\textbf{Celebrity}} & \rotatebox{65}{\textbf{Cidii}} & \rotatebox{65}{\textbf{FakeVsSatire}} & \rotatebox{65}{\textbf{Fakes}} & \rotatebox{65}{\textbf{Horne}} & \rotatebox{65}{\textbf{Infodemic}} & \rotatebox{65}{\textbf{Isot}} & \rotatebox{65}{\textbf{LIAR}} & \rotatebox{65}{\textbf{NDF}} & \rotatebox{65}{\textbf{Politifact}} \\
\midrule
\multirow{3}{*}{LR}
& Penalty & l2 & none & none & l2 & none & l2 & none & none & none & none \\
& C & 10.0 & 0.01 & 0.01 & 100.0 & 0.01 & 100.0 & 100.0 & 1.0 & 0.01 & 0.01 \\
& Solver & lbfgs & lbfgs & lbfgs & lbfgs & lbfgs & lbfgs & lbfgs & lbfgs & lbfgs & lbfgs \\
\midrule
\multirow{4}{*}{SVM} & Penalty & l1 & l1 & l1 & l1 & l1 & l1 & l1 & l1 & l1 & l1 \\
& C & 1.0 & 100.0 & 10.0 & 100.0 & 100.0 & 1.0 & 1.0 & 0.1 & 100.0 & 10.0 \\
& Loss & sh & sh & sh & sh & sh & sh & sh & sh & sh & sh \\
& Dual & 0 & 0 & 0 & 0 & 0 & 0 & 0 & 0 & 0 & 0 \\
\midrule
\multirow{2}{*}{NB}
& Alpha & 1.0 & 0.1 & 0.01 & 100.0 & 0.1 & 0.1 & 0.01 & 1.0 & 0.1 & 0.01 \\
& fit-prior & 0 & 0 & 0 & 1 & 0 & 0 & 1 & 0 & 1 & 0 \\
\midrule
\multirow{5}{*}{CNN}
& F.size & 3 & 3 & 4 & 5 & 5 & 4 & 4 & 4 & 4 & 5 \\
& Filters & 96 & 96 & 96 & 64 & 96 & 128 & 96 & 128 & 128 & 32 \\
& Dropout & 0.6 & 0.6 & 0.2 & 0.4 & 0.2 & 0.4 & 0.1 & 0.2 & 0.4 & 0.2 \\
& H.Units & 8 & 8 & 8 & 8 & 32 & 64 & 64 & 16 & 32 & 16 \\
& LR & 1e-05 & 1e-05 & 0.001 & 1e-05 & 0.001 & 1e-05 & 0.001 & 1e-06 & 0.0001 & 0.0001 \\
& Epochs & 11 & 50 & 16 & 13 & 6 & 7 & 4 & 50 & 9 & 26 \\
\midrule
\multirow{4}{*}{\fontsize{8pt}{8pt}\selectfont BiLSTM}
& Units & 32 & 96 & 128 & 96 & 16 & 16 & 96 & 128 & 96 & 16 \\
& Dropout & 0.2 & 0.1 & 0.8 & 0.6 & 0.2 & 0.8 & 0.1 & 0.6 & 0.6 & 0.4 \\
& H.Units & 8 & 16 & 32 & 16 & 8 & 8 & 64 & 64 & 16 & 16 \\
& LR & 1e-06 & 0.0001 & 1e-06 & 1e-05 & 1e-05 & 0.001 & 0.001 & 0.001 & 0.0001 \\
& Epochs & 3 & 5 & 35 & 20 & 50 & 3 & 5 & 6 & 4 & 50 \\
\midrule
\multirow{4}{*}{CNN-BERT}
& Filters & 96 & 128 & 64 & 128 & 128 & 128 & 128 & 64 & 128 & 128 \\
& K.Size & 4 & 5 & 5 & 3 & 5 & 5 & 3 & 3 & 5 & 5 \\
& D.Units & 64 & 16 & 16 & 32 & 16 & 16 & 64 & 32 & 32 & 32 \\
& LR & 0.001 & 1e-06 & 1e-06 & 1e-05 & 0.001 & 1e-05 & 0.0001 & 1e-05 & 1e-05 & 0.001 \\
& Epochs & 5 & 29 & 32 & 32 & 8 & 7 & 7 & 7 & 30 & 5 \\
\midrule
\multirow{2}{*}{BERT} & LR & 2e-5 & 2e-5 & 2e-5 & 2e-5 & 2e-5 & 4e-5 & 2e-5 & 2e-5 & 2e-5 & 4e-5 \\
& W.Decay & 0.1 & 0.1 & 0.01 & 0.001 & 0.1 & 0.1 & 0 & 0.001 & 0.01 & 0.1 \\
& Epochs & 3 & 5 & 3 & 3 & 5 & 5 & 5 & 5 & 2 & 2 \\
\midrule
\multirow{2}{*}{DeBERTa} & LR & 2e-5 & 2e-5 & 2e-5 & 2e-5 & 2e-5 & 2e-5 & 2e-5 & 2e-5 & 2e-5 & 4e-5 \\
& W.Decay & 0.01 & 0 & 0.001 & 0 & 0.001 & 0.01 & 0 & 0.001 & 0.01 & 0 \\
& Epochs & 4 & 5 & 4 & 5 & 4 & 4 & 4 & 4 & 5 & 5 \\
\bottomrule
\end{tabularx}
\end{table}

\subsection{Sensitivity studies for neural models}\label{sect:appendix_seed_neural}

Neural models inherently exhibit variability in performance due to factors such as random initialization and stochastic optimization.
To assess the robustness of the reported results in Section \ref{sect:exp1}, we reported a short seed-sensitivity study and a batch-sensitivity study focusing exclusively on the neural models trained from scratch or fine-tuned (DL and Transformers-based models).

In the case of the seed-sensitivity study, each model was trained five times using five random seeds. For every dataset–model pair, we report the mean F1-score ± standard deviation computed across these independent runs (Table \ref{tab:seed_experiment}).

\begin{table}[H]
\tiny
\caption{Average F1-score  and standard deviation obtained by the neural models across 5 random seeds.}
\label{tab:seed_experiment}
\setlength{\tabcolsep}{3pt}   
\begin{tabularx}{\textwidth}{lcccccccccc}
\toprule
\textbf{Model} &
\rotatebox{90}{\textbf{Celebrity}} &
\rotatebox{90}{\textbf{Cidii}} &
\rotatebox{90}{\textbf{FakeVsSatire}} &
\rotatebox{90}{\textbf{Fakes}} &
\rotatebox{90}{\textbf{Horne}} &
\rotatebox{90}{\textbf{Infodemic}} &
\rotatebox{90}{\textbf{Isot}} &
\rotatebox{90}{\textbf{LIAR-PLUS}} &
\rotatebox{90}{\textbf{NDF}} &
\rotatebox{90}{\textbf{Politifact}} \\
\midrule

{CNN} &
$.52\pm.02$ &
$.58\pm.00$ &
$.61\pm.03$ &
$.53\pm.03$ &
$.61\pm.02$ &
$.53\pm.01$ &
$.98\pm.00$ &
$.63\pm.03$ &
$.61\pm.03$ &
$.69\pm.02$ \\

{BiLSTM} &
$.50\pm.01$ &
$.59\pm.02$ &
$.55\pm.03$ &
$.50\pm.01$ &
$.62\pm.00$ &
$.67\pm.02$ &
$.97\pm.00$ &
$.55\pm.01$ &
$.62\pm.01$ &
$.68\pm.03$ \\

{CNN-BERT} &
$.51\pm.01$ &
$.59\pm.00$ &
$.58\pm.00$ &
$.52\pm.01$ &
$.66\pm.04$ &
$.72\pm.02$ &
$.99\pm.00$ &
$.63\pm.02$ &
$.62\pm.01$ &
$.73\pm.06$ \\
\midrule
{BERT} &
$.71\pm.03$ &
$.98\pm.00$ &
$.72\pm.02$ &
$.49\pm.03$ &
$.74\pm.06$ &
$.97\pm.00$ &
$1.0\pm.00$ &
$.65\pm.02$ &
$.82\pm.01$ &
$.85\pm.01$ \\

{DeBERTa} &
$.78\pm.02$ &
$.96\pm.00$ &
$.82\pm.04$ &
$.50\pm.02$ &
$.89\pm.05$ &
$.97\pm.00$ &
$1.0\pm.00$ &
$.65\pm.01$ &
$.92\pm.02$ &
$.86\pm.00$ \\

\bottomrule
\end{tabularx}
\end{table}

Overall, most architectures show low variance across runs, with several models (e.g., CNN-BERT on Cidii and Isot, BERT/DeBERTa on Infodemic) displaying near‑deterministic behavior. A few dataset–model pairs exhibit higher variability (e.g., CNN-BERT on Horne and Politifact), but even in these cases the variance remains small relative to the performance gaps between models, and the overall trends remain consistent.

To further examine the robustness of our findings, we conducted an additional batch‑sensitivity analysis on two datasets, Celebrity and Fakes. These datasets were selected because they capture two contrasting conditions that can stress different aspects of model stability. Celebrity is comparatively more regular and balanced in its class distribution, whereas Fakes is more heterogeneous. Evaluating both allows us to probe model behavior in settings that differ in complexity, without requiring a full sensitivity analysis across all datasets. Results are reported in Table \ref{tab:sensitivity_compact}

\begin{table}[t]
\centering
\small
\caption{Sensitivity analysis across batch sizes for the \textit{Celebrity} and \textit{Fakes} datasets.}
\begin{tabular}{llc}
\toprule
\textbf{Dataset} & \textbf{Model} & \textbf{F1 across batch sizes (bs8 → bs256)} \\
\midrule
 & BERT & 0.75, 0.69, 0.67, 0.64, 0.61, 0.67 \\
 & DeBERTa  & 0.80, 0.77, 0.75, 0.63, 0.72, 0.52 \\
Celebrity & CNN      & 0.54, 0.50, 0.50, 0.50, 0.50, 0.50 \\
 & BiLSTM   & 0.50, 0.50, 0.50, 0.50, 0.50, 0.50 \\
 & CNN-BERT & 0.53, 0.52, 0.53, 0.53, 0.50, 0.50 \\
\midrule
 & BERT & 0.52, 0.58, 0.47, 0.55, 0.52, 0.53 \\
 & DeBERTa  & 0.52, 0.52, 0.48, 0.52, 0.52, 0.52 \\
Fakes & CNN      & 0.46, 0.46, 0.47, 0.47, 0.47, 0.47 \\
 & BiLSTM   & 0.50, 0.49, 0.49, 0.49, 0.50, 0.49 \\
 & CNN-BERT & 0.52, 0.53, 0.54, 0.51, 0.52, 0.48 \\
\bottomrule
\end{tabular}
\label{tab:sensitivity_compact}
\end{table}

Overall, the results indicate that varying the batch size does not substantially alter the relative ordering of the models. CNN and BiLSTM remain consistently weak across settings, while Transformer‑based models preserve their performance advantage even under substantial changes in batch size.

\subsection{Mixed-Training/Single-Test imbalanced variant}\label{sect:mixed_imbalanced_variant}

To more closely reflect realistic multi-domain learning scenarios, we discuss an additional mixed-training experimental variant with a larger and imbalanced training set that mirrors the natural distribution of the ten datasets. Rather than sampling a fixed number of instances per domain, we draw 40\% of each dataset, using stratified sampling to preserve label proportions. This procedure yields a larger and imbalanced training set in which high-resource domains dominate the training signal. The Global test set contains 10\% of each dataset, while the remaining instances form the per‑dataset test sets. As a consequence, the test splits are not identical to those used in the balanced 2,000‑instance experiment, and the two regimes should be interpreted as complementary rather than strictly comparable.

\begin{table}[H]
\scriptsize
\caption{F1‑scores obtained by all models in the imbalanced mixed‑training/single‑test setting. Best F1-score for each dataset in bold.}
\label{tab:all_f1_models_by_dataset_mixed_imbalanced}
\begin{tabularx}{\textwidth}{@{}l*{11}{>{\centering\arraybackslash}X}@{}}
\toprule
\textbf{Model} &
\rotatebox{90}{\textbf{Celebrity}} &
\rotatebox{90}{\textbf{Cidii}} &
\rotatebox{90}{\textbf{FakeVsSatire}} &
\rotatebox{90}{\textbf{Fakes}} &
\rotatebox{90}{\textbf{Horne}} &
\rotatebox{90}{\textbf{Infodemic}} &
\rotatebox{90}{\textbf{Isot}} &
\rotatebox{90}{\textbf{LIAR-PLUS}} &
\rotatebox{90}{\textbf{NDF}} &
\rotatebox{90}{\textbf{Politifact}} &
\rotatebox{90}{\textbf{Global}} \\
\midrule
{LR}                  & .54          & .65          & .61          & .51          & .60          & .82          & .83          & .53          & .71          & .57          & .76 \\ {SVM}                 & .48          & .77          & .65          & \textbf{.53}          & .58          & .84          & .95         & .54          & .67          & .56          & .83 \\ {NB}                  & .33          & .77          & .46          & .34          & .55          & .85          & .95          & .55          & .65          & .56          & .85 \\ \midrule {CNN}                 & .58          & .79          & .52          & .48          & .60          & .47          & .95          & .57          & .63          & .62          & .87 \\ {BiLSTM}              & .50          & .73          & .56          & .49          & .57          & .41          & .95          & .58          & .71          & .60          & .86 \\ {CNN-BERT}            & .52          & .71          & .58          & .46          & .60          & .85          & .96          & .57          & .61          & .57          & .86 \\ \midrule {BERT}                & .70          & .96          & .70          & \textbf{.53}          & .78          & \textbf{.96}          & \textbf{.99}          & .59          & \textbf{.85}          & .77          & .91 \\ {DeBERTa}             & .83          & \textbf{.97}          & .76          & .49          & \textbf{.84}          & \textbf{.96}          & \textbf{.99}         & .59          & .83          & \textbf{.80}          & \textbf{.92} \\

{MERMAID}             & .77          & .91          & \textbf{.81}          & \textbf{.53}          & .57          & .72          & .94          & .56          & .66          & .77          & .84 \\
\midrule
{LLaMa3-8B (ZS)}      & .80 & .93 & .58 & .50 & .70          & .77          & .58          & .57          & .69          & .71          & .62  \\
{LLaMa3-8B (FS)}      & \textbf{.86}          & .93          & .59          & .51          & .68          & .80          & .78          & .60          & .77          & .73          & .76 \\
{Qwen3-32B (ZS)}      & .83 & .94 & .58 & .51 & .68          & .83          & .72          & .59          & .79          & .79          & .73 \\
{Qwen3-32B (FS)}      & .81          & .92          & .58          & .50          & .70          & .84          & .73          & .60          & .80          & .78          & .72 \\
{Zephyr-7B-beta (ZS)} & .68 & .79 & .46 & .50 & .55          & .74          & .59          & .55          & .70          & .58          & .59  \\
{Zephyr-7B-beta (FS)}  & .73          & .90          & .43          & .51          & .57          & .77          & .66          & \textbf{.61}          & .77          & .55          & .66 \\
\bottomrule
\end{tabularx}
\end{table}

Table \ref{tab:all_f1_models_by_dataset_mixed_imbalanced} reports the results obtained under this more realistic configuration. Several patterns emerge when compared with the balanced setup (Table \ref{tab:all_f1_models_by_dataset_mixed}). First, classical machine‑learning models (LR, SVM, NB) benefit markedly from the increased training volume, achieving substantially higher F1 in several datasets. DL models trained from scratch (CNN, BiLSTM, CNN‑BERT) exhibit a similar trend, driven by improved performance on large datasets such as Isot. Pre‑trained transformers (BERT, DeBERTa) remain the strongest models across both regimes, indicating that large pre‑trained encoders are particularly robust to domain imbalance and can effectively exploit additional training data. MERMAID also improves, showing more stable multi‑domain behavior than classical models but still under-performing compared to pre‑trained transformers. Decoder‑only LLMs exhibit more heterogeneous behavior: zero‑shot performance remains stable, while few‑shot prompting benefits moderately from the increased data volume, especially for smaller models like Zephyr, while for Qwen-32 the results are consistent with those obtained previously.

\subsection{Leave-One-Dataset-Out satire-sensitivity analysis}\label{sect:lodo_satire_analysis}

To evaluate the robustness of our findings to alternative label harmonization choices, we conduct a sensitivity analysis in the leave-one-dataset-out setting comparing model performance when satire is included as the \textit{real} class in FakeVsSatire and Horne datasets versus when those datasets are excluded from the training data. For each model $m$ and dataset $d$, we compute
\[
\Delta_{m,d} = \text{F1}_{\text{no satire}} - \text{F1}_{\text{satire}},
\]
so that positive values indicate improved performance when the two datasets are removed from training set, while negative values indicate a performance drop.

Table~\ref{tab:delta-satire} reports the F1-scores obtained by all models on each dataset in the Leave-One-Dataset-Out
setting, when satire is removed from training set and the corresponding $\Delta$ values (excluding FakeVsSatire and Horne).

\begin{table}[H]
\setlength{\tabcolsep}{3pt}
\tiny
\caption{F1-scores obtained by all models on each dataset in the Leave-One-Dataset-Out setting, when satire is removed from training set. The decoder-only models are considered both in zero-shot (ZS) and few-shot (FS) scenarios. The best F1-score for each dataset is highlighted in bold. In brackets $\Delta = \text{F1}_{\text{no satire}} - \text{F1}_{\text{satire}}$.}
\label{tab:delta-satire}
\begin{tabularx}{\textwidth}{@{}l*{8}{>{\centering\arraybackslash}X}@{}}
\toprule
\textbf{Model} &
\rotatebox{90}{\textbf{Celebrity}} &
\rotatebox{90}{\textbf{Cidii}} &
\rotatebox{90}{\textbf{Fakes}} &
\rotatebox{90}{\textbf{Infodemic}} &
\rotatebox{90}{\textbf{Isot}} &
\rotatebox{90}{\textbf{LIAR-PLUS}} &
\rotatebox{90}{\textbf{NDF}} &
\rotatebox{90}{\textbf{Politifact}} 
\\
\midrule
LR                  & .52 (-.05) & .51 (.05) & .52 (.04) & .53 (.01) & .52 (-.21) & .50 (.05) & .57 (.15) & .51 (-.10) \\
SVM                 & .51 (.00) & .52 (.01) & .51 (-.02) & .53 (.01) & .53 (-.02) & .46 (.00) & .49 (-.03) & .46 (-.01) \\
NB                  & .50 (.00) & .43 (.00) & .53 (.00) & .57 (.02) & .60 (-.01) & .53 (.00) & .39 (.00) & .53 (.01) \\ \midrule
CNN                 & .58 (.08) & .55 (.04) & .49 (-.04) & .49 (-.03) & .55 (.00) & .47 (.01) & .39 (.00) & .53 (.00) \\
BiLSTM              & .52 (.01) & .43 (-.08) & .47 (-.06) & .52 (.00) & .57 (.02) & .45 (-.01) & .52 (.00) & .57 (.10) \\
CNN-BERT            & .55 (.00) & .52 (.07) & .50 (-.01) & .50 (.06) & .49 (.00) & .47 (.00) & .44 (.06) & .58 (.01) \\ \midrule
BERT                & .55 (-.02) & .42 (-.04) & .50 (.02) & .55 (.03) & .61 (-.12) & .45 (.00) & .44 (.02) & .53 (-.08) \\
DeBERTa             & .56 (-.09) & .57 (.13) & .50 (-.01) & .64 (.04) & .80 (-.03) & .46 (.01) & .45 (-.02) & .65 (-.02) \\
MERMAID             & .54 (-.15) & .60 (.13) & .47 (-.07) & .48 (.00) & \textbf{.82} (.06) & .44 (-.03) & .52 (-.14) & .73 (-.08) \\ \midrule
LLaMa3-8B (ZS)      & \textbf{.86} (.00) & .90 (.00) & .52 (.00) & .78 (.00) & .74 (.00) & \textbf{.62} (.00) & .79 (.00) & .54 (.00) \\
LLaMa3-8B (FS)      & .85 (.00) & .89 (-.01) & .53 (.01) & .80 (-.02) & .66 (.03) & .61 (.01) & \textbf{.85} (.12) & .69 (-.04) \\
Qwen3-32B (ZS)      & .82 (.00) & \textbf{.92} (.00) & \textbf{.57} (.00) & \textbf{.83} (.00) & .75 (.00) & .58 (.00) & .83 (.00) & \textbf{.79} (.00) \\
Qwen3-32B (FS)      & .80 (-.01) & \textbf{.92} (.07) & .52 (-.06) & \textbf{.83} (.01) & .67 (-.09) & .59 (.02) & .83 (.02) & .78 (-.01) \\
Zephyr-7B-beta (ZS) & .73 (.00) & .84 (.00) & .53 (.00) & .75 (.00) & .61 (.00) & .62 (.00) & .78 (.00) & .60 (.00) \\
Zephyr-7B-beta (FS) & .76 (-.01) & .89 (.17) & .53 (.02) & .72 (-.12) & .57 (-.03) & .62 (.00) & .82 (.04) & .62 (-.03) \\
\bottomrule
\end{tabularx}
\end{table}

Overall, the results are consistent with what we observe when satirical data are included: LLMs continue to achieve the strongest performance, followed by fine-tuned Transformers, while traditional ML and DL models remain at the bottom of the ranking.
Nevertheless, we observe some variability in the results. Although the average performance change ($\Delta$), both per-model and per-dataset, is generally small, variance across specific model–dataset pairs can be larger. For example, the LR model loses 0.2 on Isot but gains 0.15 on NDF; on the NDF dataset some models improve by up to 12\%, while others lose up to 14\%. However, these cases represent outliers rather than systematic shifts. In most settings, performance discrepancies are much more modest, and average deviations are in ranges closer to zero. We also note that these discrepancies may be attributed to the changes in the training data (i.e., lack of around 800 data points and consequent change in composition of data batches).

While these results highlight the need for further investigation into label semantics (as discussed in Sections \ref{sec:limitations} and \ref{sec:conclusions}), they do not materially affect the overall conclusions of the paper.

\subsection{Prompt Optimization}\label{sect:appendix_opt_prompt}

For the decoder-only models (Llama3-8B, Qwen3-32B, and Zephyr-7B-beta), we performed a prompt optimization process separately for the zero-shot and few-shot settings. We evaluated three distinct prompts of increasing complexity for each setting with the goal of identifying the single, most robust prompt for each model across all datasets. We use deterministic decoding settings (temperature = 0, top\_k=1.0.).
 The selected prompt was the one that achieved the highest F1-score on the validation split for the largest number of datasets.

The three prompts evaluated for the zero-shot setting (P-Z1, P-Z2, P-Z3) are shown below. The \textbf{text} placeholder was replaced with the content to be labelled. The model was expected to return only the single word ``Real'' or ``Fake''.

\begin{tcolorbox}[
        colback=gray!5!white, colframe=gray!80!black, title=P-Z1,
    ]
    Classify the following text as 'Fake' or 'Real'. 

Now, label the following text.

Text: \textbf{\{ text \}}

Your label:

\end{tcolorbox}

\begin{tcolorbox}[
        colback=gray!5!white, colframe=gray!80!black, title=P-Z2,
    ]
    You have to act as a disinformation detector.

You will be provided with a piece of text (a tweet or a news article).

Your task is to label the text as 'Fake' or 'Real'.

Now, label the following text.

Text: \textbf{\{ text \}}

Your label:

\end{tcolorbox}

\begin{tcolorbox}[
        colback=gray!5!white, colframe=gray!80!black, title=P-Z3,
    ]
    You have to act as a disinformation detector. 
    
You will be provided with a piece of text (a tweet or a news article).

Read the text carefully and analyze its claims, sources, and tone. 

Your task is to label the text as 'Fake' or 'Real'.

'Real' news is factually accurate, verifiable, and presented objectively. 

'Fake' news contains verifiably false information, is intentionally misleading, or lacks credible sources.

Now, label the following text.

Text: \textbf{\{ text \}}

Your label:

\end{tcolorbox}

For the few-shot setting, we provided the models with 6 examples: 3 labeled ``Real'' and 3 labeled ``Fake''. These examples were sampled once per dataset from its training split and kept fixed across all few‑shot experiments for that dataset, ensuring consistency. For all few-shot experiments on a given dataset, the exact same 6 examples were always used in the prompt. The three prompts (P-Z1, P-Z2, P-Z3) followed a similar logic to their zero-shot counterparts, but included placeholders for the 6 examples.

Table \ref{tab:prompt_optimization} lists the prompts selected for each model.

\begin{table}[H]
\footnotesize
\centering
\caption{Best prompt selected after prompt optimization.}\label{tab:prompt_optimization}
\begin{tabular}{@{}lcc@{}}
\toprule
\textbf{Model}                               & \textbf{Modality}         & \textbf{Prompt}                 \\ \toprule
Llama3-8B & Zero-shot  & P-Z3\\ 
Qwen3-32B & Zero-shot  & P-Z3\\ 
Zephyr-7B-beta & Zero-shot  & P-Z1\\ 
\midrule
Llama3-8B & Few-shot  & P-Z3\\ 
Qwen3-32B & Few-shot  & P-Z3\\ 
Zephyr-7B-beta & Few-shot  & P-Z2   \\ \bottomrule
\end{tabular}
\end{table}

A general overview of all LLM settings is provided in Table \ref{tab:llm_summary}.

\newcolumntype{L}{>{\raggedright\arraybackslash}X}

\begin{table}[H]
\centering
\scriptsize
\caption{Summary of experimental setups across all LLM settings.}
\label{tab:llm_summary}
\begin{tabularx}{\textwidth}{@{}l l L L c@{}}
\toprule
\textbf{Model \& Setting} & \textbf{Prompt} & \textbf{Decoding Settings} & \textbf{Few-shot Sampling} & \textbf{Selection seed} \\
\midrule
Llama3-8B (Zero-shot) & P-Z3 & Temp = 0; Top-k = 1; Max new tokens = 64 & N/A & N/A \\
\addlinespace[2pt]
Llama3-8B (Few-shot)  & P-Z3 & Temp = 0; Top-k = 1; Max new tokens = 64 & 6 examples (3 per class)\textsuperscript{*} & 42 \\
\midrule
Qwen3-32B (Zero-shot) & P-Z3 & Temp = 0; Top-k = 1; Max new tokens = 64\textsuperscript{†} & N/A & N/A \\
\addlinespace[2pt]
Qwen3-32B (Few-shot)  & P-Z3 & Temp = 0; Top-k = 1; Max new tokens = 64\textsuperscript{†} & 6 examples (3 per class)\textsuperscript{*} & 42 \\
\midrule
Zephyr-7B (Zero-shot) & P-Z1 & Temp = 0; Top-k = 1; Max new tokens = 64 & N/A & N/A \\
\addlinespace[2pt]
Zephyr-7B (Few-shot)  & P-Z2 & Temp = 0; Top-k = 1; Max new tokens = 64 & 6 examples (3 per class)\textsuperscript{*} & 42 \\
\bottomrule
\addlinespace[4pt]
\multicolumn{5}{l}{\textsuperscript{*} \textit{Stratified sampling from training split, fixed per dataset.}} \\
\multicolumn{5}{l}{\textsuperscript{†} \textit{Global setting: enable\_thinking = False.}} \\
\end{tabularx}
\end{table}

\subsection{Prompt Calibration for the Leave-One-Dataset-Out Experiments}\label{sect:appendix_few_shot_calibration}

Few-shot prompting introduces an additional source of variability due to the random selection of exemplars. This effect is amplified in the leave-one-dataset-out setting, where exemplars are drawn from a large and heterogeneous pool of training datasets. To quantify the stability of the decoder-only LLMs, we repeat each evaluation three times using independently sampled few-shot exemplar sets.

Tables \ref{tab:seed_experiment} and \ref{tab:seed_experiment_median_iqr} report the variability of few-shot performance  across three runs for every held-out dataset. The first table presents the average F1-score and the corresponding standard deviation, while the second reports the median and the IQR of the three runs. The results show that some models (e.g., Llama3-8B) remain relatively stable across different exemplar draws, whereas others (e.g., Qwen3-32B on \textsc{Cidii}) exhibit substantial variance. This analysis highlights that few-shot performance can be sensitive to exemplar choice and that reporting a single run may overestimate the robustness of a model’s behavior.

\begin{table}[H]
\tiny
\caption{Average F1-score and standard deviation across three runs with different randomly sampled few-shot exemplar sets. Results are computed under the leave-one-dataset-out setting, where each model is evaluated on a held-out dataset while being prompted with exemplars drawn from the remaining datasets.}
\label{tab:seed_experiment}
\setlength{\tabcolsep}{1.8pt}
\begin{tabularx}{\textwidth}{lcccccccccc}
\toprule
\textbf{Model} &
\rotatebox{90}{\textbf{Celebrity}} &
\rotatebox{90}{\textbf{Cidii}} &
\rotatebox{90}{\textbf{FakeVsSatire}} &
\rotatebox{90}{\textbf{Fakes}} &
\rotatebox{90}{\textbf{Horne}} &
\rotatebox{90}{\textbf{Infodemic}} &
\rotatebox{90}{\textbf{Isot}} &
\rotatebox{90}{\textbf{LIAR-PLUS}} &
\rotatebox{90}{\textbf{NDF}} &
\rotatebox{90}{\textbf{Politifact}} \\
\midrule
{Llama3-8B (FS)} &
$.85 \pm .01$ &
$.88 \pm .02$ &
$.57 \pm .00$ &
$.52 \pm .00$ &
$.73 \pm .00$ &
$.81 \pm .02$ &
$.68 \pm .02$ &
$.62 \pm .00$ &
$.81 \pm .00$ &
$.74 \pm .01$ \\

{Qwen3-32B (FS)} &
$.81 \pm .00$ &
$.82 \pm .14$ &
$.58 \pm .00$ &
$.53 \pm .00$ &
$.73 \pm .00$ &
$.83 \pm .00$ &
$.71 \pm .01$ &
$.63 \pm .00$ &
$.83 \pm .01$ &
$.80 \pm .01$ \\

{Zephyr-7B-beta (FS)} &
$.66 \pm .02$ &
$.89 \pm .01$ &
$.41 \pm .00$ &
$.53 \pm .00$ &
$.58 \pm .05$ &
$.75 \pm .03$ &
$.59 \pm .00$ &
$.62 \pm .00$ &
$.79 \pm .03$ &
$.59 \pm .02$ \\

\bottomrule
\end{tabularx}
\end{table}

\begin{table}[H]
\scriptsize
\caption{Median and IQR of F1-scores across three runs with different randomly sampled few-shot exemplar sets. Results are computed under the leave-one-dataset-out setting, where each model is evaluated on a held-out dataset while being prompted with exemplars drawn from the remaining datasets.}
\label{tab:seed_experiment_median_iqr}
\setlength{\tabcolsep}{9pt}
\begin{tabularx}{\textwidth}{lcccccccccc}
\toprule
\textbf{Model} &
\rotatebox{90}{\textbf{Celebrity}} &
\rotatebox{90}{\textbf{Cidii}} &
\rotatebox{90}{\textbf{FakeVsSatire}} &
\rotatebox{90}{\textbf{Fakes}} &
\rotatebox{90}{\textbf{Horne}} &
\rotatebox{90}{\textbf{Infodemic}} &
\rotatebox{90}{\textbf{Isot}} &
\rotatebox{90}{\textbf{LIAR-PLUS}} &
\rotatebox{90}{\textbf{NDF}} &
\rotatebox{90}{\textbf{Politifact}} \\
\midrule

\textbf{LLaMA3-8B (FS)} \\
Median &
.84 & .87 & .57 & .52 & .72 & .82 & .67 & .62 & .81 & .74 \\
IQR &
.02 & .03 & .01 & .01 & .01 & .04 & .03 & .00 & .01 & .02 \\

\midrule

\textbf{Qwen3-32B (FS)} \\
Median &
.80 & .92 & .58 & .53 & .73 & .83 & .71 & .63 & .84 & .80 \\
IQR &
.02 & .16 & .00 & .01 & .01 & .00 & .02 & .01 & .02 & .02 \\

\midrule

\textbf{Zephyr-7B-beta (FS)} \\
Median &
.66 & .90 & .41 & .53 & .61 & .73 & .59 & .62 & .75 & .57 \\
IQR &
.03 & .02 & .01 & .00 & .06 & .04 & .01 & .01 & .05 & .05 \\

\bottomrule
\end{tabularx}
\end{table}

\FloatBarrier

\section{DeBERTa in-domain error analysis}\label{sec:error_analysis}
We conducted a qualitative error analysis to better understand the behavior of the best-performing model (DeBERTa) and to characterize the nature of its errors. We focus on Celebrity, FakeVsSatire, Fakes, Infodemic, NDF, and LIAR-PLUS, as these datasets allow us to identify three common failure modes: (i) \textbf{genre and structural imitation}, where the model is misled by formal reporting style or narrative complexity; (ii) \textbf{social and numerical shortcuts}, where metadata density or statistical claims act as proxies for falsity; and (iii) \textbf{sentiment artifacts}, where emotive language is confused with misinformation.

\subsection{Genre and structural imitation}
A failure mode involves the model's inability to look past the "form" of the article. In FakeVsSatire, satirical articles (labeled Real) are misclassified as Fake (Table \ref{tab:fake_satire_err}) because they adopt the stylistic cues of sensationalist misinformation. Conversely, in the Celebrity and Fakes datasets, the model fails to detect fabricated content because it mimics the structure of legitimate reporting. In Celebrity (Table \ref{tab:celebrity_err}), gossip pieces about real-world entities (e.g., Ariana Grande) use professional journalistic framing that deceive the model. In the Fakes dataset (Table \ref{tab:fakes_err}), which focuses on the Syrian war, the model systematically misclassifies fabricated reports of military operations or massacres as Real news. These samples often mention specific locations (e.g., Palmyra, Aleppo), quote official-sounding sources, and maintain a neutral, reporting tone, thereby reinforcing the illusion of credibility.

\begin{table}[ht]
\setcounter{table}{0}
\centering
\scriptsize
\caption{Examples of misclassified instances from the FakeVsSatire dataset. Confidence is the probability of the predicted class.}
\label{tab:fake_satire_err}
\begin{tabular}{p{6cm}ccc}
\toprule
\textbf{Text (Truncated)} & \textbf{True Label} & \textbf{Predicted Label} & \textbf{Confidence} \\ \midrule
BREAKING: Bridge Collapse Claims Hillary Clinton, 14 Others... & Real & Fake & 0.978 \\
BREAKING: President Trump Releases Official Boycott List... & Real & Fake & 0.966 \\ \midrule
Hillary Clinton: Women Only Voted For Trump Because Husbands... & Fake & Real & 0.508 \\
Procter \& Gamble Release One Of The Most Racist Commercials... & Fake & Real & 0.573 \\
\bottomrule
\end{tabular}
\end{table}

\begin{table}[ht]
\centering
\scriptsize
\caption{Examples of misclassified instances from Fakes dataset. Confidence is the probability of the predicted class.}
\label{tab:fakes_err}
\begin{tabular}{p{6cm}ccc}
\toprule
\textbf{Text (Truncated)} & \textbf{True Label} & \textbf{Predicted Label} & \textbf{Confidence} \\ \midrule
Daesh executed at least 128 Syrian civilians in Al-Qaryatayn... & Fake & Real & 0.786 \\
US Secretary of Defense Carter said anti-Daesh coalition killed key ministers... & Fake & Real & 0.774 \\ \midrule
Several Syrians Killed Injured in Homs Suicide Blasts. & Fake & Real & 0.501 \\
86 dead in suspected Syria chemical attack. & Real & Fake & 0.516 \\
\bottomrule
\end{tabular}
\end{table}

\begin{table}[ht]
\centering
\scriptsize
\caption{Examples of misclassified instances from the Celebrity dataset. Confidence is the probability of the predicted class.}
\label{tab:celebrity_err}
\begin{tabular}{p{6cm}ccc}
\toprule
\textbf{Text (Truncated)} & \textbf{True Label} & \textbf{Predicted Label} & \textbf{Confidence} \\ \midrule
EXCLUSIVE PICS: First photos of Ariana Grande since Manchester attack... & Fake & Real & 0.994 \\
Jennifer Lopez gets cozy with on-screen boyfriend Milo Ventimiglia... & Fake & Real & 0.985 \\ \midrule
White Witch warns Angelina against using spells to woo back Brad... & Fake & Real & 0.521 \\
Is Evian Water The Secret Behind Cameron Diaz's Great Skin? & Fake & Real & 0.544 \\
\bottomrule
\end{tabular}
\end{table}

\subsection{Social media cues and numerical bias}
In datasets involving short claims, the model relies on superficial shortcuts. In Infodemic (Table \ref{tab:infodemic_err}), official institutional tweets are labeled as Fake when they utilize viral social features like hashtags and emojis.
In LIAR-PLUS (Table \ref{tab:liar_err}), the model seems to exhibit a numerical skepticism shortcut, where factual statements containing complex statistics (e.g., "\$20 billion") are flagged as Fake. The model appears to have learned that specific numbers are a hallmark of fabricated evidence used to gain credibility, leading it to penalize data-heavy facts.

\begin{table}[ht]
\centering
\scriptsize
\caption{Examples of misclassified instances from the Infodemic Dataset. Confidence is the probability of the predicted class.}
\label{tab:infodemic_err}
\begin{tabular}{p{6cm}ccc}
\toprule
\textbf{Text (Truncated)} & \textbf{True Label} & \textbf{Predicted Label} & \textbf{Confidence} \\ \midrule
Three stories to read this morning: New lockdown...  & Real  & Fake & 0.999 \\
WHO @UNDP @UNAIDS launch COVID-19 Law Lab...  & Real  & Fake & 0.999 \\ \midrule
I thought Corona affects only humans.  & Fake  & Real & 0.558 \\
I've decided to opt out of the KCL symptoms tracker... \#frustrated & Fake & Real & 0.617 \\
\bottomrule
\end{tabular}
\end{table}

\begin{table}[ht]
\centering
\scriptsize
\caption{Examples of misclassified instances from the LIAR-PLUS Dataset. Confidence is the probability of the predicted class.}
\label{tab:liar_err}
\begin{tabular}{p{6cm}ccc}
\toprule
\textbf{Text (Truncated)} & \textbf{True Label} & \textbf{Predicted} & \textbf{Confidence} \\ \midrule
The costs of cancer and affiliated issues are over \$20 billion a year... & Real & Fake & 0.995 \\
Affordable Care Act is a major reason why weve seen 50,000 fewer deaths... & Real & Fake & 0.994 \\ \midrule
The platform of the Republican Party says deport everybody... & Fake & Real & 0.512 \\
Republican leaders are pushing to make privatizing Social Security... & Fake & Real & 0.513 \\
\bottomrule
\end{tabular}
\end{table}

\subsection{Sentiment artifacts}
In the NDF dataset (Table \ref{tab:ndf_err}), the model seems to conflate stance and emotional tone with factuality. Personal anecdotes expressing shock regarding the fire are consistently labeled as Fake.

\begin{table}[ht]
\centering
\scriptsize
\caption{Examples of misclassified instances from the NDF Dataset. Confidence is the probability of the predicted class.}
\label{tab:ndf_err}
\begin{tabular}{p{6cm}ccc}
\toprule
\textbf{Text (Truncated)} & \textbf{True Label} & \textbf{Predicted Label} & \textbf{Confidence} \\ \midrule
Notre Dame Blaze - a spiritual perspective... & Fake & Real & 0.998 \\
I've spent weekends walking around Paris then stopping in Notre Dame... & Real & Fake & 0.996 \\
So horrible to watch the massive fire at Notre Dame Cathedral... & Real & Fake & 0.994 \\
\bottomrule
\end{tabular}
\end{table}

\FloatBarrier

\section{Training and Inference times}\label{sec:appendix_training_time}

To evaluate the computational efficiency of the considered approaches, we measured the time required for model training and inference. All experiments were executed on the same hardware configuration, using a single NVIDIA A100 GPU (80GB), with CUDA 12.3. Table \ref{tab:time_cost} reports the training times (in seconds) observed across the different experimental setups: training on the specific training splits (used in Exp. 1), training on the entire datasets (used in Exp. 2 for the source domain), training on the Global Training Set (used in Exp. 4), and inference computed as average inference time per sample. As expected, traditional ML models (LR, SVM, NB) are faster than DL and Transformer-based models, with DeBERTa requiring the most computational resources.

For the MERMAID architecture, which is explicitly designed for cross-domain scenarios and utilizes a Mixture-of-Experts mechanism, the training process differs as it involves aggregating multiple datasets (as performed in the Leave-One-Dataset-Out experiments). On average, a complete training and inference run for MERMAID on the aggregated training set (comprising 9 datasets) required approximately 1056 seconds.

Regarding the decoder-only models (Llama3-8B, Qwen3-32B, and Zephyr-7B-beta), since they are employed in zero-shot and few-shot settings without fine-tuning, their computational cost is measured in terms of inference latency rather than training time. In our experiments, Llama3-8b and Zephyr-7B-beta required approximately 1 second per call to generate a classification label, while Qwen3-32B required approximately 1.5 seconds per call.

\begin{table}[H]\centering
\setcounter{table}{0}
\tiny
\caption{Training and inference times (in seconds) for each approach.} 
\label{tab:time_cost}
\begin{tabularx}{\textwidth}{@{}l*{10}{>{\centering\arraybackslash}X}@{}}
\toprule
\textbf{Dataset} & \rotatebox{65}{\textbf{LR}} & \rotatebox{65}{\textbf{SVM}} & \rotatebox{65}{\textbf{NB}} & \rotatebox{65}{\textbf{CNN}} & \rotatebox{65}{\textbf{BiLSTM}} & \rotatebox{65}{\textbf{BERT}} & \rotatebox{65}{\textbf{DeBERTa}} & \rotatebox{65}{\textbf{CNN-BERT}}\\
\midrule
Celebrity (training) & 7.63 & 5.91 & 5.53 & 192.02 & 196.97 & 189.07 & 237.05 & 50.01 \\
Celebrity (entire) & 7.11 & 8.51 & 5.46 & 194.41 & 197.93 & 210.51 & 290.84 & 42.62 \\
Celebrity (inference)& 2.70e-6 & 1.20e-6 & 2.00e-6 & 2.98e-3 & 7.00e-3 & 3.13e-3 & 5.16e-3 & 6.47e-3 \\
\midrule
Cidii (training) & 6.04 & 5.50 & 6.30 & 212.69 & 241.61 & 247.42 & 406.38 & 98.67 \\
Cidii (entire) & 5.81 & 5.59 & 6.44 & 220.64 & 227.22 & 372.50 & 532.82 & 120.75 \\
Cidii (inference)& 8.00e-7 & 7.00e-7 & 9.00e-7 & 1.07e-3 & 7.17e-3 & 2.60e-3 & 4.18e-3 & 5.03e-3 \\
\midrule
FakeVsSatire (training) & 5.85 & 6.19 & 4.73 & 197.89 & 320.77 & 112.99 & 226.94 & 165.54 \\
FakeVsSatire (entire) & 6.01 & 6.43 & 5.38 & 199.13 & 362.28 & 159.71 & 287.80 & 223.96 \\
FakeVsSatire (inference)& 2.10e-6 & 1.10e-6 & 2.10e-6 & 1.55e-3 & 1.32e-2 & 9.37e-3 & 1.03e-2 & 5.34e-3 \\
\midrule
Fakes (training) & 7.53 & 6.10 & 5.48 & 192.00 & 298.61 & 167.53 & 465.00 & 139.27 \\
Fakes (entire) & 7.54 & 8.72 & 6.23 & 195.44 & 344.57 & 229.62 & 598.94 & 164.26 \\
Fakes (inference)& 7.00e-7 & 1.10e-6 & 1.30e-6 & 1.04e-3 & 6.24e-3 & 3.42e-3 & 4.47e-3 & 4.79e-3 \\
\midrule
Horne (training) & 5.86 & 6.12 & 5.18 & 189.63 & 304.99 & 125.53 & 158.07 & 54.14 \\
Horne (entire) & 5.57 & 7.83 & 5.39 & 189.23 & 336.09 & 167.11 & 195.21 & 60.40 \\
Horne (inference)& 2.60e-6 & 1.70e-6 & 3.00e-6 & 2.27e-3 & 7.19e-3 & 3.16e-3 & 4.94e-3 & 8.37e-3 \\
\midrule
Infodemic (training) & 7.22 & 6.99 & 6.01 & 211.85 & 322.86 & 2976.10 & 3959.74 & 153.17 \\
Infodemic (entire) & 7.58 & 9.68 & 6.49 & 228.51 & 401.97 & 4975.64 & 5548.36 & 233.33 \\
Infodemic (inference)& 7.40e-6 & 1.00e-7 & 2.00e-7 & 3.80e-4 & 1.98e-3 & 1.40e-3 & 2.30e-3 & 3.60e-3 \\
\midrule
Isot (training) & 30.67 & 31.20 & 25.82 & 280.17 & 1476.91 & 15001.79 & 19485.40 & 770.47 \\ 
Isot (entire) & 40.58 & 47.25 & 29.45 & 313.20 & 2042.75 & 20772.43 & 24366.53 & 977.61 \\ 
Isot (inference)& 2.00e-7 & 2.00e-7 & 7.00e-7 & 3.25e-4 & 3.94e-3 & 2.83e-3 & 3.99e-3 & 3.35e-3 \\
\midrule
LIAR-PLUS (training) & 6.59 & 7.07 & 5.95 & 465.26 & 710.13 & 4485.00 & 5994.13 & 190.46 \\
LIAR-PLUS (entire) & 8.37 & 10.39 & 8.02 & 536.26 & 835.91 & 5342.10 & 6687.75 & 241.62 \\
LIAR-PLUS (inference)& 1.00e-7 & 1.00e-7 & 1.00e-7 & 3.96e-4 & 8.28e-3 & 1.17e-3 & 2.02e-3 & 3.33e-3 \\
\midrule
NDF (training) & 5.41 & 5.03 & 4.50 & 183.68 & 214.27 & 107.22 & 438.73 & 79.97 \\
NDF (entire) & 6.55 & 6.57 & 5.41 & 214.70 & 219.96 & 130.25 & 391.41 & 97.51 \\
NDF (inference)& 1.00e-6 & 9.00e-7 & 1.00e-6 & 1.39e-3 & 8.31e-3 & 2.63e-3 & 6.27e-3 & 6.21e-3 \\
\midrule
Politifact (training) & 6.74 & 6.15 & 5.08 & 208.39 & 353.23 & 387.96 & 291.04 & 47.67 \\
Politifact (entire) & 7.34 & 6.55 & 6.17 & 212.52 & 390.18 & 150.15 & 365.89 & 47.75 \\
Politifact (inference) & 1.90e-6 & 1.10e-6 & 2.30e-6 & 1.20e-3 & 5.67e-3 & 2.74e-3 & 4.21e-3 & 4.99e-3 \\
\midrule
\textit{Global training} & 6.68 & 9.02 & 8.34 & 193.21 & 290.87 & 859.98 & 1093.43 & 112.82 \\
Inference & 5.00e-7 & 4.00e-7 & 1.10e-6 & 6.15e-4 & 4.62e-3 & 2.69e-3 & 4.05e-3 & 3.79e-3 \\
\bottomrule
\end{tabularx}
\end{table}

Here we report model size and empirical training/test times to provide a transparent overview of computational requirements. However, the main analysis in this work is performance-driven rather than compute-normalized. Our goal is to compare robustness and generalization behavior across modeling paradigms, not to derive an efficiency ranking under a fixed computational budget.

It is worth noting that performance does not scale linearly with parameter count or training cost. Moreover, the architectures considered span different technological generations and design philosophies, ranging from traditional discriminative models (e.g., Logistic Regression) to encoder-based transformers such as DeBERTa and decoder-only LLMs. These models differ not only in parameter count, but also in pretraining regimes, optimization strategies, and typical deployment settings. As a result, a strict compute-normalized comparison would require additional assumptions about hardware, batching strategies, inference constraints, and amortization of pretraining cost.

For reference, DeBERTa achieves the highest absolute F1-scores (median 0.865 in E1; Table~\ref{tab:robustness_summary}), but with a substantially larger parameter count ($1.39 \times 10^8$) and a marked drop under domain shift (median 0.535 in E2). In contrast, more specialized architectures such as MERMAID provide competitive generalization in Leave-One-Dataset-Out settings (median 0.645; Table~\ref{tab:all_f1_models_by_dataset_lodo}) with a modular MoE structure and lower overall computational overhead. Decoder-only LLMs represent yet another configuration: while their total parameter count is higher and inference in our configuration is relatively slow, there exist highly optimized in modern implementations, and all LLMs tested here can in principle be run on commodity hardware (e.g., a consumer-grade laptop).

Overall, while a compute-aware normalization could yield a different ranking under strict resource constraints, such a perspective falls outside the scope of the present study. Our comparative analysis is intended to assess robustness and cross-domain generalization across model families, assuming comparable practical feasibility rather than a fixed compute budget. A systematic cost-normalized evaluation is left for future work.

\FloatBarrier

\newcolumntype{Y}{>{\centering\arraybackslash}X}


\renewcommand{\appendixname}{Supplementary Materials}

\begin{center}
    \LARGE \textbf{Supplementary Material}
    \vspace{1em} 
\end{center}

\setcounter{table}{0}
\renewcommand{\thetable}{S\arabic{table}}
\setcounter{figure}{0}
\renewcommand{\thefigure}{S\arabic{figure}}


This supplementary material provides a detailed breakdown of the experimental results shown in the paper. Regarding the classification performance, we present Precision (P), Recall (R), and F1-score (F1) for each class individually (Class 0 and Class 1 representing `Real' and `Fake' news, respectively). This granular analysis allows for a deeper understanding of each model's behavior, helping to reveal any potential biases towards one class over the other, which can be obscured by a single, aggregated score. The only exception concerns the Cross-Dataset Experiments, where, for the sake of brevity, we only report the F1-scores obtained by training each model on a single dataset and testing on all the others. Additionally, for each experiment, we report the results of the statistical tests performed to evaluate the statistical significance of the performance differences among the models. To this aim, we computed the Friedman test: if the p-value falls below the significance threshold (alpha=0.05), the null hypothesis is rejected, indicating that at least one model performs significantly differently from the others. In such cases (all in our experiments), the post-hoc Nemenyi test is performed to identify the specific pairs of models exhibiting statistically significant differences. In addition to the Friedman test, we computed Kendall’s W (also known as Kendall’s coefficient of concordance) to quantify the effect size. It measures the level of agreement among the different datasets regarding the ranking of the models.


\section{Dataset-Specific Experiments}\label{sec:appendix_exp1}

\setlength{\tabcolsep}{1.5pt}
This section presents the detailed experimental results, along with the application of statistical tests on the F1-scores for the Dataset-Specific experiments. 

\subsection{Detailed experimental results}

Tables \ref{tab:best_for_classes_ML}--\ref{tab:best_for_classes_Few_EXP1} present precision, recall and F1-score on the two classes obtained by using the traditional ML approaches, the DL approaches trained from scratch, and the Transformer-based approaches. 

\begin{table}[H]
\centering
\scriptsize
\caption{Precision, recall and F1-score obtained by the ML models on each dataset for each class.}
\label{tab:best_for_classes_ML}
\begin{tabularx}{\textwidth}{@{}p{1.6cm} *{18}{Y} @{}}
\toprule
\textbf{Dataset} & \multicolumn{6}{c}{\textbf{LR}} & \multicolumn{6}{c}{\textbf{SVM}} & \multicolumn{6}{c}{\textbf{NB}} \\
\cmidrule(lr){2-7} \cmidrule(lr){8-13} \cmidrule(lr){14-19}
& \multicolumn{3}{c}{\textbf{Class 0}} & \multicolumn{3}{c}{\textbf{Class 1}} & \multicolumn{3}{c}{\textbf{Class 0}} & \multicolumn{3}{c}{\textbf{Class 1}} & \multicolumn{3}{c}{\textbf{Class 0}} & \multicolumn{3}{c}{\textbf{Class 1}} \\
\cmidrule(lr){2-4} \cmidrule(lr){5-7} \cmidrule(lr){8-10} \cmidrule(lr){11-13} \cmidrule(lr){14-16} \cmidrule(lr){17-19}
& P & R & F1 & P & R & F1 & P & R & F1 & P & R & F1 & P & R & F1 & P & R & F1\\
\midrule
Celebrity & .59 & .62 & .60 & .60 & .58 & .59 & .66 & .70 & .67 & .68 & .64 & .65 & .48 & .76 & .58 & .42 & .18 & .25 \\
Cidii & 1.0 & .93 & .96 & .95 & 1.0 & .97 & .82 & .85 & .83 & .89 & .87 & .88 & .88 & .78 & .83 & .85 & .92 & .89 \\
FakeVsSatire & .87 & .34 & .49 & .67 & .96 & .79 & .68 & .48 & .57 & .69 & .84 & .76 & .81 & .31 & .45 & .65 & .94 & .77 \\
Fakes & .45 & .47 & .46 & .39 & .38 & .38 & .47 & .41 & .44 & .42 & .48 & .45 & .46 & .50 & .48 & .38 & .34 & .36 \\
Horne & .71 & 1.0 & .83 & 1.0 & .36 & .52 & .70 & .92 & .80 & .75 & .36 & .48 & .66 & 1.0 & .79 & 1.0 & .16 & .27 \\
Infodemic & .94 & .92 & .93 & .92 & .94 & .93 & .93 & .94 & .93 & .93 & .92 & .92 & .89 & .88 & .88 & .87 & .88 & .87 \\
Isot & .98 & .98 & .98 & .98 & .98 & .98 & .99 & .99 & .99 & .99 & .99 & .99 & .94 & .97 & .96 & .97 & .94 & .96 \\
LIAR-PLUS & .60 & .60 & .60 & .49 & .49 & .49 & .61 & .61 & .61 & .50 & .49 & .50 & .60 & .62 & .61 & .49 & .47 & .48 \\
NDF & .86 & .92 & .89 & .86 & .76 & .81 & .81 & .92 & .86 & .85 & .67 & .75 & .89 & .75 & .81 & .68 & .86 & .76 \\
Politifact & .72 & 1.0 & .83 & 1.0 & .28 & .44 & .78 & .91 & .84 & .75 & .52 & .61 & .69 & .94 & .80 & .71 & .23 & .35 \\
\bottomrule
\end{tabularx}
\end{table}

\begin{table}[H]
\centering
\scriptsize
\caption{Precision, recall and F1-score obtained by the CNN, BiLSTM and CNN-BERT models on each dataset for each class.}
\label{tab:best_for_classes_CNN_BILSTM_COMPLETED}
\begin{tabularx}{\textwidth}{@{}p{1.6cm} *{18}{Y} @{}}
\toprule
\textbf{Dataset} & \multicolumn{6}{c}{\textbf{CNN}} & \multicolumn{6}{c}{\textbf{BiLSTM}} & \multicolumn{6}{c}{\textbf{CNN-BERT}}\\
\cmidrule(lr){2-7} \cmidrule(lr){8-13} \cmidrule(lr){14-19}
& \multicolumn{3}{c}{\textbf{Class 0}} & \multicolumn{3}{c}{\textbf{Class 1}} & \multicolumn{3}{c}{\textbf{Class 0}} & \multicolumn{3}{c}{\textbf{Class 1}} & \multicolumn{3}{c}{\textbf{Class 0}} & \multicolumn{3}{c}{\textbf{Class 1}} \\
\cmidrule(lr){2-4} \cmidrule(lr){5-7} \cmidrule(lr){8-10} \cmidrule(lr){11-13} \cmidrule(lr){14-16} \cmidrule(lr){17-19}
& P & R & F1 & P & R & F1 & P & R & F1 & P & R & F1 & P & R & F1 & P & R & F1\\
\midrule
Celebrity & .52 & .88 & .65 & .62 & .20 & .30 & .50 & .16 & .24 & .50 & .84 & .62 & .50 & .96 & .65 & .50 & .04 & .07 \\
Cidii & .44 & .06 & .11 & .58 & .94 & .72 & .50 & .05 & .09 & .58 & .96 & .73 & .00 & .00 & .00 & .58 & 1.0 & .73 \\
FakeVsSatire & .00 & .00 & .00 & .57 & .96 & .71 & .39 & .31 & .35 & .56 & .64 & .60 & .00 & .00 & .00 & .58 & 1.0 & .73 \\
Fakes & .52 & 1.0 & .69 & .00 & .00 & .00 & .55 & .77 & .65 & .55 & .31 & .40 & .52 & 1.0 & .69 & .00 & .00 & .00 \\
Horne & .00 & .00 & .00 & .57 & .96 & .71 & .62 & 1.0 & .76 & .00 & .00 & .00 & .81 & .65 & .72 & .57 & .76 & .65 \\
Infodemic & .57 & .55 & .56 & .52 & .54 & .53 & .66 & .85 & .74 & .76 & .53 & .62 & .66 & .76 & .71 & .69 & .57 & .62 \\
Isot & .98 & .93 & .96 & .94 & .98 & .96 & .95 & .98 & .97 & .98 & .95 & .97 & .99 & .98 & .99 & .98 & .99 & .99 \\
LIAR-PLUS & .57 & .90 & .70 & .51 & .12 & .20 & .59 & .76 & .66 & .51 & .33 & .40 & .58 & .91 & .71 & .57 & .14 & .23 \\
NDF & .61 & .73 & .67 & .40 & .27 & .32 & .75 & .58 & .66 & .51 & .69 & .59 & .60 & .92 & .73 & .28 & .04 & .08 \\
Politifact & .64 & .98 & .78 & .00 & .00 & .00 & .65 & 1.0 & .78 & .00 & .00 & .00 & .66 & .88 & .76 & .47 & .19 & .27 \\
\bottomrule
\end{tabularx}
\end{table}

\begin{table}[H]
\centering
\scriptsize
\caption{Precision, recall and F1-score obtained by the BERT and DeBERTa models on each dataset for each class.}
\label{tab:best_for_classes_BERT_GPT_MODIFIED}
\begin{tabularx}{\textwidth}{@{}p{1.6cm} *{12}{Y} @{}}
\toprule
\textbf{Dataset} & \multicolumn{6}{c}{\textbf{BERT}} & \multicolumn{6}{c}{\textbf{DeBERTa}} \\
\cmidrule(lr){2-7} \cmidrule(lr){8-13}
& \multicolumn{3}{c}{\textbf{Class 0}} & \multicolumn{3}{c}{\textbf{Class 1}} & \multicolumn{3}{c}{\textbf{Class 0}} & \multicolumn{3}{c}{\textbf{Class 1}} \\
\cmidrule(lr){2-4} \cmidrule(lr){5-7} \cmidrule(lr){8-10} \cmidrule(lr){11-13}
& P & R & F1 & P & R & F1 & P & R & F1 & P & R & F1\\
\midrule
Celebrity & .75 & .78 & .76 & .77 & .74 & .75 & .80 & .86 & .83 & .85 & .78 & .81 \\
Cidii & 1.0 & .95 & .97 & .96 & 1.0 & .98 & .97 & .98 & .97 & .97 & .95 & .96 \\
FakeVsSatire & .79 & .46 & .58 & .70 & .91 & .79 & .69 & .83 & .76 & .86 & .74 & .79 \\
Fakes & .37 & .48 & .42 & .52 & .40 & .45 & .49 & .27 & .35 & .46 & .68 & .55 \\
Horne & .71 & .95 & .81 & .81 & .36 & .50 & .97 & .95 & .96 & .92 & .96 & .94 \\
Infodemic & .97 & .98 & .97 & .97 & .97 & .97 & .97 & .99 & .98 & .98 & .96 & .97 \\
Isot & 1.0 & 1.0 & 1.0 & 1.0 & 1.0 & 1.0 & 1.0 & 1.0 & 1.0 & 1.0 & 1.0 & 1.0 \\
LIAR-PLUS & .65 & .67 & .66 & .57 & .54 & .55 & .67 & .69 & .68 & .59 & .56 & .57 \\
NDF & .82 & .92 & .87 & .85 & .69 & .76 & .95 & .84 & .89 & .78 & .93 & .85 \\
Politifact & .87 & .88 & .87 & .78 & .76 & .77 & .94 & .84 & .89 & .75 & .89 & .81 \\
\bottomrule
\end{tabularx}
\end{table}

\begin{table}[H]
\centering
\scriptsize
\caption{Precision, recall and F1-score obtained by the Llama3-8B, Qwen3-32B and Zephyr-7B-beta models on each dataset for each class in a zero-shot (ZS) scenario.}
\label{tab:best_for_classes_Zero_EXP1}
\begin{tabularx}{\textwidth}{@{}p{1.6cm} *{18}{Y} @{}}
\toprule
\textbf{Dataset} & \multicolumn{6}{c}{\textbf{Llama3-8B (ZS)}} & \multicolumn{6}{c}{\textbf{Qwen3-32B (ZS)}} & \multicolumn{6}{c}{\textbf{Zephyr-7B-beta (ZS)}} \\
\cmidrule(lr){2-7} \cmidrule(lr){8-13} \cmidrule(lr){14-19}
& \multicolumn{3}{c}{\textbf{Class 0}} & \multicolumn{3}{c}{\textbf{Class 1}} & \multicolumn{3}{c}{\textbf{Class 0}} & \multicolumn{3}{c}{\textbf{Class 1}} & \multicolumn{3}{c}{\textbf{Class 0}} & \multicolumn{3}{c}{\textbf{Class 1}} \\
\cmidrule(lr){2-4} \cmidrule(lr){5-7} \cmidrule(lr){8-10} \cmidrule(lr){11-13} \cmidrule(lr){14-16} \cmidrule(lr){17-19}
& P & R & F1 & P & R & F1 & P & R & F1 & P & R & F1 & P & R & F1 & P & R & F1\\
\midrule
Celebrity & .71 & .88 & .79 & .84 & .64 & .73 & .78 & .86 & .82 & .84 & .76 & .80 & .63 & .82 & .71 & .76 & .50 & .60 \\
Cidii & .87 & .95 & .91 & .92 & .80 & .86 & .96 & .93 & .95 & .90 & .95 & .93 & .80 & 1.00 & .89 & 1.00 & .65 & .79 \\
FakeVsSatire & .00 & .00 & .00 & .58 & .98 & .73 & .00 & .00 & .00 & .58 & 1.00 & .74 & .07 & .02 & .04 & .52 & .75 & .62 \\
Fakes & .53 & 1.00 & .69 & .00 & .00 & .00 & .50 & .86 & .63 & .25 & .05 & .09 & .52 & .98 & .68 & .00 & .00 & .00 \\
Horne & .88 & .54 & .67 & .55 & .88 & .68 & .91 & .51 & .66 & .53 & .92 & .68 & .70 & .51 & .59 & .46 & .64 & .53 \\
Infodemic & .71 & .97 & .82 & .95 & .56 & .71 & .81 & .91 & .86 & .89 & .76 & .82 & .69 & 1.00 & .81 & .99 & .50 & .67 \\
Isot & .65 & 1.00 & .78 & 1.00 & .48 & .65 & .66 & 1.00 & .79 & .99 & .52 & .68 & .57 & .97 & .72 & .92 & .31 & .46 \\
LIAR-PLUS & .61 & .85 & .71 & .61 & .31 & .41 & .68 & .52 & .59 & .53 & .69 & .60 & .61 & .92 & .73 & .70 & .24 & .35 \\
NDF & .78 & .94 & .85 & .86 & .58 & .69 & .88 & .90 & .89 & .83 & .81 & .82 & .78 & .99 & .87 & .96 & .56 & .71 \\
Politifact & .76 & .84 & .80 & .86 & .49 & .62 & .78 & .96 & .86 & .86 & .49 & .62 & .73 & .69 & .71 & .61 & .51 & .56 \\
\bottomrule
\end{tabularx}
\end{table}

\begin{table}[H]
\centering
\scriptsize
\caption{Precision, recall and F1-score obtained by the Llama3-8B, Qwen3-32B and Zephyr-7B-beta models on each dataset for each class in a few-shot (FS) scenario.}
\label{tab:best_for_classes_Few_EXP1}
\begin{tabularx}{\textwidth}{@{}p{1.6cm} *{18}{Y} @{}}
\toprule
\textbf{Dataset} & \multicolumn{6}{c}{\textbf{Llama3-8B (FS)}} & \multicolumn{6}{c}{\textbf{Qwen3-32B (FS)}} & \multicolumn{6}{c}{\textbf{Zephyr-7B-beta (FS)}} \\
\cmidrule(lr){2-7} \cmidrule(lr){8-13} \cmidrule(lr){14-19}
& \multicolumn{3}{c}{\textbf{Class 0}} & \multicolumn{3}{c}{\textbf{Class 1}} & \multicolumn{3}{c}{\textbf{Class 0}} & \multicolumn{3}{c}{\textbf{Class 1}} & \multicolumn{3}{c}{\textbf{Class 0}} & \multicolumn{3}{c}{\textbf{Class 1}} \\
\cmidrule(lr){2-4} \cmidrule(lr){5-7} \cmidrule(lr){8-10} \cmidrule(lr){11-13} \cmidrule(lr){14-16} \cmidrule(lr){17-19}
& P & R & F1 & P & R & F1 & P & R & F1 & P & R & F1 & P & R & F1 & P & R & F1\\
\midrule
Celebrity & .84 & .62 & .71 & .70 & .88 & .78 & .76 & .76 & .76 & .76 & .76 & .76 & .71 & .72 & .71 & .73 & .70 & .71 \\
Cidii & .94 & 1.00 & .97 & 1.00 & .92 & .96 & .98 & .95 & .96 & .94 & .97 & .95 & .89 & .96 & .93 & .94 & .83 & .88 \\
FakeVsSatire & .00 & .00 & .00 & .58 & .93 & .72 & .50 & .02 & .05 & .58 & .98 & .73 & .00 & .00 & .00 & .57 & .91 & .70 \\
Fakes & .51 & .88 & .65 & .33 & .07 & .11 & .50 & .66 & .57 & .40 & .25 & .31 & .53 & 1.00 & .69 & .00 & .00 & .00 \\
Horne & .88 & .54 & .67 & .55 & .88 & .68 & .90 & .63 & .74 & .59 & .88 & .71 & .72 & .32 & .44 & .45 & .80 & .58 \\
Infodemic & .78 & .98 & .87 & .97 & .69 & .81 & .81 & .96 & .88 & .95 & .75 & .84 & .78 & .98 & .87 & .97 & .70 & .81 \\
Isot & .81 & 1.00 & .89 & 1.00 & .77 & .87 & .71 & 1.00 & .83 & 1.00 & .63 & .77 & .58 & .96 & .72 & .90 & .34 & .49 \\
LIAR-PLUS & .61 & .84 & .70 & .59 & .29 & .39 & .69 & .60 & .64 & .56 & .66 & .61 & .65 & .79 & .72 & .64 & .46 & .53 \\
NDF & .84 & .93 & .88 & .86 & .72 & .78 & .92 & .96 & .94 & .93 & .86 & .89 & .83 & .96 & .89 & .91 & .67 & .77 \\
Politifact & .84 & .93 & .88 & .86 & .72 & .78 & .81 & .80 & .81 & .63 & .65 & .64 & .74 & .40 & .52 & .45 & .73 & .56 \\
\bottomrule
\end{tabularx}
\end{table}

\subsection{Statistical tests}

\begin{figure}[!htb]
\centering
\includegraphics[width=0.7\linewidth]{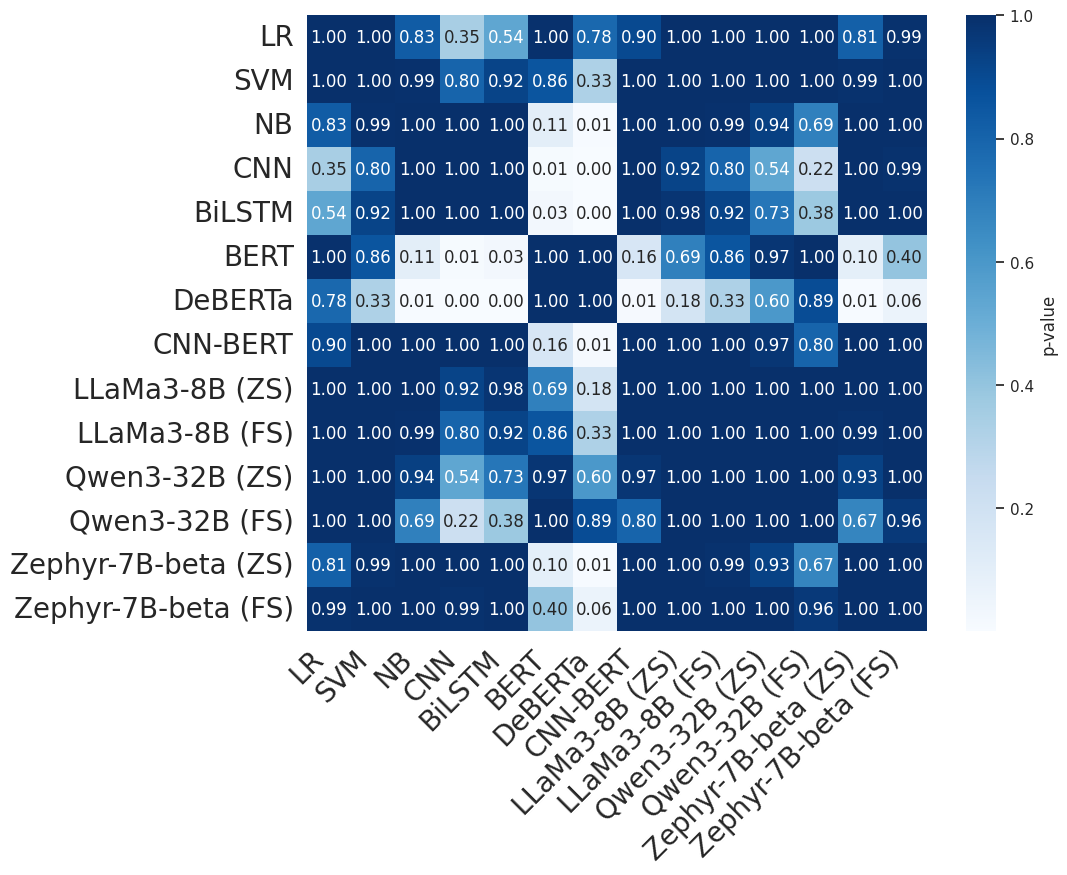} 
\caption{Heatmap of the p-values computed by the post-hoc Nemenyi Test (Pairwise Model Comparisons).}
\label{fig:heat1} 
\end{figure}

We obtained a Friedman test statistic of 46.44, a p-value of 1.2e-05, and a Kendall's W of 0.357. Since the p-value is below the significance threshold alpha=0.05, the post-hoc Nemenyi test was then performed to identify the specific pairs of models exhibiting statistically significant differences. Figure \ref{fig:heat1} reports the pairwise p-values obtained from the Nemenyi post-hoc test, where values below 0.05 indicate  statistically significant performance differences.

\FloatBarrier

\section{Cross-Dataset Experiments}\label{sec:appendix_exp2}
This section presents detailed experimental results and details on the
application of the statistical tests on the F1-score for the Cross-Dataset
experiments.

\subsection{Detailed experimental results}

\begin{figure}[H]
\centering
\begin{subfigure}{0.47\textwidth}
    \centering
    \includegraphics[width=\linewidth]{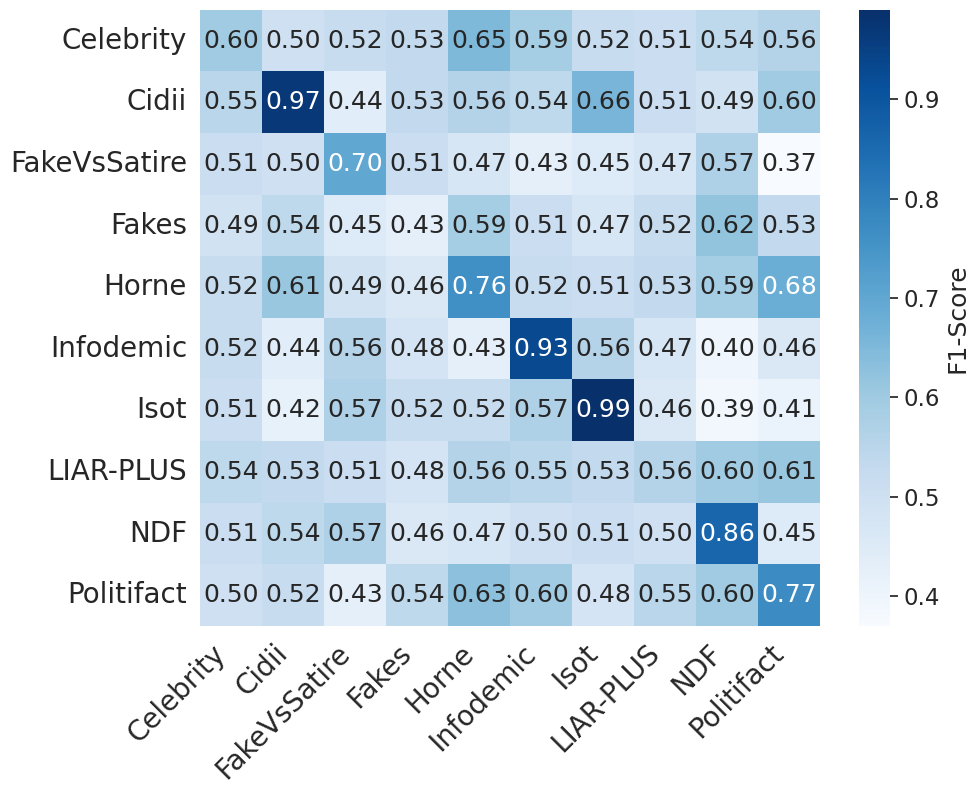}
    \caption{LR: Cross-dataset F1-score}
    \label{fig:head1}
\end{subfigure}
\hfill
\begin{subfigure}{0.47\textwidth}
    \centering
    \includegraphics[width=\linewidth]{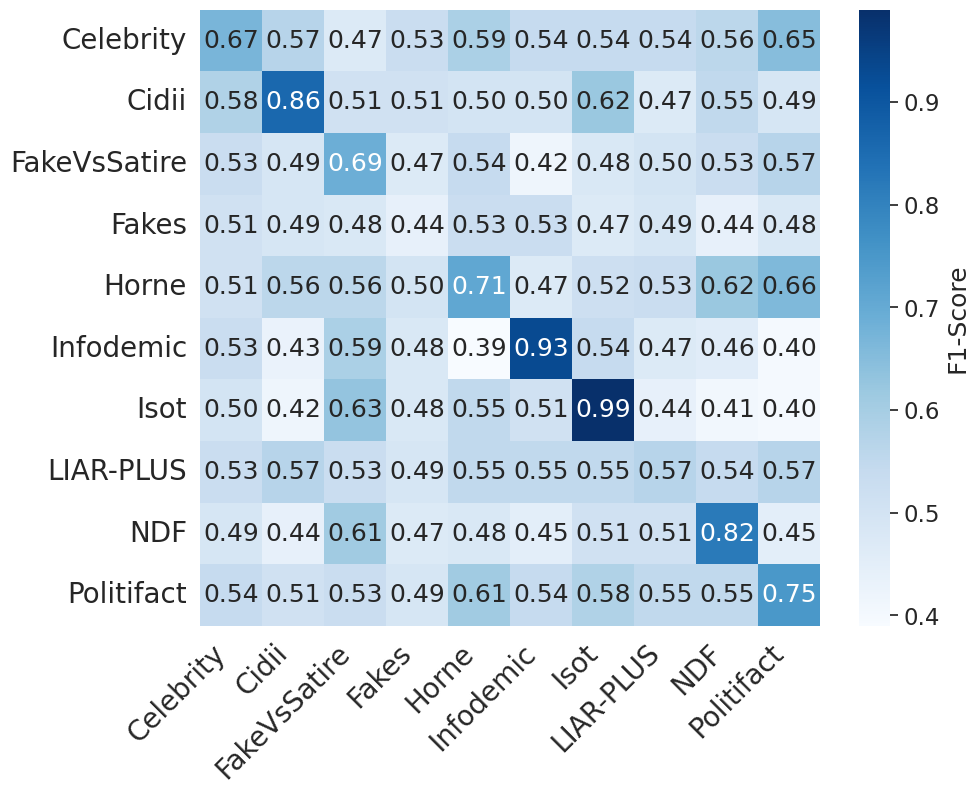}
    \caption{SVM: Cross-dataset F1-score}
    \label{fig:head2}
\end{subfigure}

\medskip

\begin{subfigure}{0.47\textwidth}
    \centering
    \includegraphics[width=\linewidth]{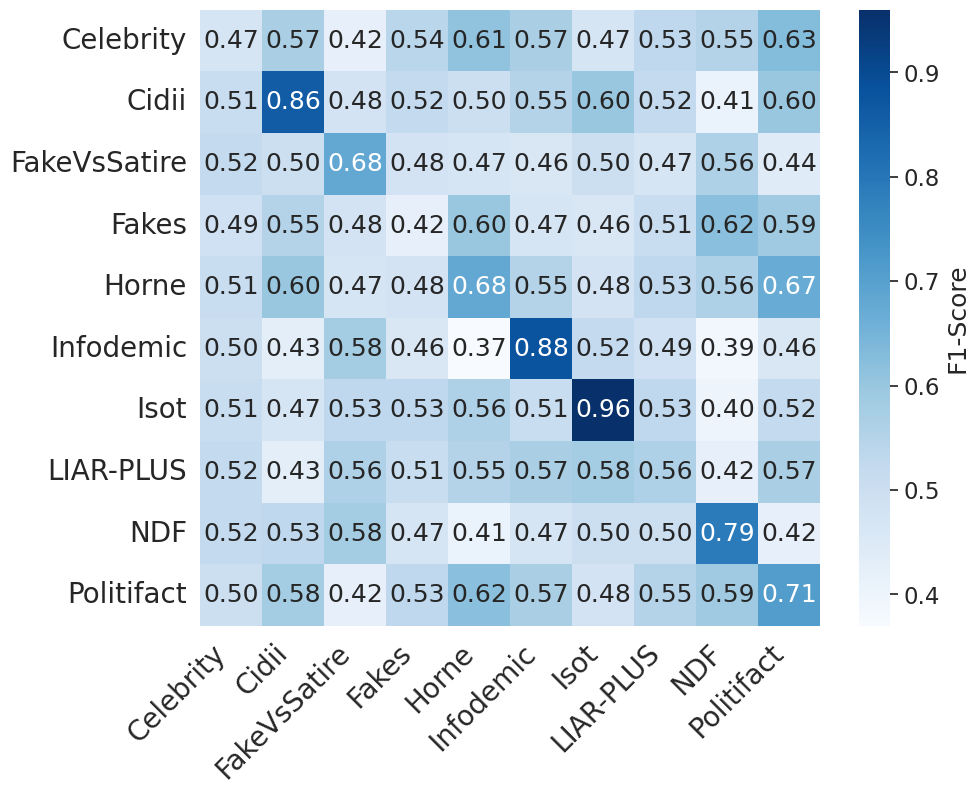}
    \caption{NB: Cross-dataset F1-score}
    \label{fig:head3}
\end{subfigure}
\caption{Cross-dataset F1-scores for traditional ML approaches. As a reference, we have reported in the diagonal of the heatmap the highest F1-scores obtained by the models in the Dataset-Specific experiments.}
\label{fig:main_ML}
\end{figure}

\begin{figure}[H]
\centering
\begin{subfigure}{0.47\textwidth}
    \centering
    \includegraphics[width=\linewidth]{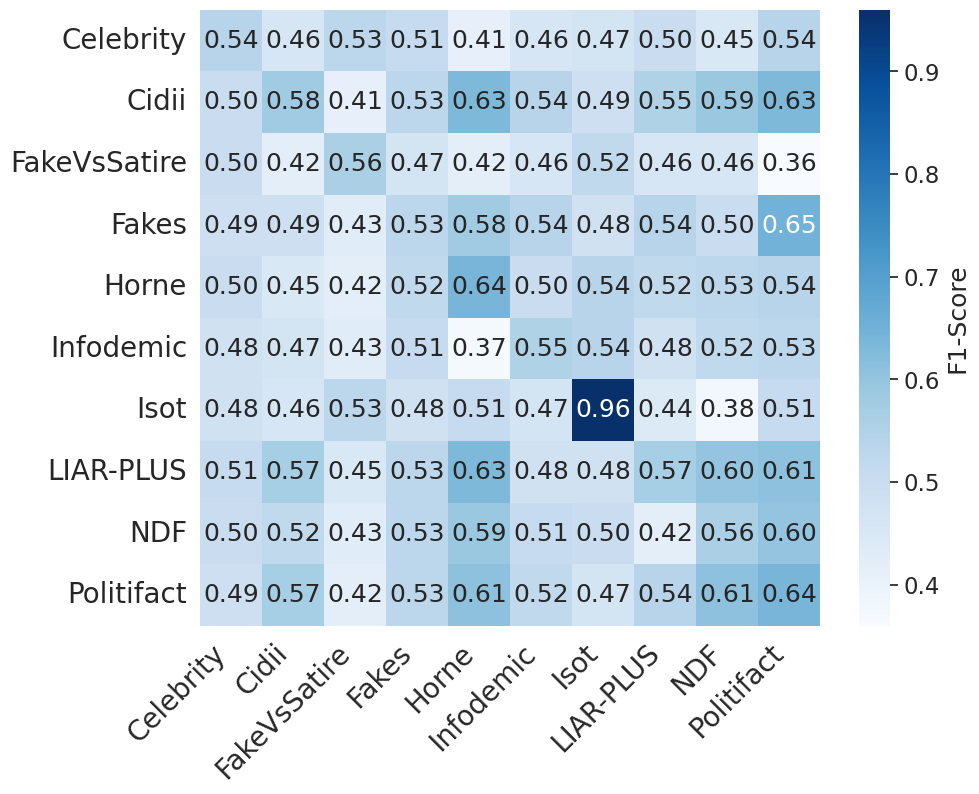}
    \caption{CNN: Cross-dataset F1-score}
    \label{fig:head4}
\end{subfigure}
\hfill
\begin{subfigure}{0.47\textwidth}
    \centering
    \includegraphics[width=\linewidth]{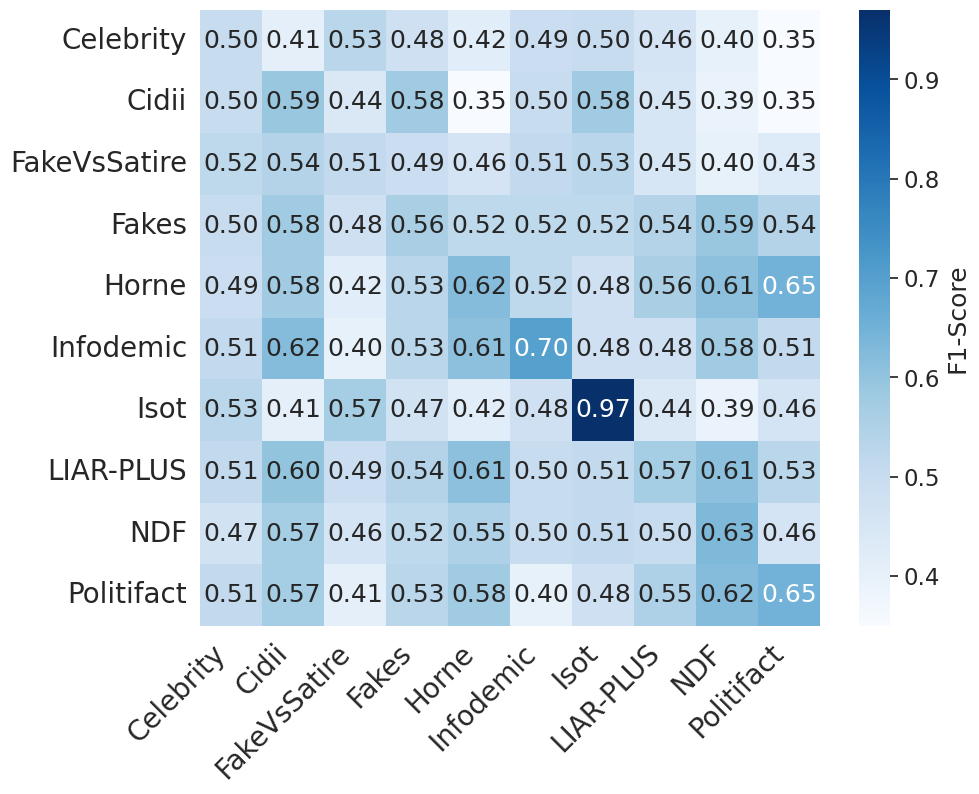}
    \caption{BiLSTM: Cross-dataset F1-score}
    \label{fig:head5}
\end{subfigure}

\medskip

\begin{subfigure}{0.47\textwidth}
    \centering
    \includegraphics[width=\linewidth]{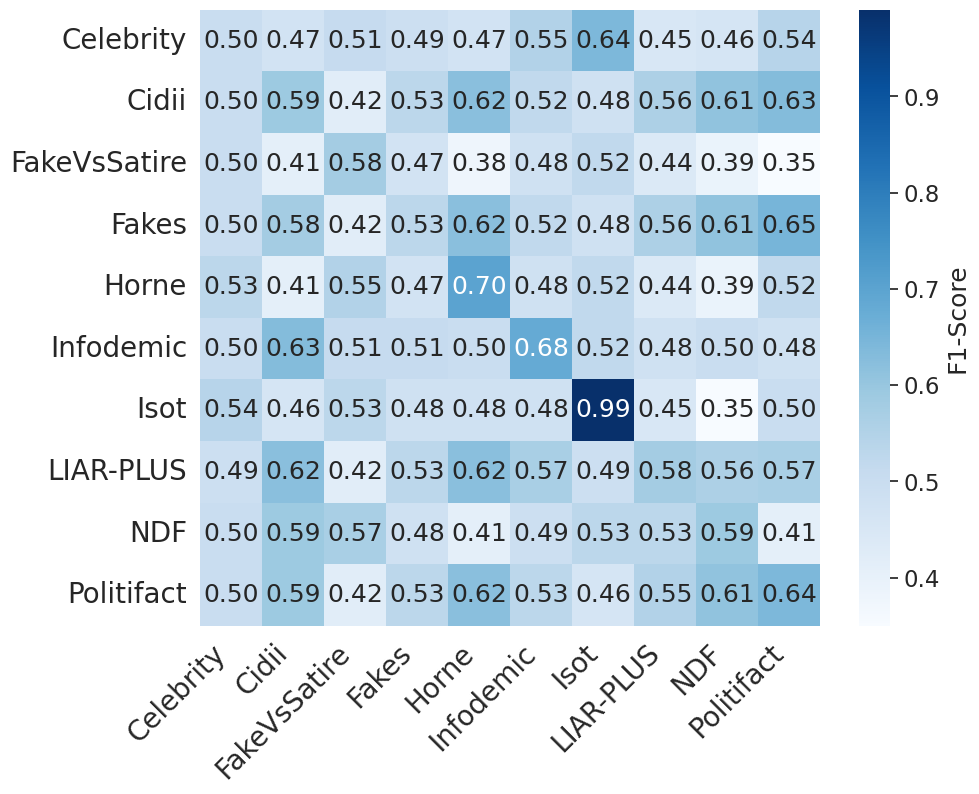}
    \caption{CNN-BERT: Cross-dataset F1-score}
    \label{fig:head6}
\end{subfigure}
\caption{Cross-dataset F1-scores for DL approaches trained from scratch. As a reference, we have reported in the diagonal of the heatmap the highest F1-scores obtained by the models in the Dataset-Specific experiments.}
\label{fig:main_DL}
\end{figure}

\begin{figure}[H]
\centering
\begin{subfigure}{0.46\textwidth}
    \centering
    \includegraphics[width=\linewidth]{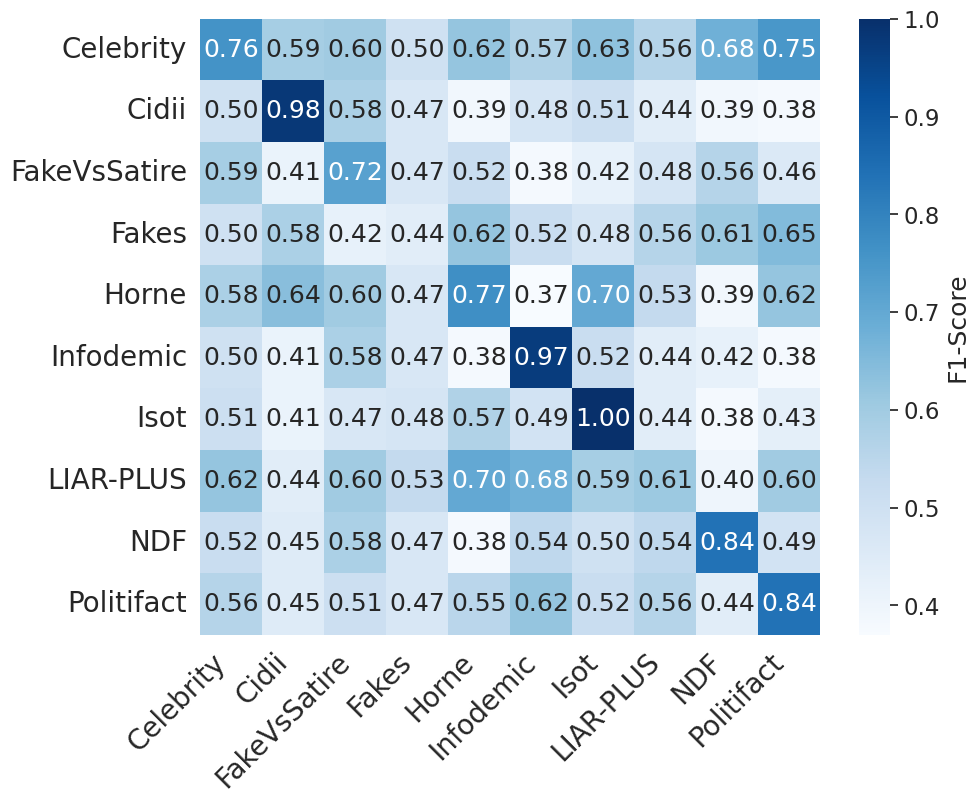}
    \caption{BERT: Cross-dataset F1-score}
    \label{fig:head7}
\end{subfigure}
\hfill
\begin{subfigure}{0.46\textwidth}
    \centering
    \includegraphics[width=\linewidth]{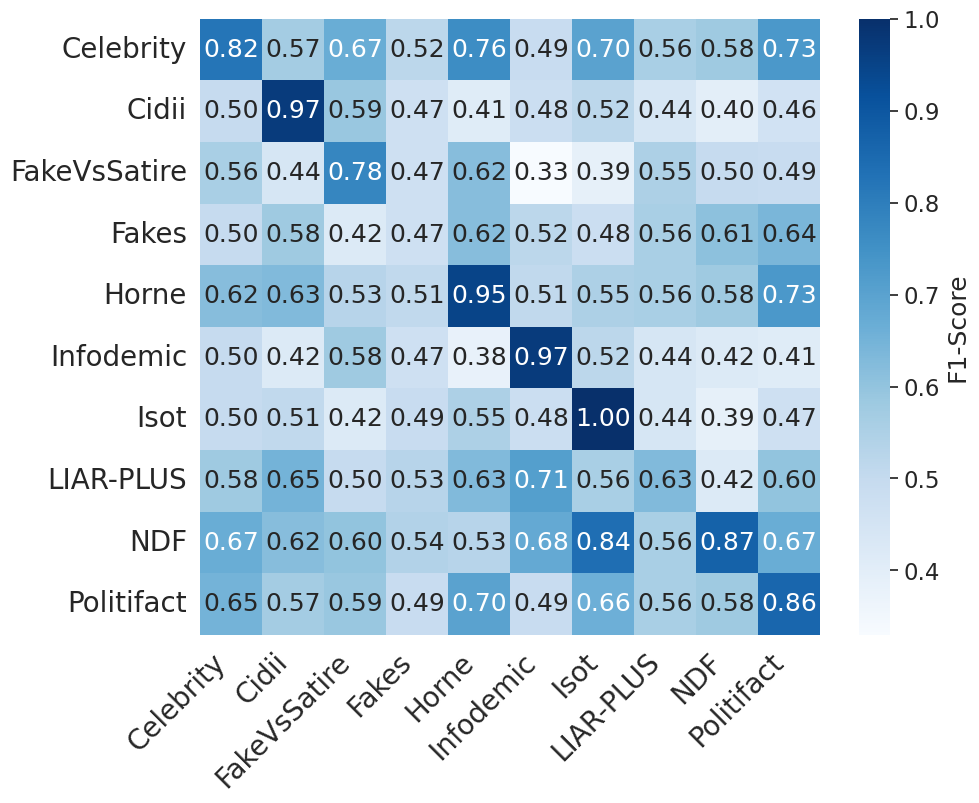}
    \caption{DeBERTa: Cross-dataset F1-score}
    \label{fig:head8}
\end{subfigure}
\caption{Cross-dataset F1-scores for Transformer-based models. As a reference, we have reported in the diagonal of the heatmap the highest F1-scores obtained by the models in the Dataset-Specific Experiments.}
\label{fig:main_TR}
\end{figure}

\begin{figure}[H]
\centering
\begin{subfigure}{0.39\textwidth}
    \centering
    \includegraphics[width=\linewidth]{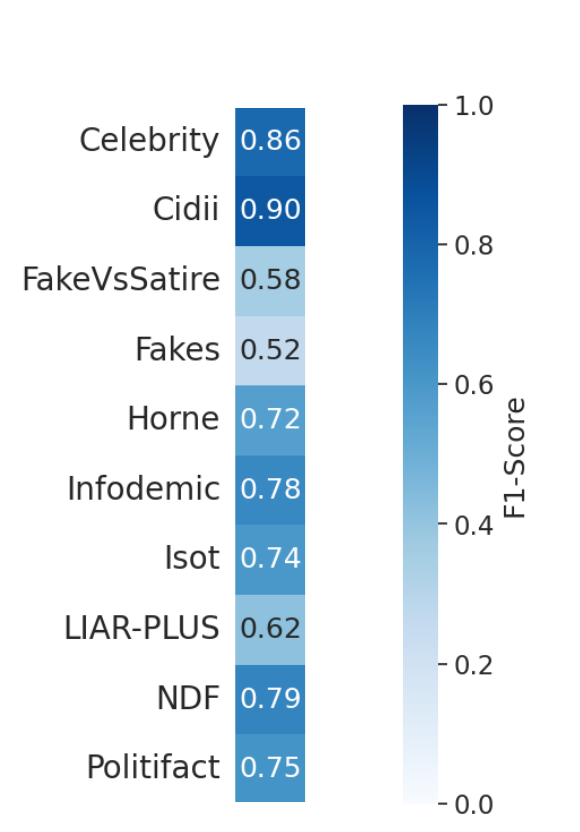}
    \caption{Llama3-8B (ZS)}
    \label{fig:headLLAMA}
\end{subfigure}
\hfill
\begin{subfigure}{0.36\textwidth}
    \centering
    \includegraphics[width=\linewidth]{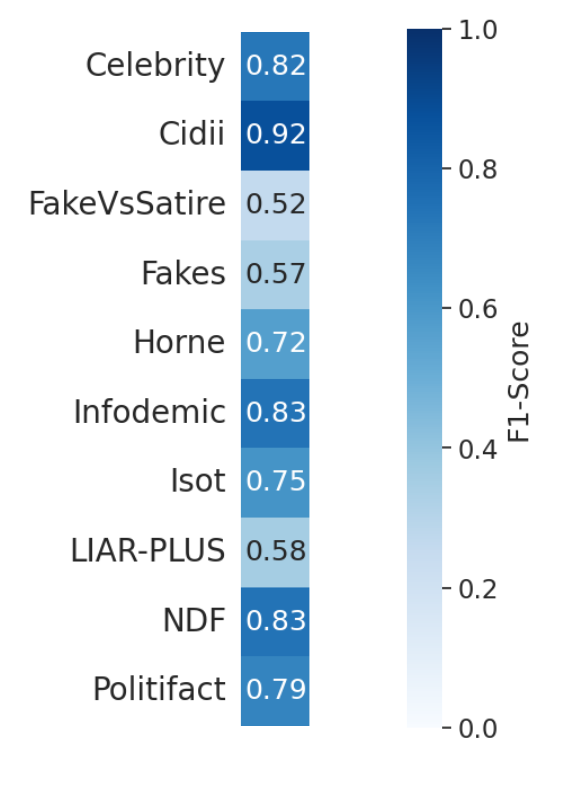}
    \caption{Qwen3-32B (ZS)}
    \label{fig:headQwen}
\end{subfigure}
\hfill
\begin{subfigure}{0.36\textwidth}
    \centering
    \includegraphics[width=\linewidth]{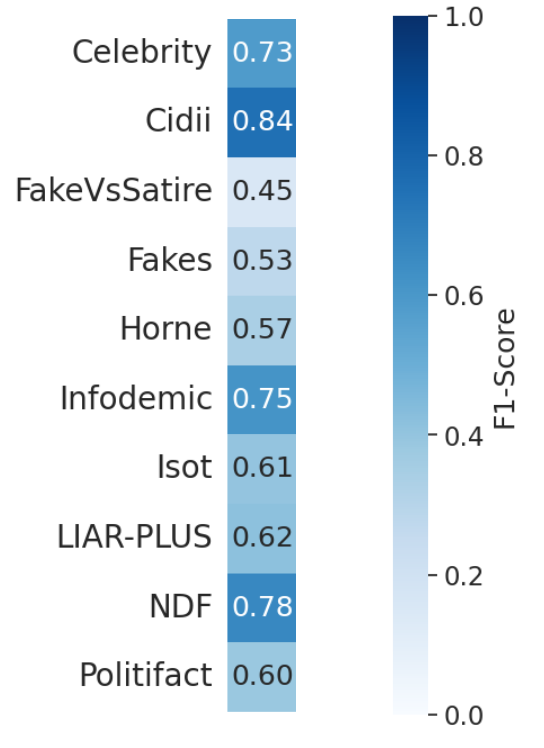}
    \caption{Zephyr-7B-beta (ZS)}
    \label{fig:headZephyr}
\end{subfigure}
\caption{F1-scores obtained in a zero-shot setting for each dataset by LLaMa3-8B (a), Qwen3-32B (b) and Zephyr-7B-beta (c).}
\label{fig:llm_zero_shot}
\end{figure}

\begin{figure}[H]
\centering
\begin{subfigure}{0.47\textwidth}
    \centering
    \includegraphics[width=\linewidth]{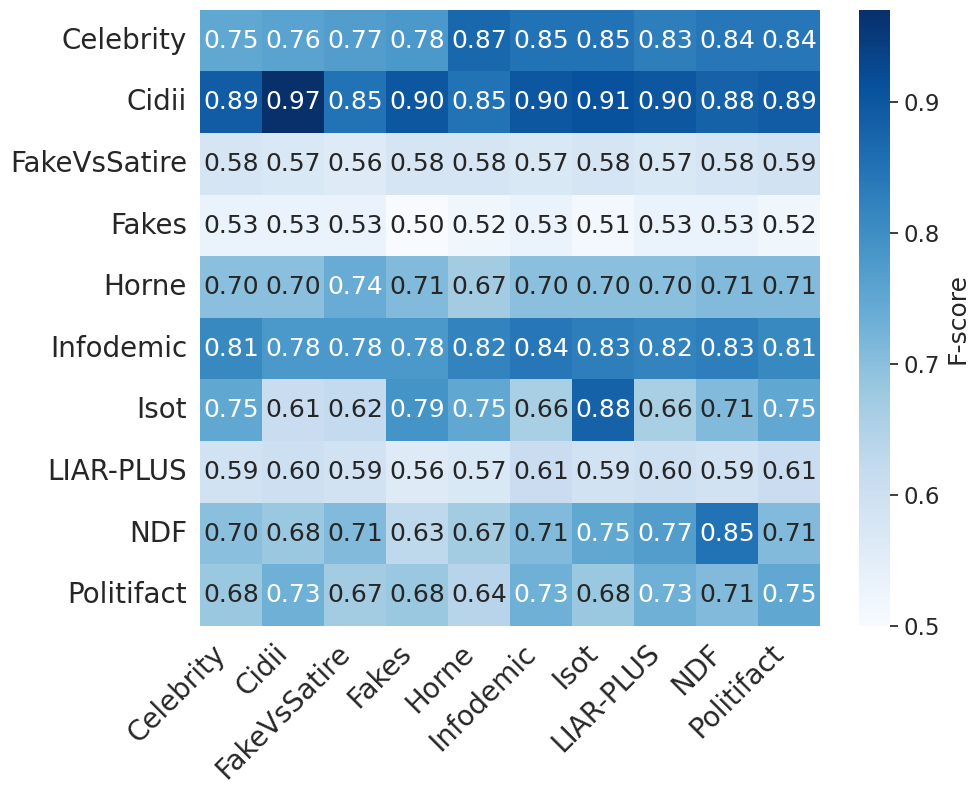}
    \caption{Llama3-8B (FS)}
    \label{fig:headLlamaFS}
\end{subfigure}
\hfill
\begin{subfigure}{0.47\textwidth}
    \centering
    \includegraphics[width=\linewidth]{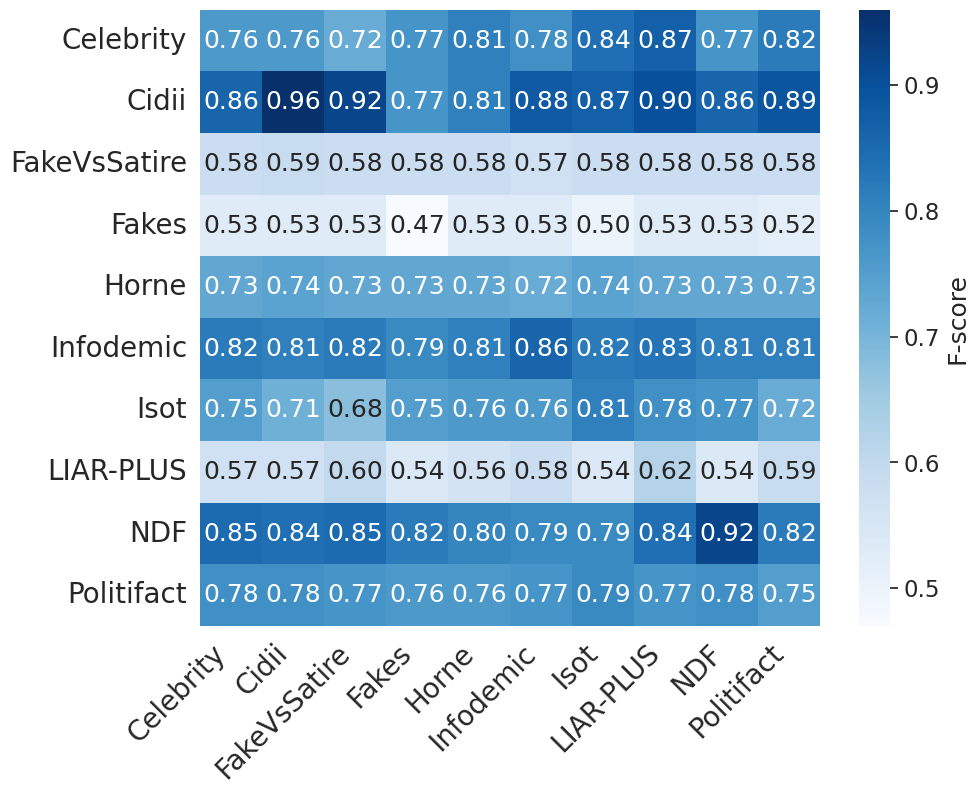}
    \caption{Qwen3-32B (FS)}
    \label{fig:headQwenFS}
\end{subfigure}

\medskip

\begin{subfigure}{0.47\textwidth}
    \centering
    \includegraphics[width=\linewidth]{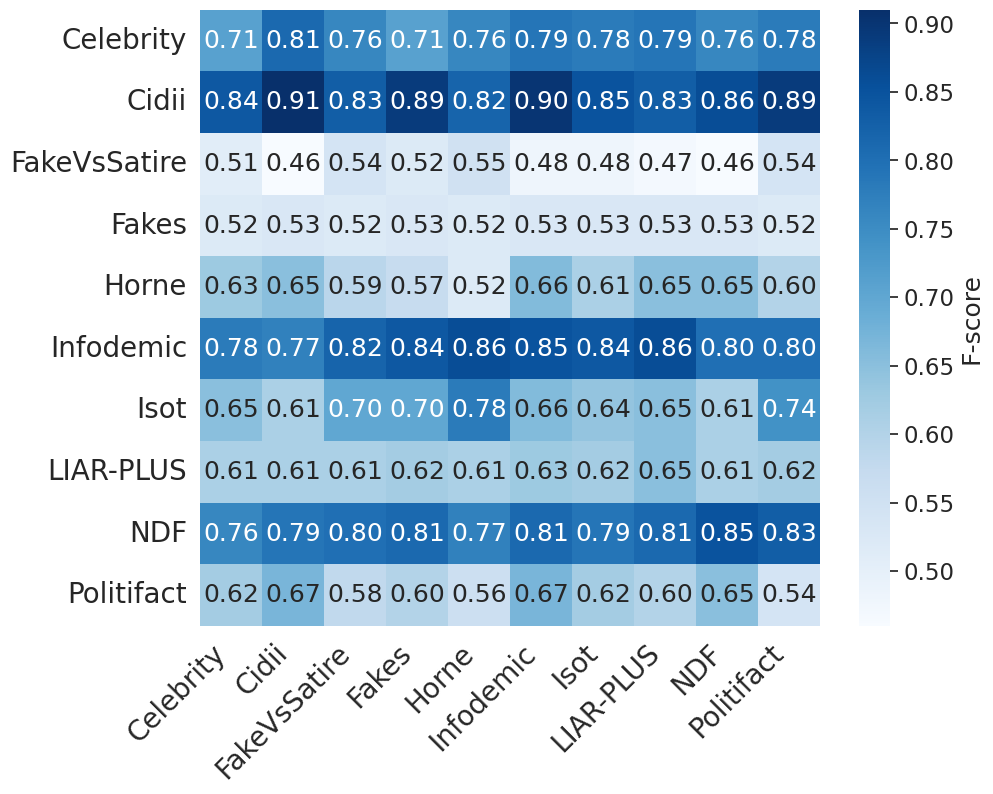}
    \caption{Zephyr-7B-beta (FS)}
    \label{fig:headZephyrFS}
\end{subfigure}
\caption{F1-scores obtained in a few-shot setting for each dataset by LLaMa3-8B (a), Qwen3-32B (b) and Zephyr-7B-beta (c).}
\label{fig:llm_few_shot}
\end{figure}

Figures \ref{fig:main_ML}--\ref{fig:llm_few_shot} present the performance obtained by using the traditional ML approaches, the DL approaches trained from scratch, and the Transformer-based approaches. Each sub-figure is a heatmap showing the F1-scores of a single model across the datasets. As a reference, the highest F1-score obtained by the models during the Dataset-Specific Experiments presented in the original paper are reported in the diagonal of the heatmap.

\subsection{Statistical tests}

\begin{figure}[!htb]
\centering
\includegraphics[width=0.8\linewidth]{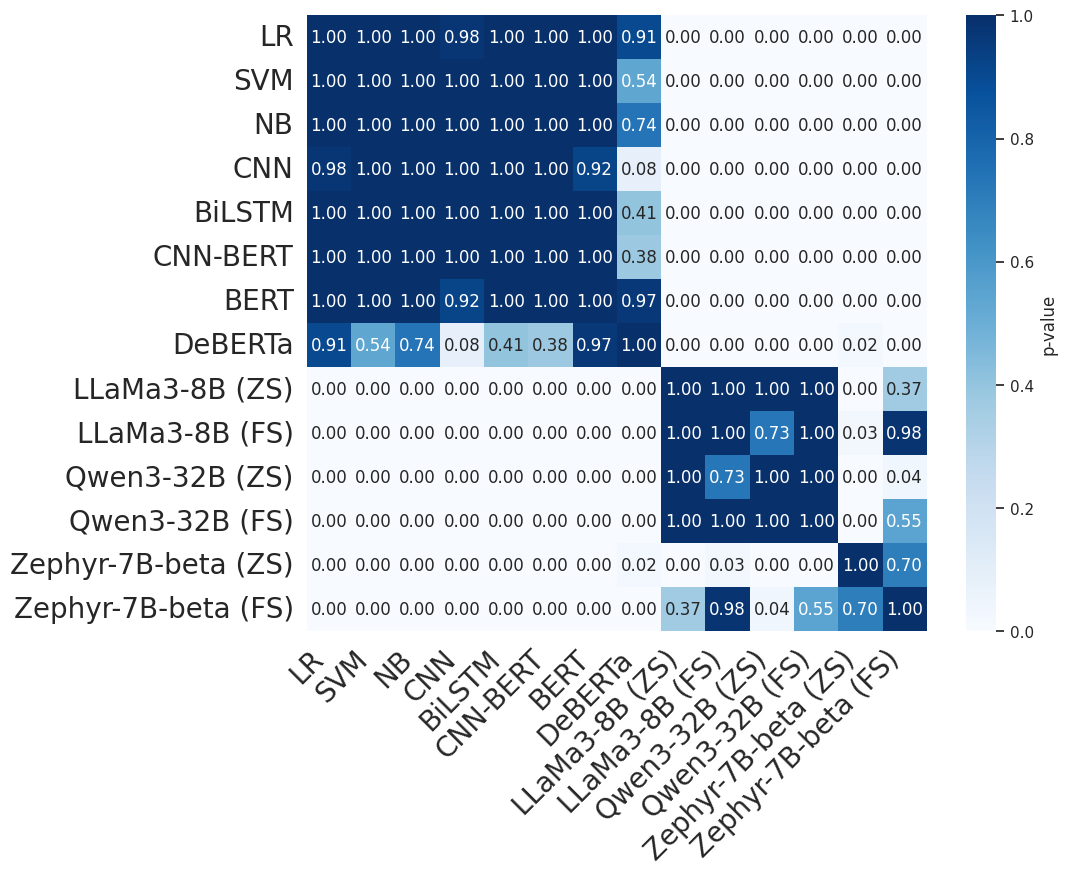} 
\caption{Heatmap of the p-values computed by the post-hoc Nemenyi Test (Pairwise Model Comparisons).}
\label{fig:heat2} 
\end{figure}

We obtained a Friedman test statistic of 608.31, a p-value of 1.29e-121 and a Kendall's W of 0.519. Since the p-value is below the significance threshold alpha=0.05, the post-hoc Nemenyi test is then performed to identify the specific pairs of models exhibiting statistically significant differences. Figure \ref{fig:heat2} reports the pairwise p-values obtained from the Nemenyi post-hoc test, where values below 0.05 indicate  statistically significant performance differences. 

We further computed whether statistically significant differences exist among the models per-stratum narrow-generic and articles-others. Results are reported in Table \ref{tab:statistical_stratification}. 

\begin{table}[htb]
\centering
\caption{Stratified Statistical Analysis (Friedman Test and Kendall's W Effect Size) across different dataset categories.}
\label{tab:statistical_stratification}
\begin{small}
\begin{tabular}{lcccc}
\toprule
\textbf{Stratum} & \textbf{Friedman Stat} & \textbf{p-value} & \textbf{Kendall's $W$} \\ \midrule
Generic          & 371.83                 & 1.96e-71 & 0.530  \\
Narrow           & 249.79                 & 7.08e-46 & 0.534 \\
Articles         & 380.00                 & 3.72e-73 & 0.541 \\
Others           & 233.63                 & 1.58e-42 & 0.499 \\ \bottomrule
\end{tabular}
\end{small}
\end{table}

Results show that the performance gap remains highly significant across all strata and the effect size is remarkably consistent.

\begin{figure}[!htb]
\centering
\includegraphics[width=0.8\linewidth]{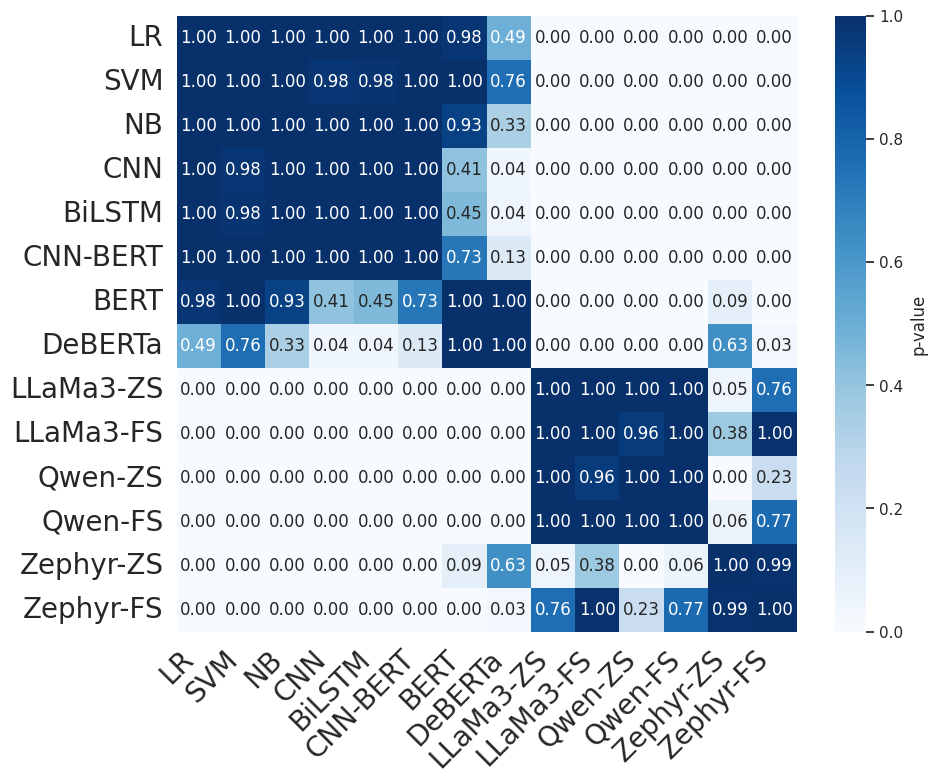} 
\caption{Heatmap of the p-values computed by the post-hoc Nemenyi Test (Pairwise Model Comparisons) on the Generic datasets (topic analysis).}
\label{fig:heat2.1} 
\end{figure}

\begin{figure}[!htb]
\centering
\includegraphics[width=0.8\linewidth]{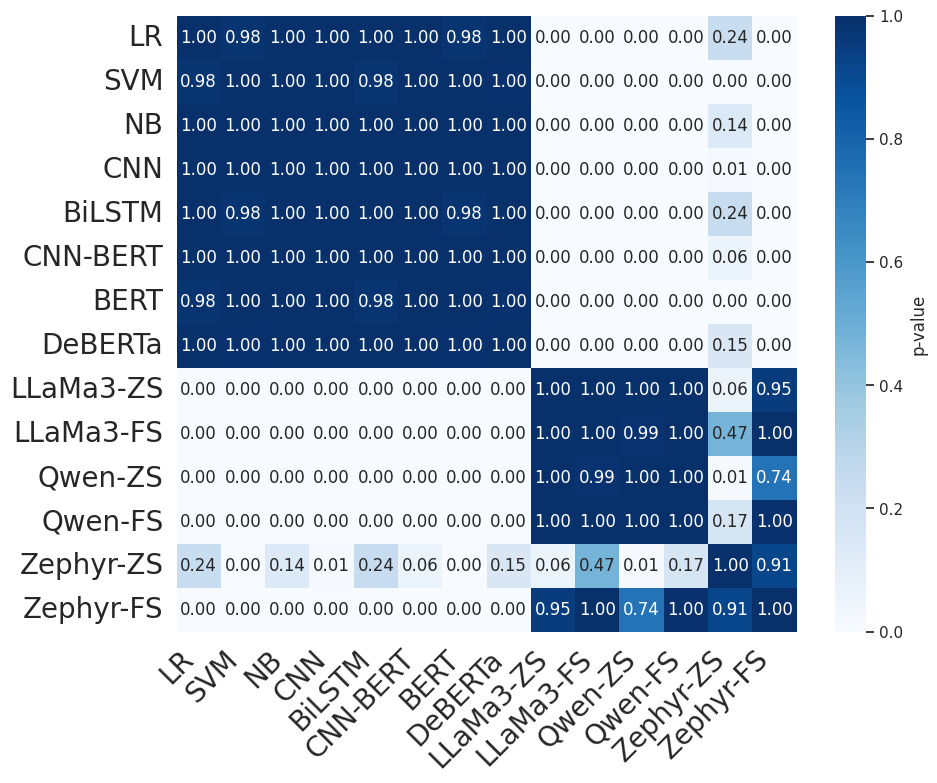} 
\caption{Heatmap of the p-values computed by the post-hoc Nemenyi Test (Pairwise Model Comparisons) on the Narrow datasets (topic analysis).}
\label{fig:heat2.2} 
\end{figure}

\begin{figure}[!htb]
\centering
\includegraphics[width=0.8\linewidth]{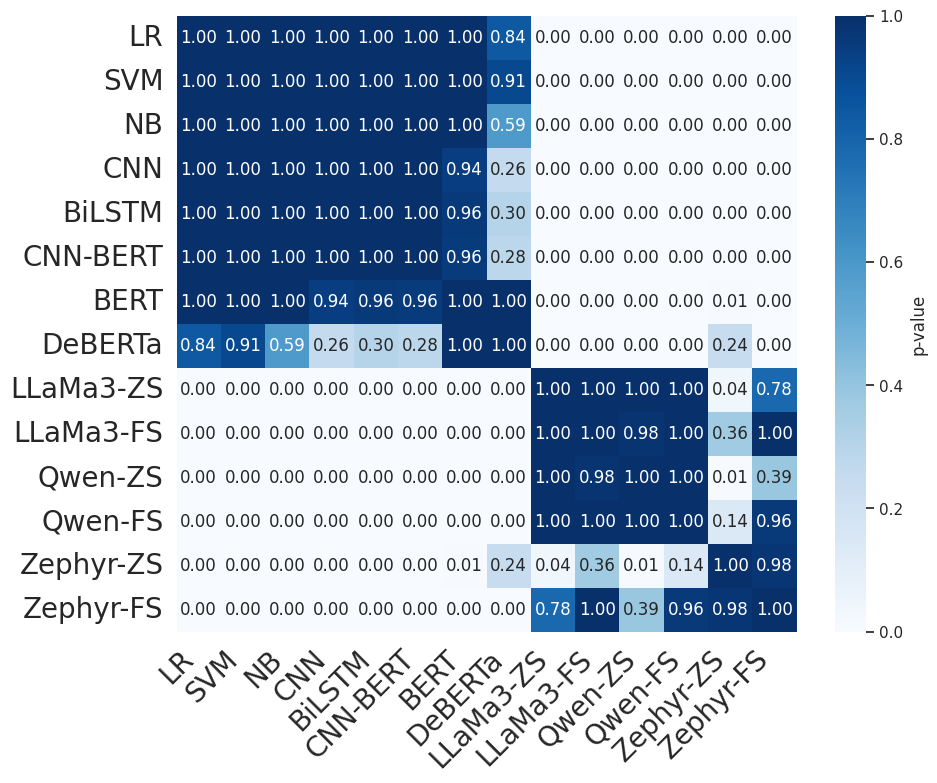} 
\caption{Heatmap of the p-values computed by the post-hoc Nemenyi Test (Pairwise Model Comparisons) on the Articles datasets (genre analysis).}
\label{fig:heat2.3} 
\end{figure}

\begin{figure}[!htb]
\centering
\includegraphics[width=0.8\linewidth]{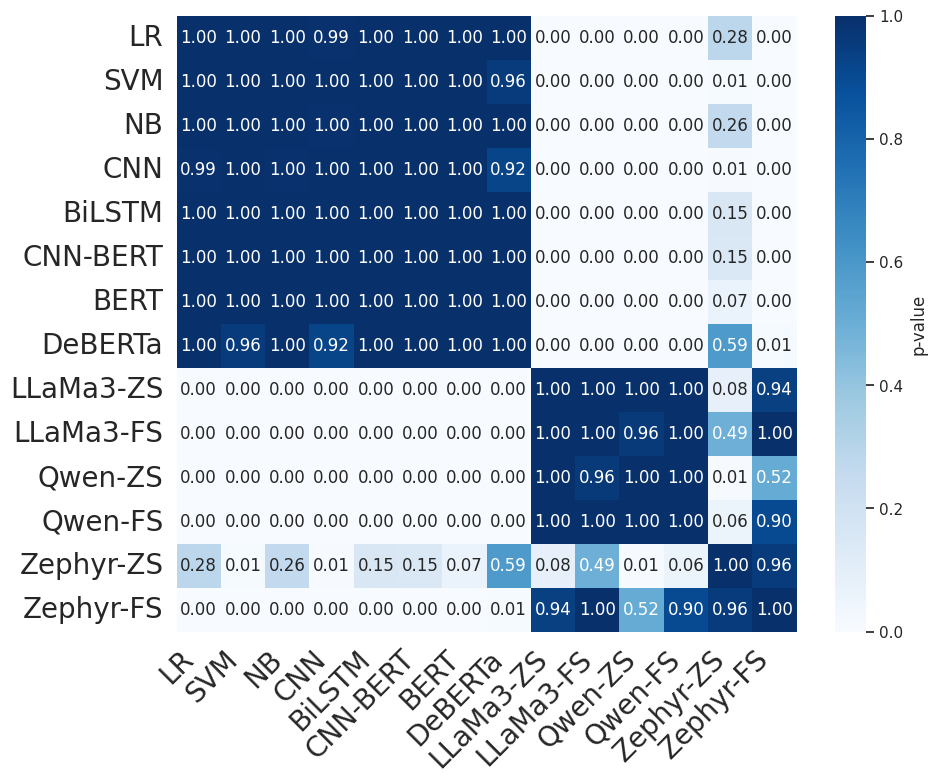} 
\caption{Heatmap of the p-values computed by the post-hoc Nemenyi Test (Pairwise Model Comparisons) on the Others datasets (genre analysis).}
\label{fig:heat2.4} 
\end{figure}

Figures \ref{fig:heat2.1},\ref{fig:heat2.2},  \ref{fig:heat2.3}, and \ref{fig:heat2.4} report the pairwise p-values obtained by the Nemenyi post-hoc test per-stratum, where values below 0.05 indicate  statistically significant performance differences. The relative behavior of decoder-only LLMs remains consistent across strata with the exception of Zephyr in its zero-shot setting.

\FloatBarrier

\section{Mixed-Training/Single-Test Experiments}\label{sec:appendix_exp3}
This section presents detailed experimental results and details on the
application of the statistical tests on the F1-score for the Mixed-Training/Single-Test
experiments.

\subsection{Detailed experimental results}

Tables \ref{tab:best_for_classes_ML_EXP3}--\ref{tab:best_for_classes_FEW_EXP3} show precision, recall and F1-score obtained by each model on each dataset for each class. 

\begin{table}[H]
\centering
\scriptsize
\caption{Precision, recall and F1-score obtained by the ML models on each dataset for each class.}
\label{tab:best_for_classes_ML_EXP3}
\begin{tabularx}{\textwidth}{@{}p{1.6cm} *{18}{Y} @{}}
\toprule
\textbf{Dataset} & \multicolumn{6}{c}{\textbf{LR}} & \multicolumn{6}{c}{\textbf{SVM}} & \multicolumn{6}{c}{\textbf{NB}} \\
\cmidrule(lr){2-7} \cmidrule(lr){8-13} \cmidrule(lr){14-19}
& \multicolumn{3}{c}{\textbf{Class 0}} & \multicolumn{3}{c}{\textbf{Class 1}} & \multicolumn{3}{c}{\textbf{Class 0}} & \multicolumn{3}{c}{\textbf{Class 1}} & \multicolumn{3}{c}{\textbf{Class 0}} & \multicolumn{3}{c}{\textbf{Class 1}} \\
\cmidrule(lr){2-4} \cmidrule(lr){5-7} \cmidrule(lr){8-10} \cmidrule(lr){11-13} \cmidrule(lr){14-16} \cmidrule(lr){17-19}
& P & R & F1 & P & R & F1 & P & R & F1 & P & R & F1 & P & R & F1 & P & R & F1\\
\midrule
Celebrity & .57 & .61 & .59 & .58 & .54 & .56 & .68 & .59 & .63 & .64 & .73 & .68 & .52 & .74 & .61 & .56 & .34 & .42 \\
Cidii & .87 & .87 & .87 & .79 & .80 & .80 & .88 & .77 & .82 & .69 & .83 & .75 & .77 & .84 & .80 & .70 & .6 & .64 \\
FakeVsSatire & .59 & .68 & .63 & .80 & .73 & .77 & .61 & .61 & .61 & .78 & .78 & .78 & .46 & .81 & .59 & .82 & .47 & .60 \\
Fakes & .53 & .50 & .51 & .45 & .48 & .46 & .56 & .55 & .56 & .48 & .49 & .49 & .53 & .59 & .56 & .45 & .39 & .42 \\
Horne & .86 & .77 & .81 & .30 & .43 & .35 & .86 & .66 & .75 & .26 & .52 & .34 & .84 & .86 & .85 & .33 & .30 & .31 \\
Infodemic & .74 & .81 & .78 & .77 & .70 & .73 & .76 & .79 & .78 & .76 & .73 & .74 & .65 & .74 & .69 & .66 & .55 & .60 \\
Isot & .62 & .83 & .71 & .77 & .53 & .63 & .68 & .84 & .75 & .81 & .64 & .71 & .52 & .83 & .64 & .64 & .28 & .39 \\
LIAR-PLUS & .57 & .61 & .59 & .46 & .43 & .45 & .57 & .68 & .62 & .48 & .36 & .41 & .57 & .62 & .59 & .46 & .41 & .43 \\
NDF & .86 & .86 & .86 & .72 & .71 & .71 & .85 & .85 & .85 & .70 & .70 & .70 & .80 & .70 & .75 & .51 & .65 & .57 \\
Politifact & .8 & .79 & .79 & .46 & .46 & .46 & .87 & .77 & .82 & .53 & .69 & .60 & .78 & .84 & .81 & .49 & .39 & .44 \\
Global test & .59 & .71 & .64 & .63 & .50 & .56 & .59 & .61 & .60 & .60 & .59 & .59 & .56 & .75 & .64 & .62 & .41 & .49 \\
\bottomrule
\end{tabularx}
\end{table}

\begin{table}[H]
\centering
\scriptsize
\caption{Precision, recall and F1-score obtained by the CNN, BiLSTM and CNN-BERT models on each dataset for each class.}
\label{tab:best_for_classes_CNN_BILSTM_COMPLETED_EXP3}
\begin{tabularx}{\textwidth}{@{}p{1.6cm} *{18}{Y} @{}}
\toprule
\textbf{Dataset} & \multicolumn{6}{c}{\textbf{CNN}} & \multicolumn{6}{c}{\textbf{BiLSTM}} & \multicolumn{6}{c}{\textbf{CNN-BERT}}\\
\cmidrule(lr){2-7} \cmidrule(lr){8-13} \cmidrule(lr){14-19}
& \multicolumn{3}{c}{\textbf{Class 0}} & \multicolumn{3}{c}{\textbf{Class 1}} & \multicolumn{3}{c}{\textbf{Class 0}} & \multicolumn{3}{c}{\textbf{Class 1}} & \multicolumn{3}{c}{\textbf{Class 0}} & \multicolumn{3}{c}{\textbf{Class 1}} \\
\cmidrule(lr){2-4} \cmidrule(lr){5-7} \cmidrule(lr){8-10} \cmidrule(lr){11-13} \cmidrule(lr){14-16} \cmidrule(lr){17-19}
& P & R & F1 & P & R & F1 & P & R & F1 & P & R & F1 & P & R & F1 & P & R & F1 \\
\midrule
Celebrity & .52 & .77 & .62 & .57 & .30 & .40 & .49 & .82 & .61 & .47 & .16 & .23 & .52 & .69 & .59 & .54 & .36 & .43 \\
Cidii & .64 & .85 & .73 & .51 & .25 & .33 & .65 & .87 & .75 & .57 & .26 & .36 & .62 & .61 & .62 & .39 & .40 & .39 \\
FakeVsSatire & .37 & .87 & .52 & .73 & .19 & .31 & .34 & .77 & .47 & .56 & .16 & .25 & .36 & .65 & .46 & .65 & .36 & .46 \\
Fakes & .54 & .89 & .67 & .50 & .12 & .20 & .55 & .66 & .60 & .48 & .36 & .41 & .55 & .69 & .61 & .49 & .34 & .40 \\
Horne & .82 & .73 & .77 & .20 & .30 & .24 & .79 & .82 & .80 & .05 & .04 & .04 & .83 & .76 & .79 & .22 & .30 & .25 \\
Infodemic & .52 & .70 & .59 & .47 & .29 & .36 & .54 & .78 & .64 & .52 & .26 & .35 & .57 & .55 & .56 & .53 & .55 & .54 \\
Isot & .50 & .84 & .63 & .61 & .23 & .33 & .48 & .84 & .61 & .54 & .17 & .26 & .55 & .69 & .61 & .62 & .48 & .54 \\
LIAR-PLUS & .57 & .77 & .65 & .47 & .26 & .34 & .56 & .80 & .66 & .47 & .22 & .30 & .55 & .46 & .50 & .43 & .51 & .47 \\
NDF & .69 & .78 & .73 & .39 & .29 & .33 & .67 & .72 & .70 & .33 & .28 & .30 & .67 & .59 & .63 & .33 & .42 & .37 \\
Politifact & .72 & .81 & .76 & .24 & .15 & .18 & .72 & .84 & .77 & .27 & .15 & .19 & .76 & .73 & .74 & .35 & .38 & .36 \\
Global test & .49 & .79 & .60 & .46 & .18 & .26 & .51 & .80 & .62 & .54 & .23 & .32 & .51 & .59 & .55 & .52 & .45 & .48 \\
\bottomrule
\end{tabularx}
\end{table}

\begin{table}[H]
\centering
\scriptsize
\caption{Precision, recall and F1-score obtained by the BERT, DeBERTa and MERMAID models on each dataset for each class.}
\label{tab:best_for_classes_BERT_MERMAID_EXP3_MODIFIED}
\begin{tabularx}{\textwidth}{@{}p{1.6cm} *{18}{Y} @{}}
\toprule
\textbf{Dataset} & \multicolumn{6}{c}{\textbf{BERT}} & \multicolumn{6}{c}{\textbf{DeBERTa}} & \multicolumn{6}{c}{\textbf{MERMAID}} \\
\cmidrule(lr){2-7} \cmidrule(lr){8-13} \cmidrule(lr){14-19}
& \multicolumn{3}{c}{\textbf{Class 0}} & \multicolumn{3}{c}{\textbf{Class 1}} & \multicolumn{3}{c}{\textbf{Class 0}} & \multicolumn{3}{c}{\textbf{Class 1}} & \multicolumn{3}{c}{\textbf{Class 0}} & \multicolumn{3}{c}{\textbf{Class 1}} \\
\cmidrule(lr){2-4} \cmidrule(lr){5-7} \cmidrule(lr){8-10} \cmidrule(lr){11-13} \cmidrule(lr){14-16} \cmidrule(lr){17-19}
& P & R & F1 & P & R & F1 & P & R & F1 & P & R & F1 & P & R & F1 & P & R & F1\\
\midrule
Celebrity & .85 & .66 & .74 & .72 & .88 & .79 & .83 & .83 & .83 & .83 & .83 & .83 & .59 & .84 & .69 & .72 & .41 & .52 \\
Cidii & .98 & .84 & .90 & .79 & .97 & .87 & .98 & .92 & .95 & .89 & .97 & .93 & .15 & .22 & .18 & .38 & .26 & .31 \\
FakeVsSatire & .50 & .80 & .61 & .83 & .54 & .66 & .68 & .75 & .71 & .85 & .80 & .82 & .60 & .65 & .63 & .85 & .82 & .83 \\
Fakes & .54 & .73 & .62 & .47 & .28 & .35 & .62 & .17 & .26 & .47 & .88 & .62 & .52 & .51 & .52 & .43 & .44 & .43 \\
Horne & .92 & .70 & .80 & .36 & .73 & .48 & .92 & .91 & .92 & .62 & .65 & .64 & .98 & .67 & .80 & .07 & .66 & .12 \\
Infodemic & .89 & .85 & .87 & .84 & .89 & .86 & .88 & .91 & .90 & .90 & .87 & .88 & .80 & .92 & .85 & .89 & .77 & .84 \\
Isot & .90 & .99 & .95 & .99 & .90 & .94 & .96 & 1.0 & .98 & 1.0 & .96 & .98 & .80 & .92 & .85 & .91 & .78 & .84 \\
LIAR-PLUS & .64 & .51 & .56 & .50 & .63 & .56 & .66 & .53 & .59 & .52 & .65 & .58 & .68 & .38 & .49 & .50 & .78 & .61 \\
NDF & .93 & .74 & .83 & .63 & .89 & .74 & .94 & .89 & .91 & .79 & .89 & .84 & .86 & .71 & .78 & .47 & .69 & .56 \\
Politifact & .90 & .86 & .88 & .68 & .75 & .72 & .87 & .94 & .90 & .78 & .61 & .69 & .88 & .90 & .89 & .48 & .42 & .45 \\
Global test & .68 & .71 & .69 & .69 & .67 & .68 & .73 & .72 & .73 & .72 & .73 & .73 & .65 & .69 & .67 & .69 & .62 & .64 \\
\bottomrule
\end{tabularx}
\end{table}

\begin{table}[H]
\centering
\scriptsize
\caption{Precision, recall and F1-score obtained by the Llama3-8B, Qwen3-32B and Zephyr-7B-beta models on each dataset for each class in a zero-shot (ZS) scenario.}
\label{tab:best_for_classes_Zero_EXP3}
\begin{tabularx}{\textwidth}{@{}p{1.6cm} *{18}{Y} @{}}
\toprule
\textbf{Dataset} & \multicolumn{6}{c}{\textbf{Llama3-8B (ZS)}} & \multicolumn{6}{c}{\textbf{Qwen3-32B (ZS)}} & \multicolumn{6}{c}{\textbf{Zephyr-7B-beta (ZS)}} \\
\cmidrule(lr){2-7} \cmidrule(lr){8-13} \cmidrule(lr){14-19}
& \multicolumn{3}{c}{\textbf{Class 0}} & \multicolumn{3}{c}{\textbf{Class 1}} & \multicolumn{3}{c}{\textbf{Class 0}} & \multicolumn{3}{c}{\textbf{Class 1}} & \multicolumn{3}{c}{\textbf{Class 0}} & \multicolumn{3}{c}{\textbf{Class 1}} \\
\cmidrule(lr){2-4} \cmidrule(lr){5-7} \cmidrule(lr){8-10} \cmidrule(lr){11-13} \cmidrule(lr){14-16} \cmidrule(lr){17-19}
& P & R & F1 & P & R & F1 & P & R & F1 & P & R & F1 & P & R & F1 & P & R & F1\\
\midrule
Celebrity & .84  & .91  & .87  & .90  & .82  & .86  & .86 & .83 & .84 & .84 & .86 & .85 & .68 & .87 & .76 & .86 & .58 & .70 \\
Cidii & .92  & .94  & .93  & .90  & .84  & .87  & .95 & .92 & .94 & .87 & .92 & .90 & .81 & 1.00 & .89 & 1.00 & .59 & .75 \\
FakeVsSatire & .00  & .00  & .00  & .66  & .99  & .79  & .00 & .00 & .00 & .66 & 1.00 & .80 & .08 & .04 & .05 & .61 & .76 & .68 \\
Fakes & .54  & .99  & .70  & .00  & .00  & .00  & .54 & .90 & .67 & .40 & .08 & .14 & .54 & .99 & .70 & .50 & .01 & .02 \\
Horne & .97  & .56  & .71  & .32  & .91  & .48  & .95 & .54 & .69 & .30 & .87 & .44 & .84 & .48 & .61 & .22 & .61 & .32 \\
Infodemic & .71  & .98  & .82  & .96  & .56  & .70  & .79 & .92 & .85 & .89 & .74 & .81 & .68 & 1.00 & .81 & .99 & .48 & .65 \\
Isot & .57 & 1.00 & .73 & 1.00 & .30 & .46 & .66 & 1.00 & .79 & .99 & .52 & .68 & .56 & .99 & .72 & .96 & .26 & .41 \\
LIAR-PLUS & .61 & .84 & .71 & .62 & .32 & .42 & .67 & .50 & .57 & .52 & .68 & .59 & .61 & .90 & .73 & .69 & .25 & .37 \\
NDF & .85 & .93 & .88 & .79 & .61 & .69 & .91 & .89 & .90 & .75 & .80 & .78 & .83 & .95 & .89 & .83 & .55 & .66 \\
Politifact & .82 & .84 & .83 & .53 & .41 & .46 & .85 & .96 & .90 & .76 & .44 & .56 & .79 & .71 & .75 & .61 & .40 & .48 \\
Global & .60 & .71 & .65 & .66 & .53 & .58 & .63 & .68 & .65 & .65 & .61 & .63 & .53 & .70 & .60 & .58 & .38 & .46 \\
\bottomrule
\end{tabularx}

\end{table}

\begin{table}[H]
\centering
\scriptsize
\caption{Precision, recall and F1-score obtained by the Llama3-8B, Qwen3-32B and Zephyr-7B-beta models on each dataset for each class in a few-shot (FS) scenario.}
\label{tab:best_for_classes_FEW_EXP3}
\begin{tabularx}{\textwidth}{@{}p{1.6cm} *{18}{Y} @{}}
\toprule
\textbf{Dataset} & \multicolumn{6}{c}{\textbf{Llama3-8B (FS)}} & \multicolumn{6}{c}{\textbf{Qwen3-32B (FS)}} & \multicolumn{6}{c}{\textbf{Zephyr-7B-beta (FS)}} \\
\cmidrule(lr){2-7} \cmidrule(lr){8-13} \cmidrule(lr){14-19}
& \multicolumn{3}{c}{\textbf{Class 0}} & \multicolumn{3}{c}{\textbf{Class 1}} & \multicolumn{3}{c}{\textbf{Class 0}} & \multicolumn{3}{c}{\textbf{Class 1}} & \multicolumn{3}{c}{\textbf{Class 0}} & \multicolumn{3}{c}{\textbf{Class 1}} \\
\cmidrule(lr){2-4} \cmidrule(lr){5-7} \cmidrule(lr){8-10} \cmidrule(lr){11-13} \cmidrule(lr){14-16} \cmidrule(lr){17-19}
& P & R & F1 & P & R & F1 & P & R & F1 & P & R & F1 & P & R & F1 & P & R & F1\\
\midrule
Celebrity & .89 & .82 & .86 & .84 & .90 & .87 & .79 & .85 & .82 & .84 & .78 & .81 & .66 & .87 & .75 & .83 & .54 & .65 \\
Cidii & .93 & .93 & .93 & .88 & .87 & .87 & .98 & .80 & .88 & .75 & .98 & .85 & .84 & .95 & .89 & .90 & .70 & .79 \\
FakeVsSatire & .00 & .00 & .00 & .66 & .99 & .69 & .50 & .01 & .02 & .66 & .99 & .80 & .07 & .04 & .05 & .61 & .74 & .67 \\
Fakes & .54 & .99 & .70 & .33 & .00 & .01 & .53 & .93 & .68 & .28 & .03 & .06 & .54 & .99 & .70 & .50 & .01 & .02 \\
Horne & .97 & .57 & .72 & .33 & .91 & .48 & .97 & .58 & .73 & .33 & .91 & .48 & .87 & .53 & .66 & .27 & .65 & .38 \\
Infodemic & .77 & .95 & .85 & .93 & .69 & .79 & .81 & .87 & .84 & .84 & .78 & .81 & .72 & 1.00 & .84 & .99 & .57 & .72 \\
Isot & .63 & 1.00 & .77 & .99 & .46 & .63 & .65 & 1.00 & .79 & 1.00 & .50 & .66 & .54 & .97 & .70 & .91 & .22 & .36 \\
LIAR-PLUS & .65 & .58 & .61 & .53 & .60 & .56 & .71 & .44 & .55 & .52 & .77 & .62 & .63 & .82 & .71 & .64 & .37 & .47 \\
NDF & .82 & .81 & .81 & .58 & .60 & .59 & .91 & .89 & .90 & .77 & .79 & .78 & .83 & .92 & .87 & .77 & .56 & .65 \\
Politifact & .82 & .74 & .78 & .57 & .49 & .53 & .85 & .86 & .86 & .54 & .52 & .53 & .78 & .60 & .68 & .60 & .44 & .51 \\
Global & .57 & .66 & .61 & .60 & .49 & .54 & .66 & .67 & .66 & .66 & .66 & .66 & .54 & .69 & .60 & .59 & .40 & .47 \\
\bottomrule
\end{tabularx}
\end{table}

\subsection{Statistical tests}

\begin{figure}[!htb]
\centering
\includegraphics[width=0.8\linewidth]{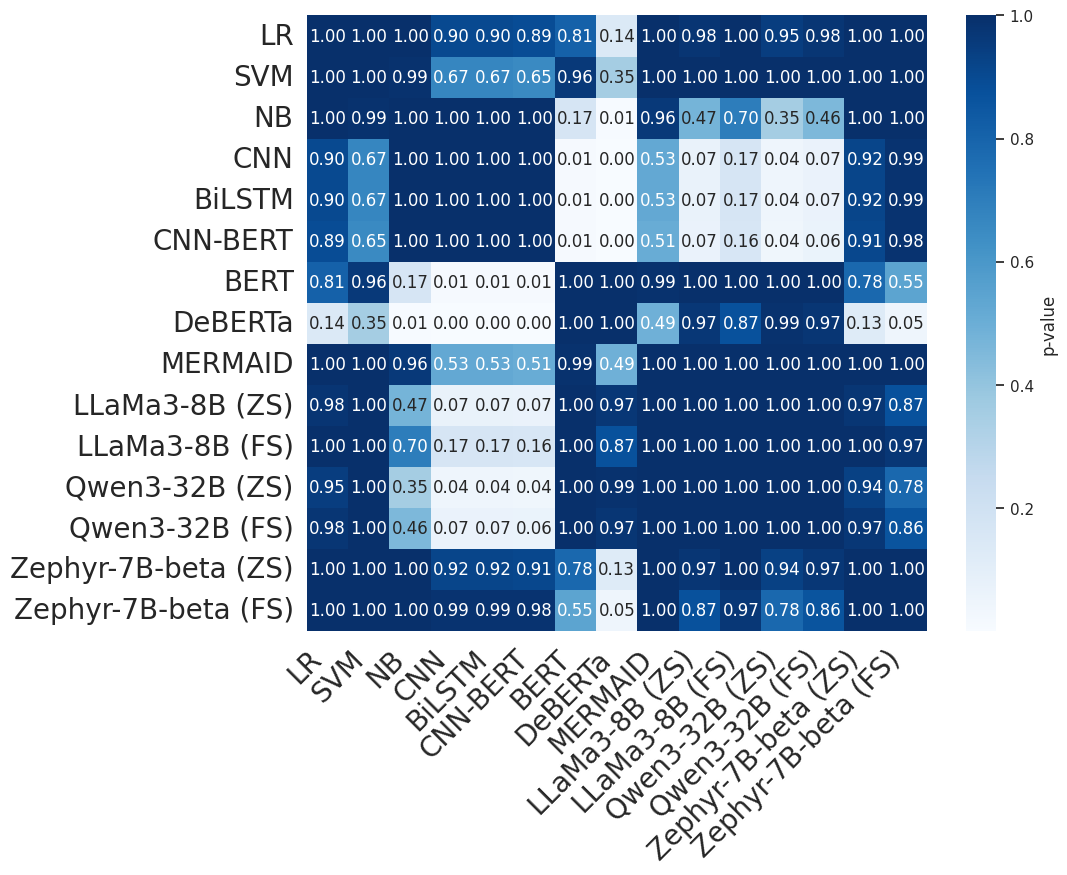} 
\caption{Heatmap of the p-values computed by the post-hoc Nemenyi Test (Pairwise Model Comparisons).}
\label{fig:heat3} 
\end{figure}

We obtained a Friedman Test statistic of 63.12, a p-value of 3.3e-08, and a Kendall's W of 0.410. Since the p-value is below the significance threshold alpha=0.05, the post-hoc Nemenyi test is then performed to identify the specific pairs of models exhibiting statistically significant differences. Figure \ref{fig:heat3} reports the pairwise p-values obtained from the Nemenyi post-hoc test, where values below 0.05 indicate  statistically significant performance differences.

\FloatBarrier

\section{Leave-One-Dataset-Out Experiments}\label{sect:appendix_exp4}

This section presents detailed experimental results and details on the ap-
plication of the statistical tests on the F1-score for the Leave-One-Dataset-Out experiments.

\subsection{Detailed experimental results}

Tables \ref{tab:best_for_classes_ML_EXP4}--\ref{tab:best_for_classes_FEW_EXP4} show precision, recall and F1-score Obtained by each model on each dataset for each class. These detailed metrics provide insight into which specific classes are misclassified when the model encounters a completely unseen domain.

\begin{table}[H]
\centering
\scriptsize
\caption{Precision, recall and F1-score obtained by the ML models on each dataset for each class.}
\label{tab:best_for_classes_ML_EXP4}
\begin{tabularx}{\textwidth}{@{}p{1.6cm} *{18}{Y} @{}}
\toprule
\textbf{Dataset} & \multicolumn{6}{c}{\textbf{LR}} & \multicolumn{6}{c}{\textbf{SVM}} & \multicolumn{6}{c}{\textbf{NB}} \\
\cmidrule(lr){2-7} \cmidrule(lr){8-13} \cmidrule(lr){14-19}
& \multicolumn{3}{c}{\textbf{Class 0}} & \multicolumn{3}{c}{\textbf{Class 1}} & \multicolumn{3}{c}{\textbf{Class 0}} & \multicolumn{3}{c}{\textbf{Class 1}} & \multicolumn{3}{c}{\textbf{Class 0}} & \multicolumn{3}{c}{\textbf{Class 1}} \\
\cmidrule(lr){2-4} \cmidrule(lr){5-7} \cmidrule(lr){8-10} \cmidrule(lr){11-13} \cmidrule(lr){14-16} \cmidrule(lr){17-19}
& P & R & F1 & P & R & F1 & P & R & F1 & P & R & F1 & P & R & F1 & P & R & F1\\
\midrule
Celebrity & .58 & .27 & .37 & .52 & .81 & .64 & .54 & .12 & .19 & .50 & .90 & .65 & .60 & .01 & .02 & .50 & .99 & .67 \\
Cidii & .65 & .46 & .54 & .46 & .64 & .54 & .68 & .32 & .43 & .45 & .79 & .57 & .75 & .05 & .09 & .42 & .98 & .59 \\
FakeVsSatire & .45 & .51 & .48 & .61 & .54 & .57 & .45 & .38 & .41 & .60 & .66 & .63 & .42 & .18 & .25 & .58 & .82 & .68 \\
Fakes & .54 & .62 & .57 & .48 & .40 & .43 & .54 & .65 & .59 & .50 & .38 & .43 & .53 & .99 & .69 & .50 & .01 & .03 \\
Horne & .73 & .54 & .62 & .47 & .67 & .55 & .71 & .53 & .61 & .45 & .64 & .53 & .74 & .38 & .50 & .43 & .78 & .56 \\
Infodemic & .56 & .40 & .47 & .50 & .65 & .57 & .54 & .49 & .51 & .49 & .55 & .52 & .60 & .44 & .51 & .52 & .68 & .59 \\
Isot & .51 & .71 & .59 & .56 & .36 & .44 & .53 & .55 & .54 & .57 & .55 & .56 & .65 & .43 & .51 & .59 & .78 & .67 \\
LIAR-PLUS & .59 & .32 & .42 & .46 & .72 & .56 & .63 & .09 & .15 & .45 & .94 & .61 & .60 & .49 & .54 & .47 & .58 & .52 \\
NDF & .76 & .37 & .50 & .45 & .82 & .58 & .69 & .40 & .50 & .43 & .71 & .54 & .50 & .03 & .06 & .39 & .95 & .55 \\
Politifact & .64 & .55 & .59 & .37 & .47 & .41 & .61 & .47 & .53 & .34 & .48 & .40 & .71 & .41 & .52 & .41 & .71 & .52 \\
\bottomrule
\end{tabularx}
\end{table}

\begin{table}[H]
\centering
\scriptsize
\caption{Precision, recall and F1-score obtained by the CNN, BiLSTM and CNN-BERT models on each dataset for each class.}
\label{tab:best_for_classes_CNN_BILSTM_COMPLETED_EXP4}
\begin{tabularx}{\textwidth}{@{}p{1.6cm} *{18}{Y} @{}}
\toprule
\textbf{Dataset} & \multicolumn{6}{c}{\textbf{CNN}} & \multicolumn{6}{c}{\textbf{BiLSTM}} & \multicolumn{6}{c}{\textbf{CNN-BERT}}\\
\cmidrule(lr){2-7} \cmidrule(lr){8-13} \cmidrule(lr){14-19}
& \multicolumn{3}{c}{\textbf{Class 0}} & \multicolumn{3}{c}{\textbf{Class 1}} & \multicolumn{3}{c}{\textbf{Class 0}} & \multicolumn{3}{c}{\textbf{Class 1}} & \multicolumn{3}{c}{\textbf{Class 0}} & \multicolumn{3}{c}{\textbf{Class 1}} \\
\cmidrule(lr){2-4} \cmidrule(lr){5-7} \cmidrule(lr){8-10} \cmidrule(lr){11-13} \cmidrule(lr){14-16} \cmidrule(lr){17-19}
& P & R & F1 & P & R & F1 & P & R & F1 & P & R & F1 & P & R & F1 & P & R & F1 \\
\midrule
Celebrity & .58 & .38 & .46 & .54 & .72 & .62 & .72 & .20 & .31 & .53 & .92 & .68 & .64 & .21 & .32 & .53 & .88 & .66 \\
Cidii & .79 & .18 & .29 & .45 & .93 & .60 & .78 & .08 & .14 & .43 & .97 & .59 & .61 & .17 & .27 & .42 & .84 & .56 \\
FakeVsSatire & .59 & .06 & .12 & .59 & .97 & .73 & .47 & .17 & .25 & .59 & .86 & .70 & .52 & .30 & .38 & .61 & .80 & .69 \\
Fakes & .51 & .61 & .56 & .44 & .34 & .38 & .49 & .29 & .36 & .45 & .67 & .54 & .54 & .58 & .56 & .48 & .45 & .46 \\
Horne & .86 & .32 & .47 & .45 & .91 & .60 & .87 & .23 & .37 & .43 & .94 & .59 & .82 & .55 & .66 & .52 & .80 & .63 \\
Infodemic & .43 & .15 & .22 & .46 & .78 & .58 & .68 & .08 & .14 & .49 & .96 & .65 & .43 & .21 & .29 & .44 & .69 & .54 \\
Isot & .59 & .48 & .53 & .59 & .69 & .63 & .56 & .59 & .57 & .59 & .56 & .58 & .46 & .27 & .34 & .51 & .70 & .59 \\
LIAR-PLUS & .70 & .10 & .17 & .45 & .95 & .61 & .69 & .01 & .03 & .44 & .99 & .61 & .64 & .12 & .20 & .45 & .91 & .60 \\
NDF & .68 & .04 & .07 & .39 & .97 & .56 & .81 & .10 & .18 & .41 & .96 & .57 & .46 & .09 & .15 & .37 & .84 & .51 \\
Politifact & .72 & .51 & .60 & .43 & .66 & .52 & .68 & .61 & .64 & .42 & .51 & .46 & .76 & .48 & .59 & .44 & .73 & .55 \\
\bottomrule
\end{tabularx}
\end{table}

\begin{table}[H]
\centering
\scriptsize
\caption{Precision, recall and F1-score obtained by the BERT, DeBERTa and MERMAID models on each dataset for each class.}
\label{tab:best_for_classes_BERT_MERMAID_EXP4}
\begin{tabularx}{\textwidth}{@{}p{1.6cm} *{18}{Y} @{}}
\toprule
\textbf{Dataset} & \multicolumn{6}{c}{\textbf{BERT}} & \multicolumn{6}{c}{\textbf{DeBERTa}} & \multicolumn{6}{c}{\textbf{MERMAID}} \\
\cmidrule(lr){2-7} \cmidrule(lr){8-13} \cmidrule(lr){14-19}
& \multicolumn{3}{c}{\textbf{Class 0}} & \multicolumn{3}{c}{\textbf{Class 1}} & \multicolumn{3}{c}{\textbf{Class 0}} & \multicolumn{3}{c}{\textbf{Class 1}} & \multicolumn{3}{c}{\textbf{Class 0}} & \multicolumn{3}{c}{\textbf{Class 1}} \\
\cmidrule(lr){2-4} \cmidrule(lr){5-7} \cmidrule(lr){8-10} \cmidrule(lr){11-13} \cmidrule(lr){14-16} \cmidrule(lr){17-19}
& P & R & F1 & P & R & F1 & P & R & F1 & P & R & F1 & P & R & F1 & P & R & F1\\
\midrule
Celebrity & .59 & .49 & .53 & .56 & .65 & .60 & .72 & .48 & .58 & .61 & .82 & .70 & .71 & .63 & .67 & .67 & .74 & .70 \\
Cidii & .68 & .16 & .26 & .43 & .90 & .58 & .80 & .05 & .09 & .42 & .98 & .59 & .73 & .17 & .28 & .42 & .91 & .58 \\
FakeVsSatire & .62 & .29 & .39 & .63 & .88 & .73 & .58 & .13 & .21 & .60 & .93 & .73 & .60 & .17 & .26 & .63 & .93 & .75 \\
Fakes & .53 & .32 & .40 & .47 & .67 & .55 & .55 & .37 & .44 & .48 & .67 & .56 & .54 & .91 & .68 & .53 & .12 & .19 \\
Horne & .85 & .72 & .78 & .63 & .79 & .70 & .79 & .80 & .80 & .66 & .64 & .65 & .96 & .78 & .86 & .67 & .93 & .78 \\
Infodemic & .58 & .26 & .36 & .50 & .80 & .61 & .73 & .37 & .49 & .55 & .85 & .67 & .40 & .00 & .00 & .48 & .99 & .64 \\
Isot & .65 & .93 & .77 & .89 & .54 & .67 & .77 & .91 & .83 & .90 & .75 & .82 & .84 & .62 & .71 & .71 & .89 & .79 \\
LIAR-PLUS & .68 & .04 & .08 & .45 & .98 & .61 & .65 & .05 & .10 & .45 & .96 & .61 & .68 & .11 & .19 & .45 & .93 & .61 \\
NDF & .74 & .07 & .13 & .40 & .96 & .56 & .83 & .17 & .28 & .42 & .94 & .58 & .68 & .87 & .77 & .57 & .30 & .39 \\
Politifact & .72 & .63 & .67 & .47 & .57 & .51 & .72 & .79 & .76 & .56 & .48 & .51 & .80 & .95 & .87 & .84 & .53 & .65 \\
\bottomrule
\end{tabularx}
\end{table}

\begin{table}[H]
\centering
\scriptsize
\caption{Precision, recall and F1-score obtained by the Llama3-8B, Qwen3-32B and Zephyr-7B-beta models on each dataset for each class in a zero-shot (ZS) scenario.}
\label{tab:best_for_classes_Zero_EXP4}
\begin{tabularx}{\textwidth}{@{}p{1.6cm} *{18}{Y} @{}}
\toprule
\textbf{Dataset} & \multicolumn{6}{c}{\textbf{Llama3-8B (ZS)}} & \multicolumn{6}{c}{\textbf{Qwen3-32B (ZS)}} & \multicolumn{6}{c}{\textbf{Zephyr-7B-beta (ZS)}} \\
\cmidrule(lr){2-7} \cmidrule(lr){8-13} \cmidrule(lr){14-19}
& \multicolumn{3}{c}{\textbf{Class 0}} & \multicolumn{3}{c}{\textbf{Class 1}} & \multicolumn{3}{c}{\textbf{Class 0}} & \multicolumn{3}{c}{\textbf{Class 1}} & \multicolumn{3}{c}{\textbf{Class 0}} & \multicolumn{3}{c}{\textbf{Class 1}} \\
\cmidrule(lr){2-4} \cmidrule(lr){5-7} \cmidrule(lr){8-10} \cmidrule(lr){11-13} \cmidrule(lr){14-16} \cmidrule(lr){17-19}
& P & R & F1 & P & R & F1 & P & R & F1 & P & R & F1 & P & R & F1 & P & R & F1\\
\midrule
Celebrity & .82  & .90  & .86  & .91  & .81  & .85  & .83 & .81 & .82 & .81 & .84 & .83 & .67 & .88 & .76 & .87 & .57 & .69 \\
Cidii & .89  & .95  & .92  & .93  & .82  & .87  & .93 & .93 & .93 & .91 & .91 & .91 & .79 & 1.00 & .88 & 1.00 & .62 & .76 \\
FakeVsSatire & .38  & .01  & .03  & .58  & .98  & .73  & .25 & .01 & .02 & .58 & .98 & .73 & .13 & .05 & .08 & .52 & .73 & .61 \\
Fakes & .53  & .99  & .69  & .00  & .00  & .00  & .53 & .90 & .67 & .45 & .09 & .15 & .53 & .98 & .69 & .36 & .01 & .02 \\
Horne & .94  & .57  & .71  & .58  & .94  & .72  & .94 & .59 & .72 & .58 & .94 & .72 & .71 & .51 & .59 & .47 & .65 & .54 \\
Infodemic & .71  & .98  & .82  & .96  & .56  & .70  & .79 & .91 & .85 & .89 & .74 & .81 & .68 & 1.00 & .81 & .99 & .48 & .65 \\
Isot & .65 & 1.00 & .79 & 1.00 & .49 & .65 & .66 & 1.00 & .79 & .99 & .52 & .68 & .56 & .99 & .72 & .96 & .26 & .41 \\
LIAR-PLUS & .61 & .84 & .71 & .62 & .33 & .43 & .67 & .51 & .58 & .52 & .68 & .59 & .61 & .90 & .73 & .69 & .25 & .37 \\
NDF & .77 & .92 & .84 & .82 & .58 & .68 & .85 & .87 & .86 & .79 & .75 & .77 & .65 & .94 & .84 & .85 & .50 & .63 \\
Politifact & .77 & .84 & .80 & .74 & .55 & .63 & .78 & .93 & .85 & .82 & .54 & .65 & .68 & .68 & .68 & .61 & .42 & .53 \\
\bottomrule
\end{tabularx}
\end{table}

\begin{table}[H]
\centering
\scriptsize
\caption{Precision, recall and F1-score obtained by the Llama3-8B, Qwen3-32B and Zephyr-7B-beta models on each dataset for each class in a few-shot (FS) scenario.}
\label{tab:best_for_classes_FEW_EXP4}
\begin{tabularx}{\textwidth}{@{}p{1.6cm} *{18}{Y} @{}}
\toprule
\textbf{Dataset} & \multicolumn{6}{c}{\textbf{Llama3-8B (FS)}} & \multicolumn{6}{c}{\textbf{Qwen3-32B (FS)}} & \multicolumn{6}{c}{\textbf{Zephyr-7B-beta (FS)}} \\
\cmidrule(lr){2-7} \cmidrule(lr){8-13} \cmidrule(lr){14-19}
& \multicolumn{3}{c}{\textbf{Class 0}} & \multicolumn{3}{c}{\textbf{Class 1}} & \multicolumn{3}{c}{\textbf{Class 0}} & \multicolumn{3}{c}{\textbf{Class 1}} & \multicolumn{3}{c}{\textbf{Class 0}} & \multicolumn{3}{c}{\textbf{Class 1}} \\
\cmidrule(lr){2-4} \cmidrule(lr){5-7} \cmidrule(lr){8-10} \cmidrule(lr){11-13} \cmidrule(lr){14-16} \cmidrule(lr){17-19}
& P & R & F1 & P & R & F1 & P & R & F1 & P & R & F1 & P & R & F1 & P & R & F1\\
\midrule
Celebrity & .86 & .84 & .85 & .85 & .84 & .85 & .80 & .83 & .82 & .83 & .80 & .81 & .74 & .83 & .78 & .83 & .71 & .76 \\
Cidii & .88 & .96 & .92 & .93 & .81 & .87 & .97 & .77 & .86 & .75 & .96 & .84 & .89 & .60 & .72 & .61 & .89 & .73 \\
FakeVsSatire & .29 & .02 & .05 & .58 & .96 & .72 & .50 & .02 & .05 & .58 & .98 & .73 & .11 & .03 & .05 & .54 & .79 & .64 \\
Fakes & .53 & .95 & .68 & .41 & .04 & .07 & .53 & .82 & .64 & .47 & .17 & .25 & .52 & .77 & .62 & .45 & .21 & .28 \\
Horne & .88 & .60 & .71 & .58 & .87 & .69 & .96 & .59 & .73 & .58 & .96 & .73 & .80 & .56 & .66 & .54 & .76 & .64 \\
Infodemic & .81 & .87 & .84 & .84 & .77 & .81 & .82 & .83 & .82 & .81 & .80 & .81 & .77 & .99 & .86 & .98 & .67 & .79 \\
Isot & .57 & 1.00 & .72 & .99 & .29 & .45 & .67 & 1.00 & .80 & .99 & .53 & .70 & .55 & 1.00 & .71 & .99 & .22 & .37 \\
LIAR-PLUS & .64 & .63 & .64 & .54 & .55 & .55 & .71 & .40 & .51 & .51 & .79 & .62 & .61 & .87 & .72 & .66 & .30 & .41 \\
NDF & .78 & .77 & .77 & .65 & .65 & .65 & .88 & .79 & .83 & .72 & .84 & .77 & .81 & .84 & .83 & .73 & .69 & .71 \\
Politifact & .76 & .74 & .75 & .81 & .60 & .69 & .80 & .89 & .85 & .77 & .62 & .68 & .72 & .60 & .65 & .77 & .56 & .65 \\
\bottomrule
\end{tabularx}
\end{table}

\subsection{Statistical tests}

\begin{figure}[!htb]
\centering
\includegraphics[width=0.8\linewidth]{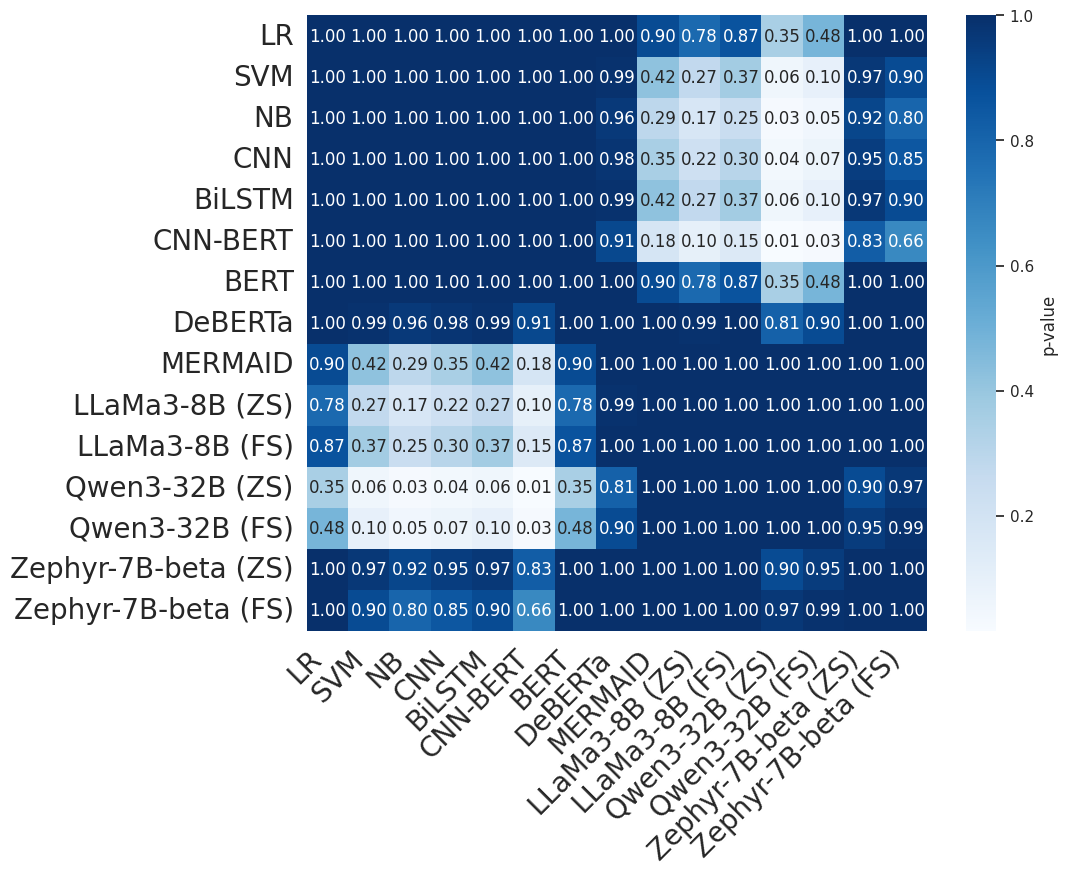} 
\caption{Heatmap of the p-values computed by the post-hoc Nemenyi Test (Pairwise Model Comparisons).}
\label{fig:heat4} 
\end{figure}

We obtained a Friedman Test statistic of 49.97, a p-value of 6.1e-06, and a Kendall's W of 0.357. Since the p-value is below the significance threshold alpha=0.05, the post-hoc Nemenyi test is then performed to identify the specific pairs of models exhibiting statistically significant differences. Figure \ref{fig:heat4} reports the pairwise p-values obtained from the Nemenyi post-hoc test, where values below 0.05 indicate  statistically significant performance differences.

\FloatBarrier

\FloatBarrier

\end{document}